\documentclass[nohyperref]{article}

\usepackage{microtype}
\usepackage{graphicx}
\usepackage{subfigure}
\usepackage{booktabs} 

\usepackage{hyperref}

\usepackage[accepted]{icml2022}

\usepackage{amsmath}
\usepackage{amssymb}
\usepackage{mathtools}
\usepackage{amsthm}

\usepackage[capitalize,noabbrev]{cleveref}

\theoremstyle{plain}

\theoremstyle{definition}

\theoremstyle{remark}

\usepackage[textsize=tiny]{todonotes}
\icmltitlerunning{Deep Networks on Toroids: Removing Symmetries Reveals the Structure of Flat Regions in the Landscape Geometry}

\begin{document}

\twocolumn[
\icmltitle{Deep Networks on Toroids: Removing Symmetries Reveals the Structure of Flat Regions in the Landscape Geometry}




\begin{icmlauthorlist}
\icmlauthor{Fabrizio Pittorino}{yyy}
\icmlauthor{Antonio Ferraro}{yyy}
\icmlauthor{Gabriele Perugini}{yyy,sch}
\icmlauthor{Christoph Feinauer}{yyy}
\icmlauthor{Carlo Baldassi}{yyy}
\icmlauthor{Riccardo Zecchina}{yyy}
\end{icmlauthorlist}

\icmlaffiliation{yyy}{AI Lab, Institute for Data Science and Analytics, Bocconi University, 20136 Milano, Italy}
\icmlaffiliation{sch}{Dept. of Applied Science and Technology, Politecnico di Torino, 10129 Torino, Italy}

\icmlcorrespondingauthor{Fabrizio Pittorino}{fabrizio.pittorino@unibocconi.it}
\icmlcorrespondingauthor{Carlo Baldassi}{carlo.baldassi@unibocconi.it}

\icmlkeywords{Machine Learning, ICML}

\vskip 0.3in
]



\printAffiliationsAndNotice{} 

\begin{abstract}
We systematize the approach to the investigation of deep neural network landscapes by basing it on the geometry of the space of implemented functions rather than the space of parameters. Grouping classifiers into equivalence classes, we develop a standardized parameterization in which all symmetries are removed, resulting in a toroidal topology. On this space, we explore the error landscape rather than the loss.
This lets us derive a meaningful notion of the flatness of minimizers and of the geodesic paths connecting them.
Using different optimization algorithms that sample minimizers with different flatness we study the mode connectivity and relative distances.
Testing a variety of state-of-the-art architectures and benchmark datasets, we confirm the correlation between flatness and generalization performance; we further show that in function space flatter minima are closer to each other and that the barriers along the geodesics connecting them are small. We also find that minimizers found by variants of gradient descent can be connected by zero-error paths composed of two straight lines in parameter space, i.e. polygonal chains with a single bend.
We observe similar qualitative results in neural networks with binary weights and activations, providing one of the first results concerning the connectivity in this setting.
Our results hinge on symmetry removal, and are in remarkable agreement with the rich phenomenology described by some recent analytical studies performed on simple shallow models.

\end{abstract}

\section{Introduction}

The loss landscape of a typical deep neural network performing a supervised learning task is in general highly non-convex. Moreover, even small networks (by the current standards) have a huge number of configurations of small loss, corresponding to zero or near-zero  training error. In this sense, most modern networks operate in a strongly over-parameterized regime. Understanding how simple variants of first-order algorithms are able to escape bad local minima and yet avoid overfitting is a fundamental problem, which has received a lot of attention from several perspectives~\cite{belkin2019reconciling,rocks2020memorizing}.

A natural and promising approach for addressing this issue is to investigate the geometrical properties of the loss landscape.
Broadly speaking, there are two related but conceptually distinct main research directions in this area: one is about the dynamics of gradient-based learning algorithms (e.g.~\citet{feng2021inverse}); the other concerns a static description of the geometry, its overall structure and its relation to the generalization properties of the network on unseen data (e.g.~\citet{gotmare2018}). In this paper, we focus on the latter.

A first basic observation is that (near-)minimizers of the loss, corresponding to (near-)zero training error, can have dramatically different generalization properties~\cite{keskar2016large,bad_minima_sgd,pittorino2021}. A growing amount of evidence shows a consistent correlation between the flatness of the minima of the loss and the test accuracy, across a large number of models and with several alternative measures of flatness, see e.g.~\citet{dziugaite2017computing,Jiang2020Fantastic,pittorino2021,yue2020salr}. Moreover, several studies indicate that stochastic gradient descent (SGD) and its variants introduce a bias, compared to full-batch gradient descent, towards flatter minima~\cite{keskar2016large,chaudhari2018stochastic,feng2021inverse,pittorino2021}. This effect seems to be amplified by other operating procedures, e.g. the use of the cross-entropy loss function, drop-out, judicious initialization, ReLU transfer functions~\cite{baldassi2018role,Baldassi20,relu_locent,bad_minima_sgd,zhang2021variance}. Therefore, in practical applications, bad minima are seldom reported or observed,
even though they exist in the landscape.

In this paper, we present a coherent empirical exploration of the structure of the minima of the landscape. Our work has two main features that, taken together, sets it apart from the majority of existing literature (see also sec.~\ref{sec:related} on related work): 1) The main object of our study is the (train) error (also called ``energy'') landscape, rather than the landscape given by the objective loss with which networks are optimized; 2) Our underlying geometrical space and topology is that of networks, intended as functional relations, rather than the space of parameters. 

The first point is less crucial to our results, although it affects the second one. Providing a detailed description of the dynamics requires studying the train loss (usually the cross-entropy) landscape. However, we argue that the train error landscape is a similar but more basic object of study for a static analysis, since it is directly related to the observable behavior of the network, especially the generalization error. This is assuming that the end goal of the training is to obtain a classifier whose output is the $\mathrm{argmax}$ over the last layer. The objective loss function on the other hand uses the entire output of the layer, which is not normally needed after the training is complete. The train error landscape is also more amenable to theoretical analysis, e.g.~\citet{baldassi2015subdominant,dziugaite2017computing}.

The second point is inspired from similar considerations. We posit that, after training, two networks that implement the same input-output relation (not only on the training set, but on any input) must be identified, even if their parameterization differs, and as such we group them into equivalence classes. In order to define a topology and a metric over this space, we standardize the parameterization of the networks (not during the training, but only for the geometrical description), thereby removing the symmetries that affect the usual parameterizations. In most common networks there are two of them: a continuous one, a scale invariance that allows to renormalize the weights, and a discrete one, a permutation symmetry that reflects the fact that in a hidden layer the labels of units (or the labels of filters in convolutional layers) can be exchanged. The latter in particular means that
two networks may appear to be very distant from each other in parameter space even if they are very similar or even identical in the function they implement. For example, this could happen if we measure the Euclidean distance between the parameters of two networks, where one is identical to the other up to a permutation of the hidden units within the layers. The scale symmetry can also have this effect, and moreover it may affect many measures of flatness making them imprecise at best and misleading at worst~\cite{dinh2017sharp}. These problems obviously affect every other investigation, such as studying the paths that join two configurations.

Our approach is thus as follows: 1) we choose a normalization method that leaves the network behavior unchanged while fixing the norm of each of the hidden units, projecting them onto a hyper-toroidal manifold; 2) when we compare two networks, we normalize and align them first, and consider the geodesic paths between them in normalized space.
With respect to the original goal of exploring function space, this approach is approximate, mainly for computational reasons (all our procedures are efficient and have polynomial running time with the number of parameters), 
and because in our characterization of the landscape we neglect the biases and batch-norm parameters.
Yet, the approximation appears to be very good and the effect
is critical. We have applied these techniques to several networks (continuous and discrete\footnote{For the discrete networks that we consider the rescaling symmetry is substituted with another discrete (sign-reversal) symmetry.}, multi-layer perceptrons and convolutional networks) and datasets (both real-world and synthetic). In each case, we used different training protocols aimed at sampling zero-error configurations (also called solutions) with different characteristics. In particular, we compared the kind of solutions found by SGD or its variants with momentum, with the (typically flatter and more accurate ones) found by the Replicated-SGD (RSGD) algorithm \cite{pittorino2021}, and with some poorly-generalizing solutions found by  adversarial initialization \cite{bad_minima_sgd}. With these, we could explore the local landscape of each solution and the paths connecting any two of them.

Remarkably, our results display some qualitative features that are shared by all networks and datasets, but which are visible and stable only in the space of networks as described above, and not in that of their parameters. Besides confirming that flatter minima generalize better than sharper ones, we found that they are also closer to each other in parameter space\footnote{See Fig.~\ref{fig:distances} for a discussion concerning the distances among different types of solutions.} (with respect to the Euclidean distance for networks with continuous weights and with respect to the Hamming distance for networks with binary weights), and that geodesic paths between them encounter lower barriers. 
Also, with few exceptions, the solutions we find can be connected by paths of (near-)zero error with a single bend.
Overall, our results are compatible with the analysis of the geometry of the space of solutions in binary, shallow networks reported in~\citet{baldassi2021learning,baldassi2021unveiling}, according to which efficient algorithms target large connected structures of solutions with the more robust (flatter) ones at the center, radiating out into progressively sharper ones.

\section{Related Work\label{sec:related}}

Early works relating flatness and generalization performance are \citet{hochreiter1997flat, hinton1993keeping}. In \citet{keskar2016large}, the authors show that minimizers with different geometrical properties can be found by varying algorithmic choices like the batch-size.
In \citet{Jiang2020Fantastic} a large scale experiment exploring the correlation between generalization and different complexity measures reported some flatness-based measures as the most robust predictors of good generalization performance.
Several optimization algorithms explicitly designed for finding flatter minima have been presented in the literature, resulting in improved generalization performance in several settings~\cite{chaudhari2019entropy,pittorino2021,yue2020salr}.
Some analytical investigations of phenomena related to flatness, their relation to generalization and algorithmic implications can be found in~\citet{baldassi2015subdominant,baldassi2016unreasonable,zhou2018non,dziugaite2017computing}.

Several recent works analyze the topic of mode connectivity empirically and analytically. In \citet{draxler2018}, the authors construct low-loss paths between minima of networks trained on image data using a variant of the nudged elastic band algorithm \cite{jonsson1998nudged}. In the closely connected work \citet{garipov2018}, the authors develop a method for finding low-loss paths as simple as a polygonal chain with a single bend and use this insight for creating a fast ensembling method. In \citet{gotmare2018} the authors show that minima found with different training and initialization strategies can be connected by high accuracy paths. Notably for the present work, the authors note that some of these choices like batch size or the optimizers are expected to have an effect on the flatness of the found solutions. Mode connectivity in the context of adversarial attacks has been studied in \citet{Zhao2020Bridging}, where the authors also propose to exploit mode connectivity to \textit{repair} tampered models. 
In \citet{kuditipudi2019explaining} it is shown analytically that, given some suitable assumptions on the robustness of the network, a low-loss path can be constructed between the solutions of ReLU networks with multiple layers. A similar result is derived in \citet{shevchenko20a} for networks trained specifically with SGD. 

The importance of symmetries for the question of flatness has been highlighted in \citet{dinh2017sharp}, where symmetries are used to show that simple notions of flatness are problematic. In \citet{brea2019weight}, low-loss paths between minimizers are constructed that cross 'permutation points', which are points where the weights associated with two hidden neurons in the same layer coincide. In \citet{tatro2020}, a method for aligning neurons based on matching activation distributions is introduced and the authors show analytically and empirically that this method increases mode connectivity.
In \citet{singh2020model} a neuron alignment method based on optimal transport is introduced in the context of \textit{model fusion}, where one tries to merge two or more trained models into a single model. Among other results, the authors show that matching is crucial when averaging the weights of models trained in the same or different settings (for example trained on different subsets of the labels).
In \citet{entezari2021role}, the authors test the hypothesis that barriers on the \textit{linear} path between minima of ReLU networks found by SGD vanish if the permutation symmetry is taken into account. They present evidence in the form of extensive numerical tests, exploring different settings with respect to the width, depth and architectures used.

\section{Numerical tools for the study of the energy landscape geometry}

As stated in the introduction, our aim is to study the geometry of the error landscape (which we will also call the ``energy'' landscape) in the space of networks. 
In particular, after sampling solutions (zero-error configurations) with different characteristics we study their flatness profiles and two types of paths connecting them: \emph{geodesic} and \emph{optimized} paths. The latter are obtained by finding a midpoint between two networks, using it as the starting point of a new optimization process that reaches zero error, and then connecting the new solution to the two original endpoints via two geodesic paths (this is a modified procedure of what is called \emph{polygonal chains with one bend} in \citet{garipov2018}, where the authors use euclidean geometry and do not account for the permutation symmetry). To make these definitions precise, we first need to describe the tools by which we remove the symmetries in the neural networks (code available at \url{https://gitlab.com/bocconi-artlab/matchingnn}).

We will use the following notation. We denote a whole neural network (NN) with $\Theta$, and with $L$ the number of its layers. We use the index $\ell=1,\dots,L$ for the layers, and denote the number of units in the layer by $H^\ell$. Each layer has associated parameters that we collectively call $\theta^\ell$ (which include batch-norm parameters if used). The input weights for the $k$-th unit are $w^\ell_k$ and its bias $b^\ell_k$. For all layers except the last, the activations are $x^\ell_k=\mathrm{ReLU}\left(\Delta^\ell_k\right)$, where $\Delta^\ell_k$ are the pre-activations. In binary networks we change the $\mathrm{ReLU}$ with a $\mathrm{sign}$. The last layer is linear and the output of the network is given by $\mathrm{argmax}_k (x^L_k)$. In binary classification tasks we use a single output unit, and a $\mathrm{sign}$ instead of an $\mathrm{argmax}$.

\subsection{Breaking the symmetries}



\subsubsection{Normalization}

Networks with continuous parameters and $\mathrm{ReLU}$ activation functions are invariant to the rescaling of the weights of the units in each hidden layer by a positive real value (that may be different for each unit), since the $\mathrm{ReLU}$ function has the property $\mathrm{ReLU}(a \Delta)=a\,\mathrm{ReLU}(\Delta)$ for any $a>0$. To break this symmetry, we apply the following normalization procedure, starting from the first layer and proceeding upward, see alg.~\ref{alg:rescaling}.
\begin{algorithm}[ht]
   \caption{Neural Network Normalization}
\begin{algorithmic}
   \STATE {\bfseries Input:} A NN with continuous weights, $L$ layers, parameters $\{\theta^{\ell}\}_{\ell=1}^{L}$ and ReLU activations.
   \FOR{$\ell=1$ {\bfseries to} $L-1$}
       \STATE $\{|w^{\ell}_k|\}_{k=1}^{H^{\ell}} = \mathrm{ComputeUnitNorms}(\theta^{\ell})$
       \STATE $\mathrm{Rescale}(\theta^{\ell}, \{|w^{\ell}_k|\}_{k=1}^{H^{\ell}})$ 
       \STATE $\mathrm{InverseRescale}(\theta^{\ell+1},\{|w^{\ell}_k|\}_{k=1}^{H^{\ell}})$ 
   \ENDFOR
   \STATE $|w|^L = \mathrm{ComputeLayerNorm}(\theta^L)$
   \STATE $\mathrm{Normalize}(\theta^L, |w|^L)$
\end{algorithmic}
\label{alg:rescaling}
\end{algorithm}

For each layer $\ell$, starting from the first and up to $L-1$, we calculate the norm $|w_k^{\ell}|$ of each hidden unit $k$ in the layer, and we multiply its incoming weights, bias and batch-norm parameters by $|w_k^{\ell}|^{-1}$, and its outgoing weights by $|w_k^{\ell}|$. As a result, all the units in the layer have norm $1$, while the function realized by the network (and also its loss) remains unaffected. The last layer does not have this symmetry, but it can be globally rescaled by a positive factor since the output of the network is invariant with respect to this operation (as a consequence of using the $\mathrm{argmax}$ operation). For consistency with the other layers, we normalize the last layer to $\sqrt{H^L}$.
The resulting space is a product of normalized hyper-spheres (one for each hidden unit apart from the ones in the last layer, which is globally normalized to a single hyper-sphere), inducing a generalized hyper-toroidal topology.

This normalization choice is rather natural, results in simple expressions for the computation of the geodesics (see below), and leads to sensible results; it is certainly not the only possible (or reasonable) one, and other possibilities might be worth exploring.

\subsubsection{Alignment}

Multi-layer networks (continuous or discrete) also have a discrete permutation symmetry, that allows to exchange the units in the hidden layers (neurons in fully-connected layers and filters in convolutional layers). When comparing two networks $\Theta_a$ and $\Theta_b$, we break the symmetry by first normalizing both networks, and then by applying the following alignment procedure, again starting from the first layer and proceeding upward, see alg.~\ref{alg:match_and_perm}.
\begin{algorithm}[ht]
   \caption{Neural Network Alignment}
\begin{algorithmic}
   \STATE {\bfseries Input:} Two normalized NNs of the same type $\Theta_a$, $\Theta_b$ with $L$ layers and parameters $\{\theta^{\ell}_a\}_{\ell=1}^L$, $\{\theta^{\ell}_b\}_{\ell=1}^L$.
   \FOR{$\ell=1$ {\bfseries to} $L-1$}
       \STATE $\pi^{\ell} = \text{Match}(\theta_a^{\ell}, \theta_b^{\ell})$ 
       \STATE PermutePrev$(\theta_b^{\ell}, \pi^{\ell})$ 
       \STATE PermuteNext$(\theta_b^{\ell+1}, \pi^{\ell})$
   \ENDFOR
\end{algorithmic}
\label{alg:match_and_perm}
\end{algorithm}

For each layer $\ell=1,\dots,L-1$, we use a matching algorithm to find the permutation $\pi$ of the indices of the second network that maximizes the cosine similarity\footnote{We can ignore the norms, thanks to our choice of the normalization and the fact that we do not match the last layer. For the same reasons, this is also equivalent to minimizing a squared distance.} $\sum_{k=1}^{H^\ell} (w^a)^\ell_k \cdot (w^b)^\ell_{\pi(k)}$. The permutation is applied to both the ingoing and the outgoing indices of the weights of layer $\ell$ of the second network (as well as other parameters associated to the units such as the biases and the batch-norm parameters).
For the matching algorithm, we use \texttt{linear\_sum\_assignment} from SciPy \cite{2020SciPy-NMeth}, which uses a quadratic modified Jonker-Volgenant algorithm. Any algorithm solving the minimum weight matching in bipartite graphs problem would be appropriate.

In the case of discrete networks, we use the $\mathrm{sign}$ activation function instead of the $\mathrm{ReLU}$, thus instead of a continuous rescaling symmetry there is a discrete sign-reversal one (allowing to flip the signs of all ingoing and outgoing weights of any hidden unit). We break this symmetry during the alignment step: in the matching step, we use the absolute value of the cosine similarities in the optimization objective, then we apply the permutation, and finally for each unit we either flip its sign or not based on which choice maximizes the cosine similarity.

This procedure is not guaranteed to realize a global distance minimization between the two networks in a worst-case scenario (and it is not clear whether such goal would be computationally feasible). It also does not take into account the biases, if present. However, it guarantees that if $\Theta_a$ and $\Theta_b$ are the same network with shuffled hidden units labels, at the end they will be perfectly matched almost surely. Furthermore, it is simple to implement, computationally efficient, and rather general (it applies to fully-connected and convolutional layers, to continuous or discrete weights). Basing the matching on the weights (as opposed to using e.g. the activations) is also data-independent and consistent with our overall geometrical picture.

\subsubsection{Geodesic paths}

Given two (non-normalized) continuous networks $\Theta_a$ and $\Theta_b$, we want to
consider the \emph{geodesic paths} between the networks in function space. Formally, this is defined as the shortest path between all possible permutations of the networks in the normalized space. 
We approximate this with the path between the normalized-and-aligned networks, which can be computed easily thanks to our choice of the normalization. Consider first a hyper-sphere of norm $n$, and two vectors $v_1,v_2$ on it. The angle between them is $\phi=\mathrm{arccos}(v_1\cdot v_2/n^2)$, and their (geodesic) distance is $n \phi$. A generic point along the geodesic, located at distance $x n \phi$ from $v_1$ where $x\in[0,1]$, can be expressed as $n\,\mathrm{Normalize}(v_1+t(x)(v_2-v_1))$ where
$t(x)=\left(1-\cos(\phi)+\sin(\phi)/\tan(\phi x)\right)^{-1}$.
This is easily extended to the metric on the full network: given two normalized-and-aligned networks, we can simply apply this formula independently (but with the same $x$) to each hidden unit of the first $L-1$ layers (with norm $1$) and to the last layer (with norm $\sqrt{H^L}$). Similarly, the overall squared geodesic distance is just the sum of the squared geodesic distances within each spherical sub-manifold.

For discrete binary networks we use the Hamming distance to measure the discrepancy between two (aligned) networks. In this case there is no analog to the geodesic path, since there are multiple shortest paths connecting any two networks: each path corresponds to a choice of the order in which to change the weights that differ between the two networks. In our tests, we simply pick one such path at random.
In this case, all curves are averaged over different random paths realizations, and although there is of course some variability (see error bars in the plots), it does not affect the results. Optimizing the curves using e.g. simulated annealing turned out to be too computationally expensive.

\subsection{\label{sec:hunt}Minima with Different Flatness and Generalization}

We sample different kinds of solutions by using different algorithms. The $\emph{standard}$ minima are found by using the Stochastic Gradient Descent (SGD) algorithm with Nesterov momentum. In order to find $\emph{flatter}$ minima, we use replicated-SGD (RSGD), see \citet{pittorino2021}, which was designed for this purpose. To find $\emph{sharper}$ minima, we use the the adversarial initialization described in \citet{bad_minima_sgd} followed by the SGD algorithm without momentum. We call this method ADV; it was developed to overfit the dataset and produce poorly generalizing solutions, but since there is a known correlation between flatness and generalization error, we expect (and indeed confirm a posteriori) that these solutions are sharp.

In all cases, we do not use $\ell_2$ regularization or data augmentation; the only image pre-processing in our experiments is normalization of images to zero mean and unit variance.

In the case of discrete binary networks we used the same techniques, but based on top of the BinaryNet training scheme (see e.g. \citet{simons2019review}), which is a variant of SGD tailored to binary weights. Our implementation differs from the original one~\cite{hubara2016binarized} in that the output layer is also binary
(details in SI Sec.~\ref{SI:binaryNNs}).



\section{Neural Networks with Continuous Weights}
In this section we present results on the energy landscape geometry and connectivity obtained on NNs with continuous weights. We consider 3 different architectures of increasing complexity: a) a multi-layer perceptron (MLP) with $2$ hidden layers of $512$ units; b) a small LeNet-like architecture with $2$ convolutional layers of $20$ and $50$ $5\times5$ filters followed by a $2\times2$ MaxPool and one fully-connected layer of $500$ units; c) the VGG$16$ architecture~\cite{zhang2015accelerating} with batch normalization (the only network we train with batch-normalization: the corresponding parameters and running means/variances have to be permuted/rescaled coherently with the other layers' parameters). We train architectures a) and b) on MNIST, Fashion-MNIST and CIFAR-10, while architecture c) is trained on the CIFAR-10.

\begin{figure}[t]
  \centering
    \includegraphics[width=1.0\linewidth]{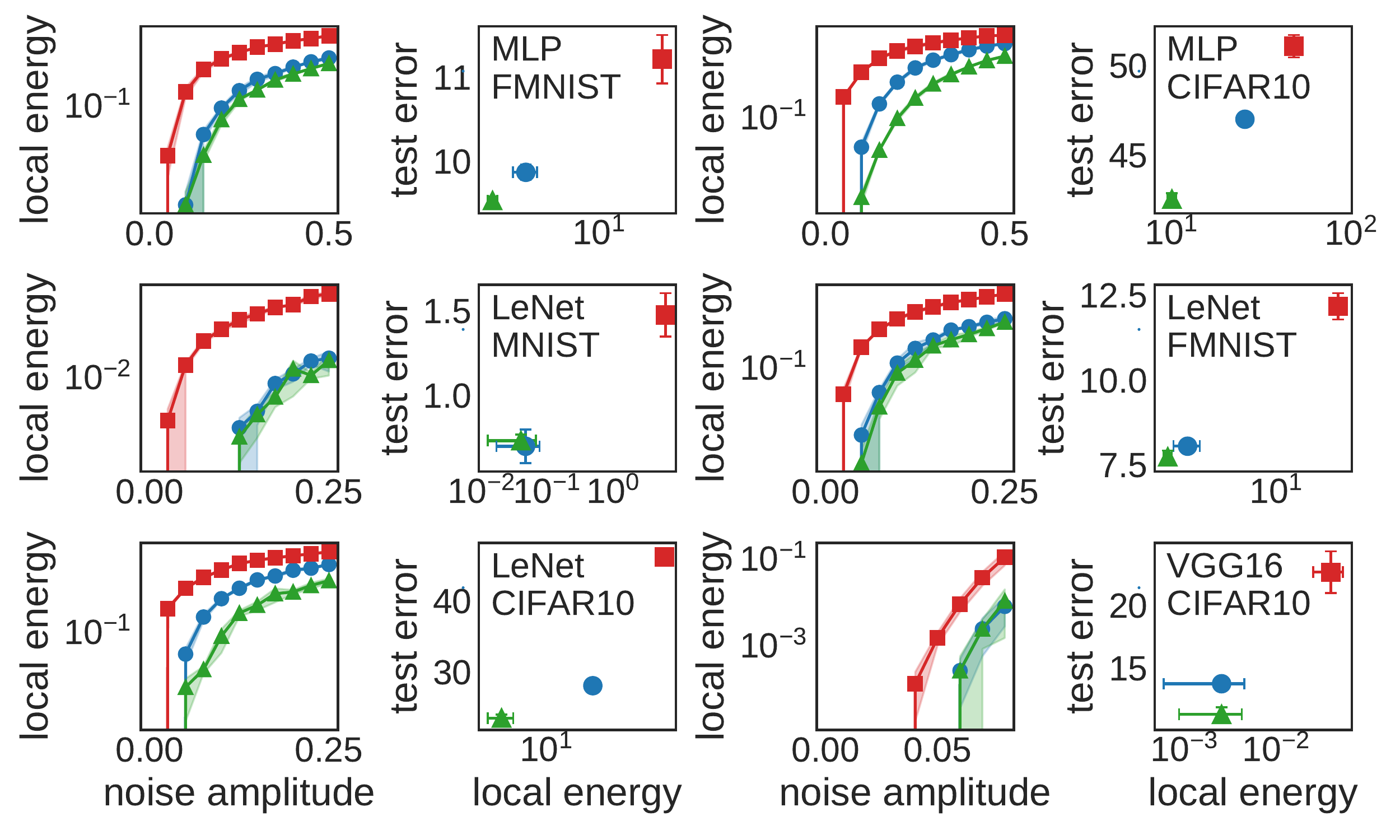}
    \vskip -0.05in
  \caption{Flatness and generalization for several algorithms. Green triangles: RSGD; blue circles: SGD; red squares: ADV. Columns 1 and 3: local energies as a function of noise amplitude; columns 2 and 4: corresponding local energies (at maximal reported noise) versus test error. Shades and error-bars are standard deviations (over $100$ sampled configurations for each noise amplitude).
  }
  \label{fig:locen}
    \vskip -0.2in
\end{figure}
\subsection{Flatness and generalization}
We train these 3 networks with the 3 algorithms described in Sec.~\ref{sec:hunt} (RSGD, SGD, ADV) for $300$ epochs, sufficient to find a configuration at zero training error (or with error $<1\%$ \cite{Jiang2020Fantastic}). In Fig.~\ref{fig:locen} we show that the flatness of the solutions, measured by the \emph{local energy}, follows the expected ranking between the algorithms, and the expected correlation with the generalization error, confirming that flatter minima generalize better.
The local energy $\delta E_\text{train}$ is defined as the average train error difference obtained by perturbing a given configuration $w$ with a multiplicative Gaussian noise of varying amplitude $\sigma$, in formulas: $\delta E_\text{train}(w, \sigma) = \mathbb{E}_z\, E_\text{train}(w+\sigma z \odot w) - E_{\text{train}}(w)$
where $\odot$ is element-wise multiplication and $z\sim\mathcal{N}(0,1)$.
We notice here that the local energy is one among possible flatness definitions (with respect to weights perturbations) that correlates with generalization, and that other choices are possible, e.g. the robustness with respect to input perturbations.

\subsection{Landscape geometry and connectivity}
\label{subsec:continuous_landscape}
Analyzing the connectivity of minima with different flatness levels allows us to explore the fine structure of the energy landscape, shedding light on the geometry of possible connected basins and/or on the presence of isolated regions.
To this aim we choose $5$ distinct pairs of solutions (in order to calculate standard deviations, represented by shaded areas in the figures) for each of the $6$ possible paths among the $3$ types of solutions: the paths between solutions of the same flatness (RSGD-RSGD, SGD-SGD, ADV-ADV) and of different flatness (RSGD-SGD, RSGD-ADV, SGD-ADV). 
We calculate the train error of configurations along the paths connecting these different solutions and define the barrier along the path as the highest encountered train error (solutions are at training error $<1\%$).

Let us notice that what we call \emph{energy} along the paper, i.e. the train error, is directly related to applications and exhibits more symmetries (e.g., in the norm of the last layer) than the objective loss with which the networks have been optimized (in our work the cross-entropy). When calculating the loss along the one-dimensional paths and the bi-dimensional sections, defining representatives for the equivalence classes is less straightforward, and results are less uniform across models. This is possibly due to the fact that the last layer, that is not normalized when considering the loss rescaling symmetry, may result in having very different norms across different solutions.

\begin{figure}[ht]
    \begin{center}
    \includegraphics[width=1.0\linewidth]{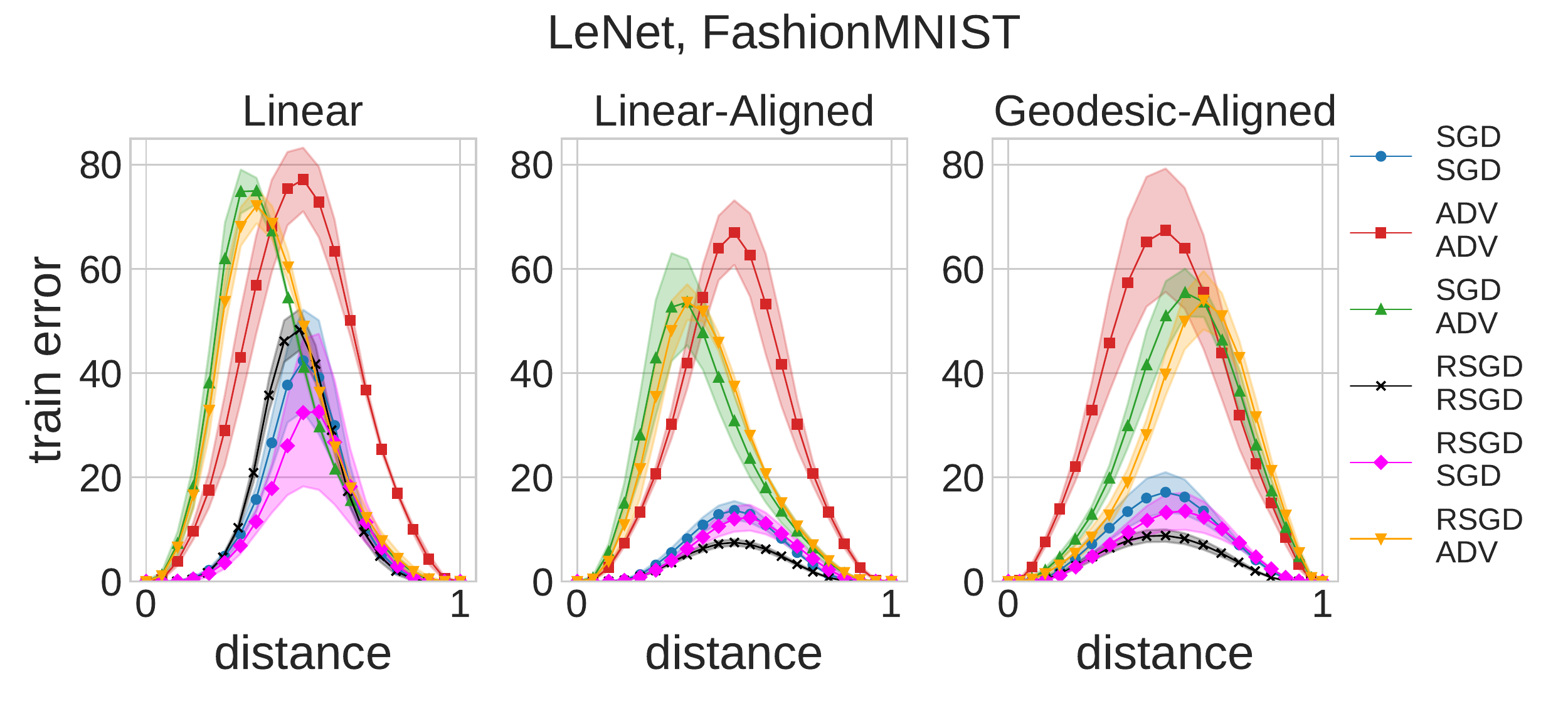}
    \includegraphics[width=0.33\linewidth]{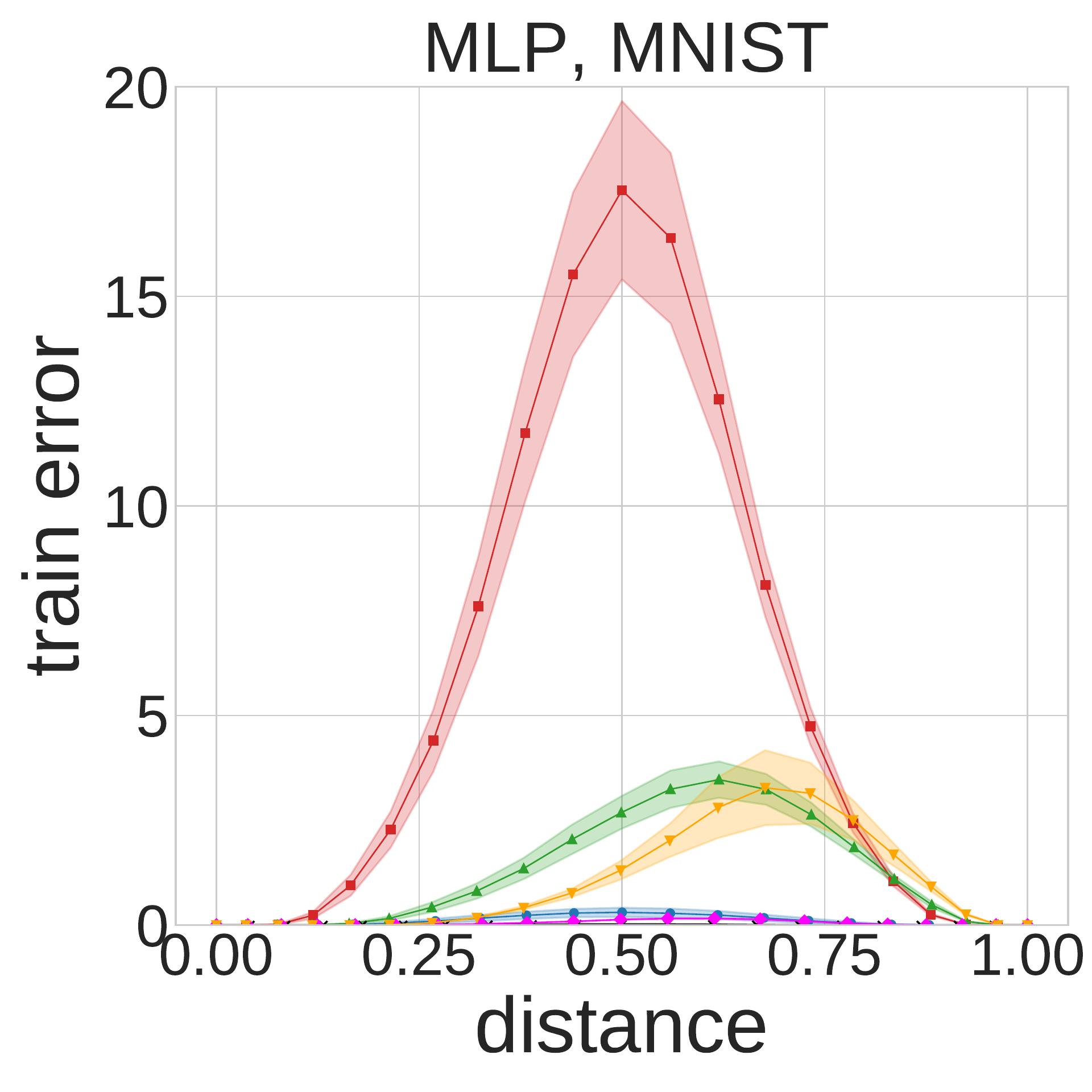}\includegraphics[width=0.33\linewidth]{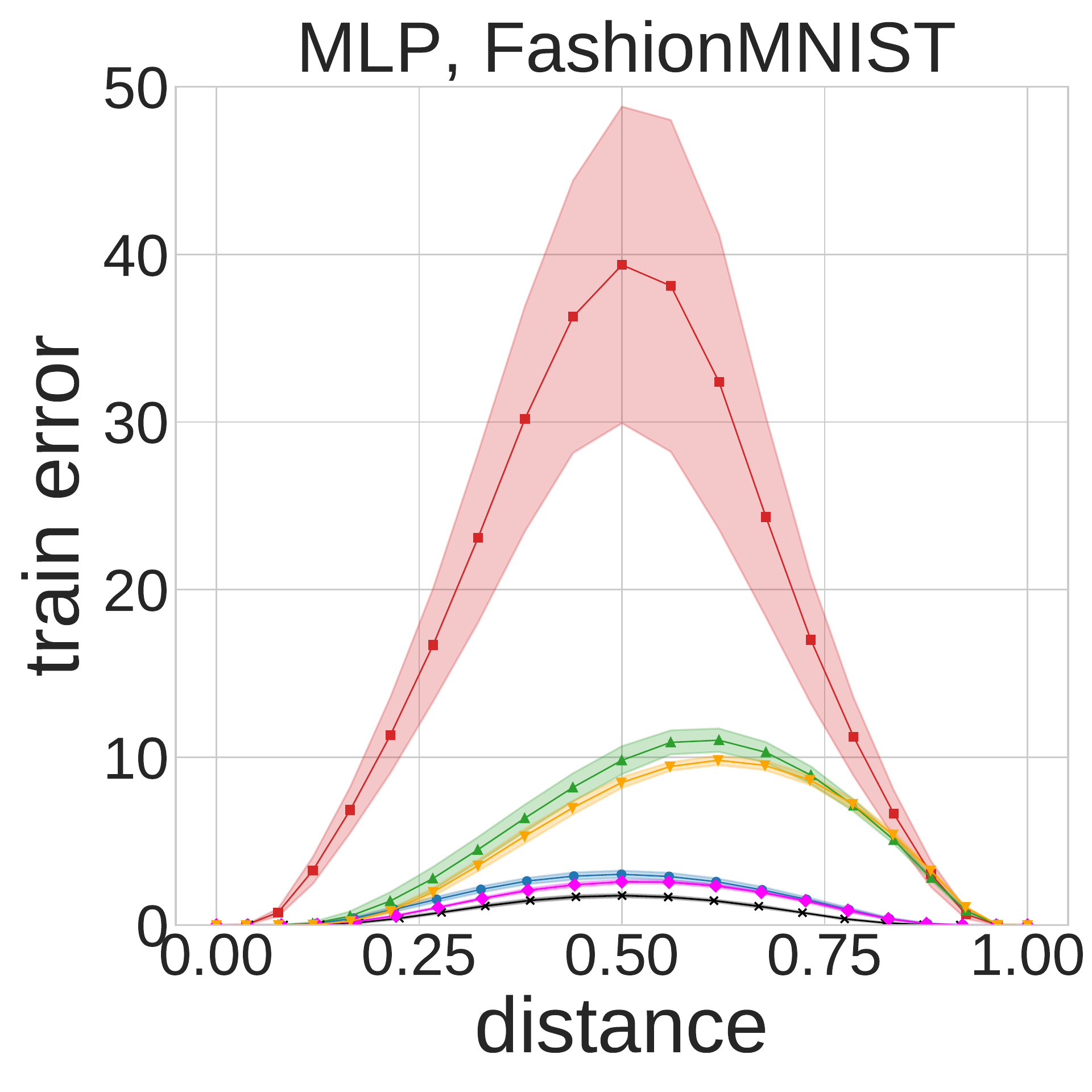}\includegraphics[width=0.33\linewidth]{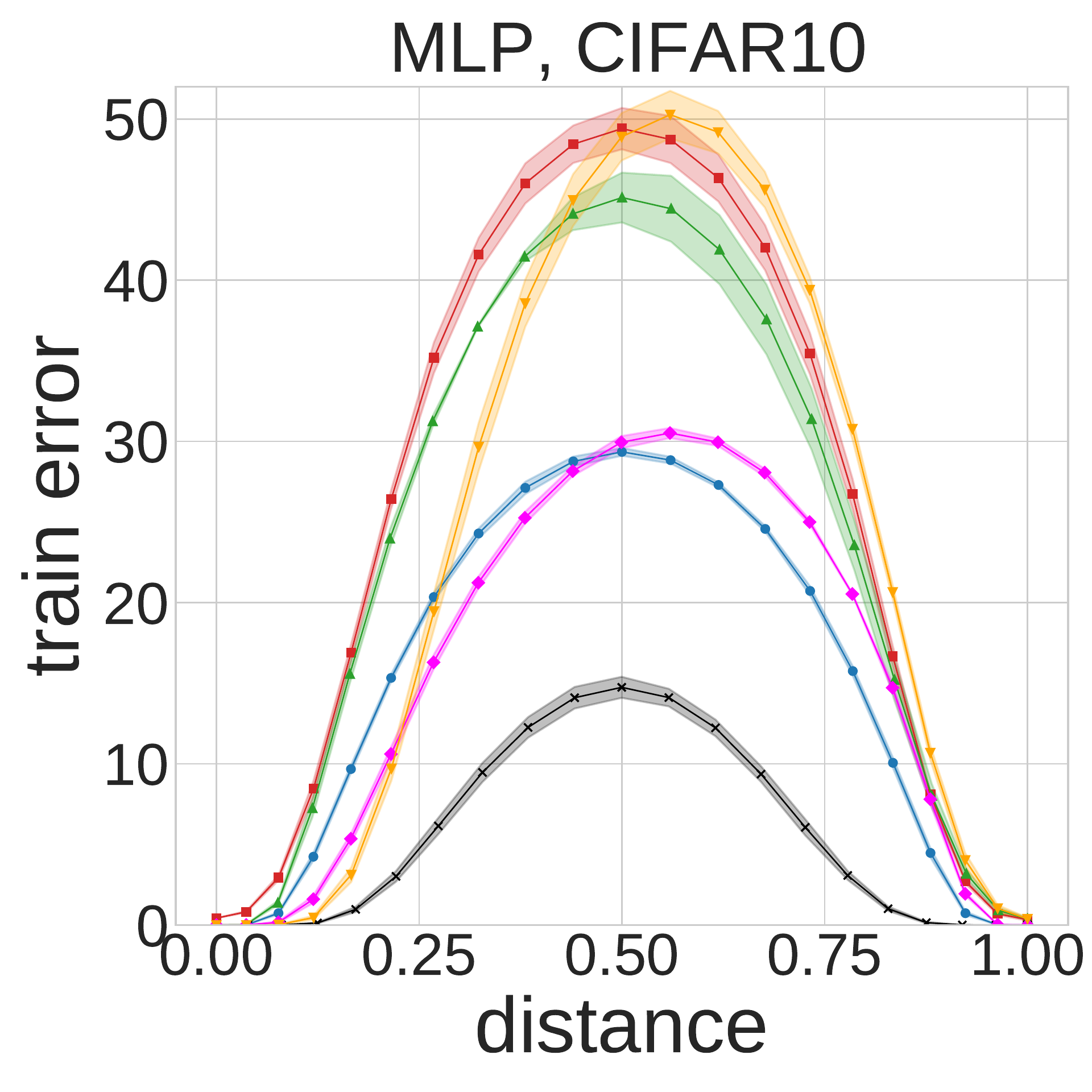}
    \includegraphics[width=0.33\linewidth]{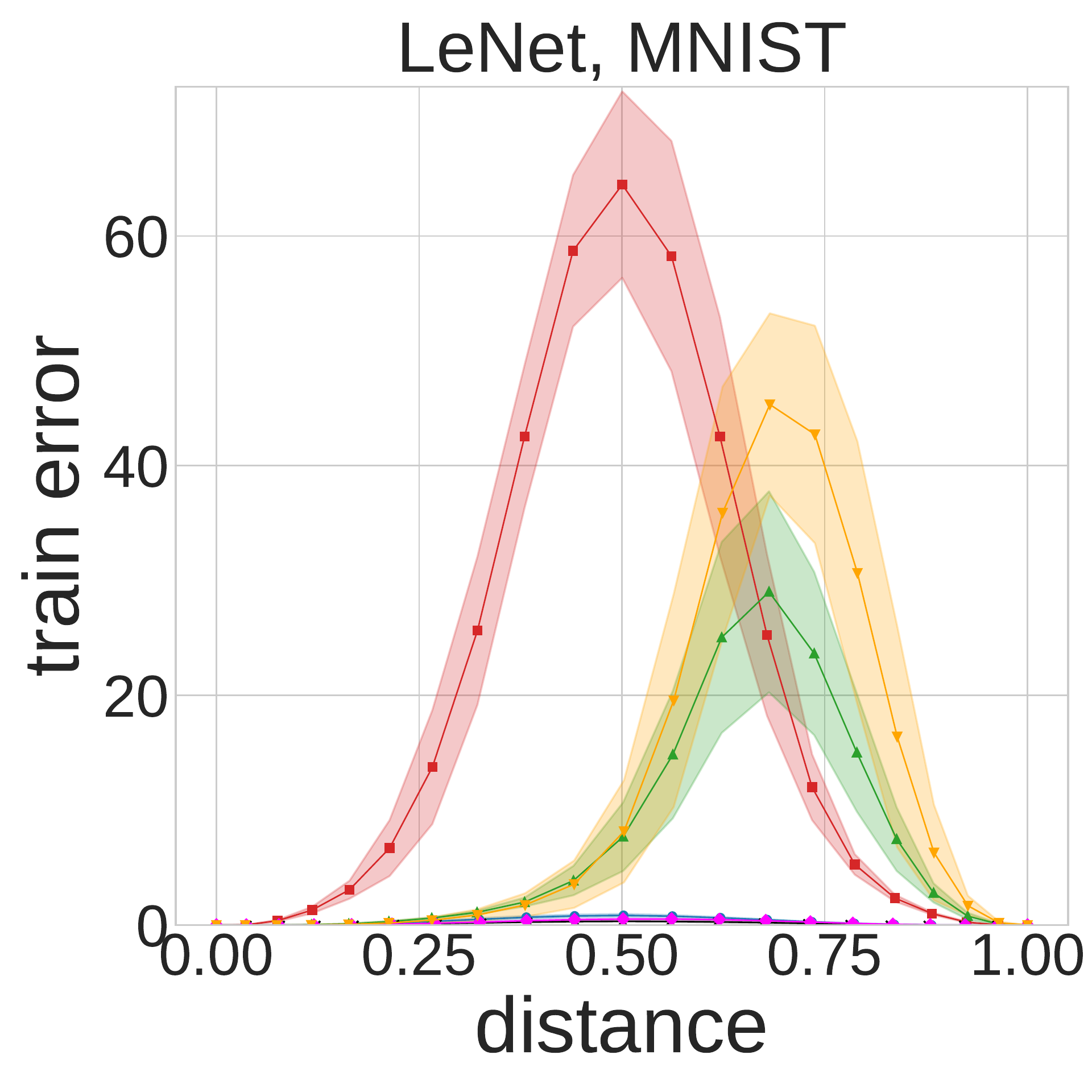}\includegraphics[width=0.33\linewidth]{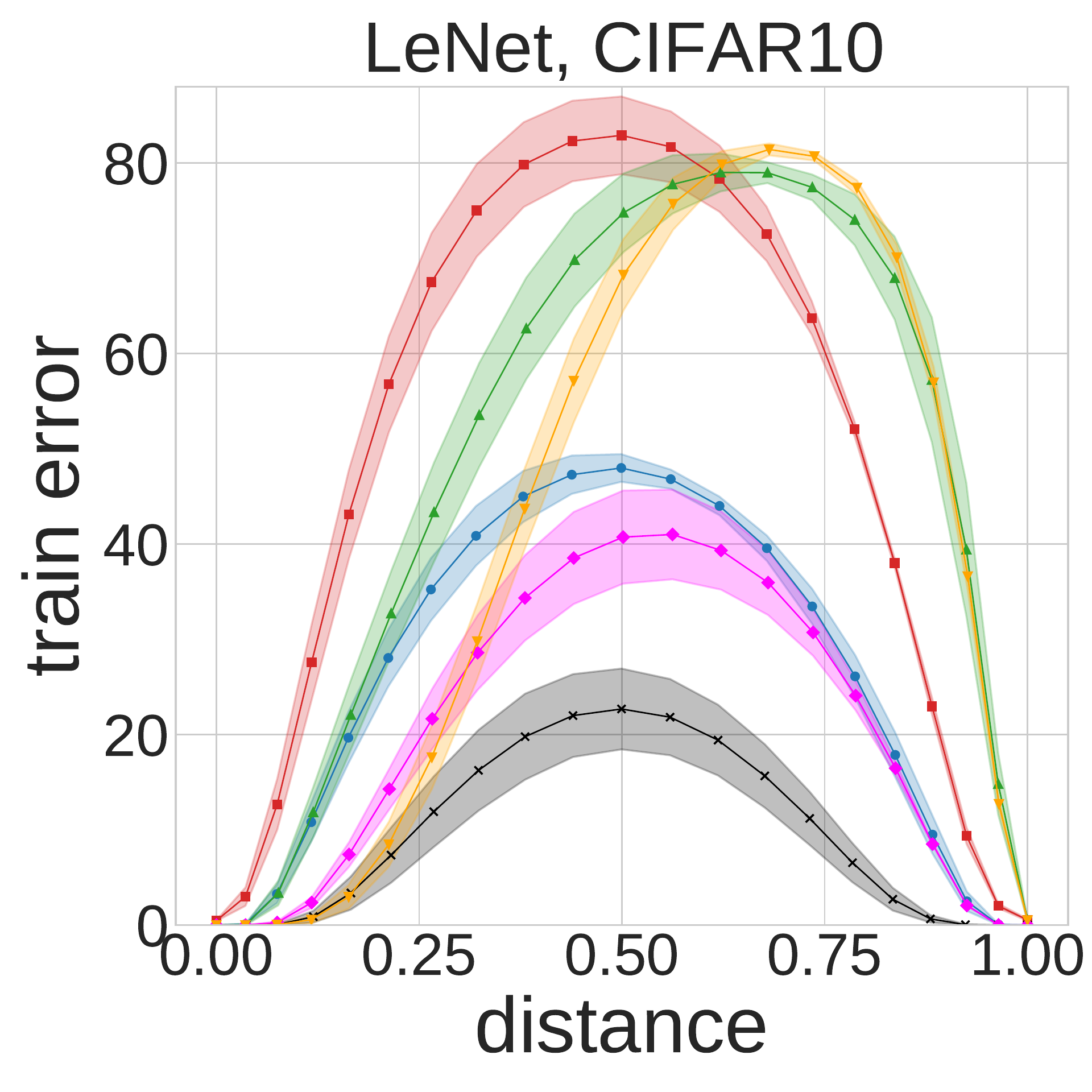}\includegraphics[width=0.33\linewidth]{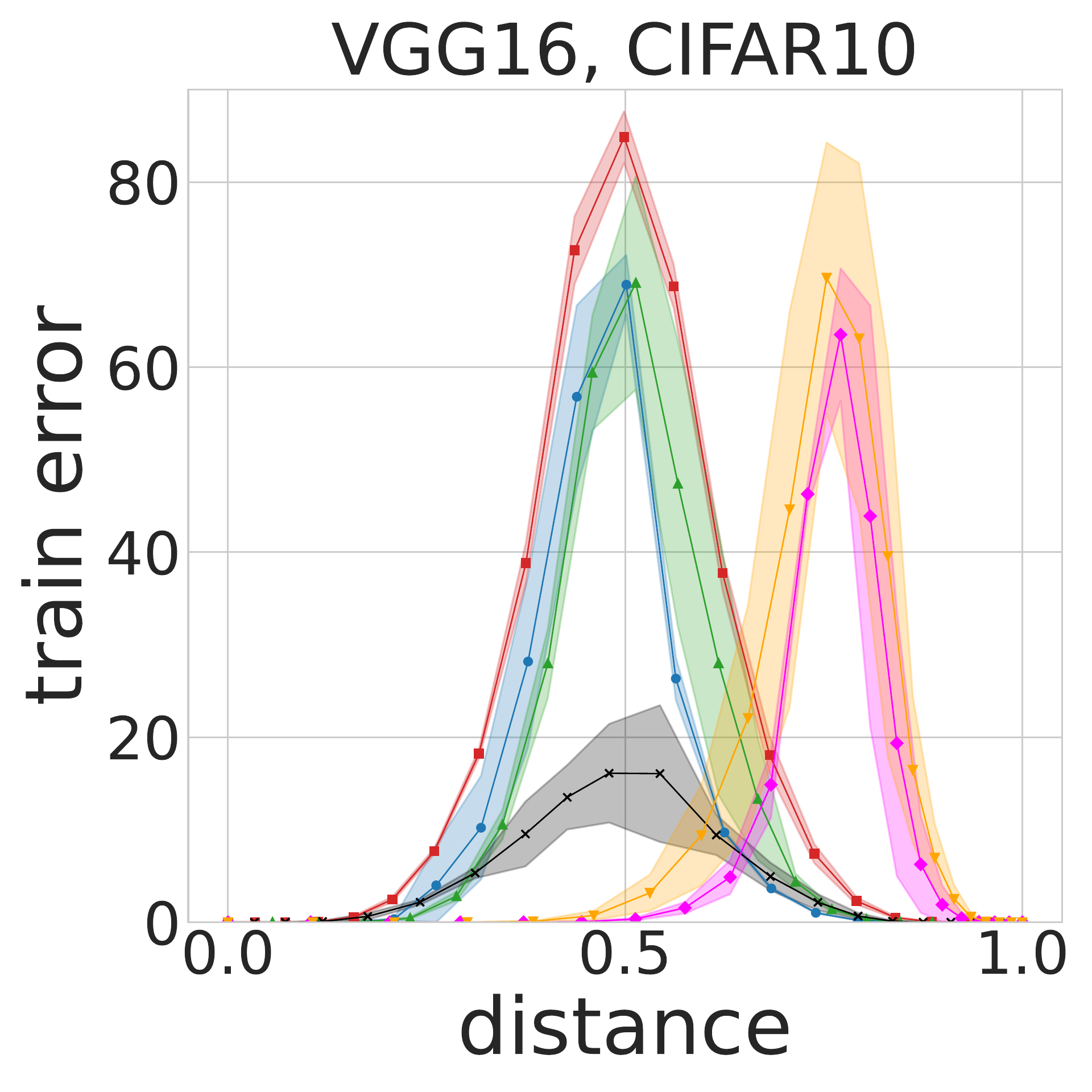}
    \caption{Error landscape along one-dimensional paths connecting minima of different flatness (minima appear on left/right following the up/down order of the legend). All distances have been rescaled in $[0,1]$. The top row highlights the general effect of the symmetries in a representative example (LeNet on Fashion-MNIST). All other panels report the geodesic-aligned paths representing the final result of our analysis. Shades are standard deviations over $5$ distinct pairs of solutions.}
    \label{fig:linearpaths}
    \end{center}
    \vskip -0.2in
\end{figure}

\paragraph{Geodesic paths}
We show the error along the geodesic paths connecting pairs of minima for the different networks and datasets in Fig.~\ref{fig:linearpaths}.
The top row in the figure demonstrates the effect of accounting for the symmetries, by showing three sets of curves - one set per panel, left to right: linear paths between raw configurations as output by the algorithms; linear paths in which the endpoint networks are aligned but not normalized\footnote{We still maximize the layers' cosine similarity when performing the alignment.}; the geodesic path between the normalized-and-aligned networks. We can see that: a) taking into account the permutation symmetry the barriers are lowered;
b) following the geodesic removes the distortion in the paths, such that they appear flatter towards flatter solutions as they should
c) considering these two symmetries the barriers heights (in particular their maximal value) follow a general overall ranking which is correlated to the flatness level of the corresponding solutions: the RSGD-RSGD one is consistently the lowest, followed by RSGD-SGD and SGD-SGD, then RSGD-ADV and SGD-ADV, and finally ADV-ADV is consistently the highest.

\begin{figure}[ht]
  \centering
    \includegraphics[width=1.0\linewidth]{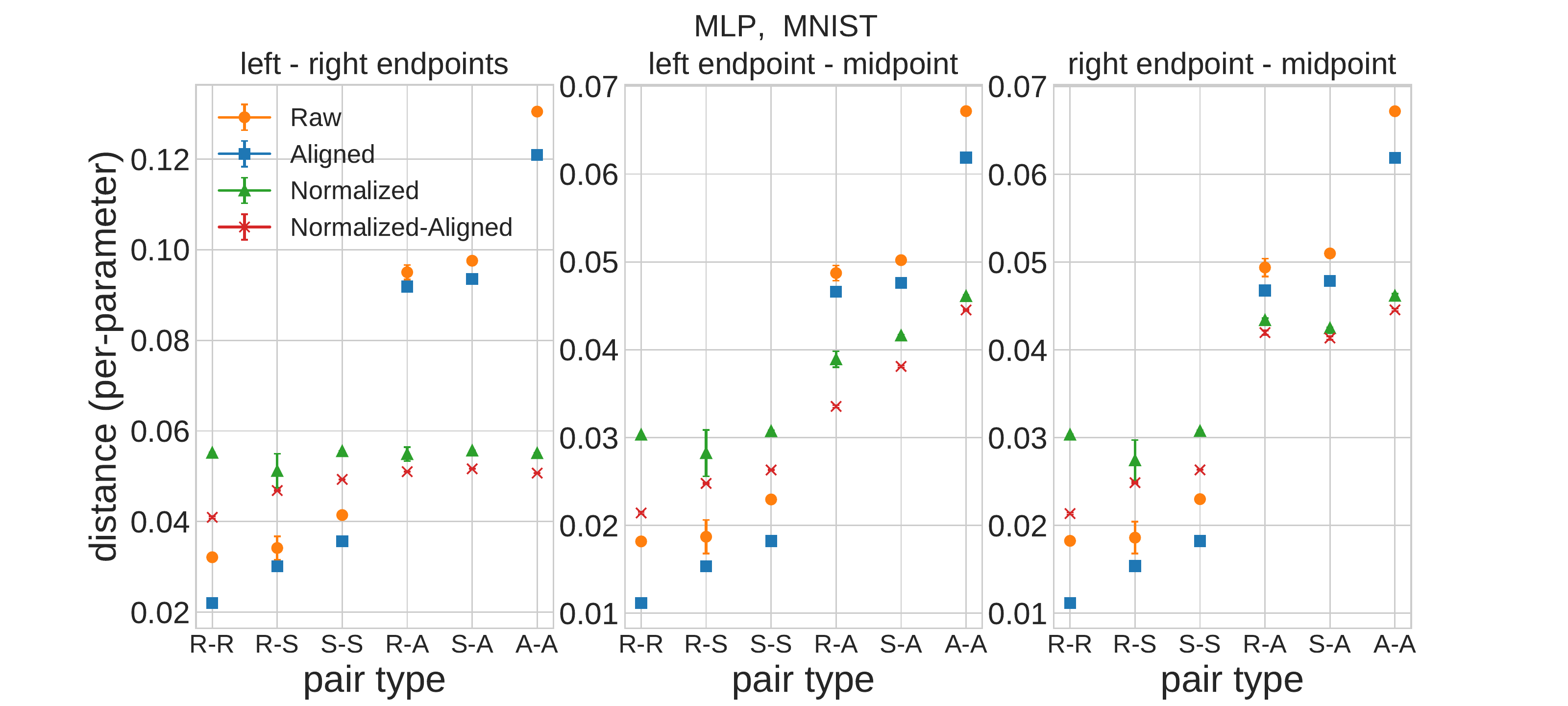}
  \caption{Distances between configurations, MLP on MNIST. (Left panel) Euclidean distance between configurations independently sampled by different algorithms (on the $x$-axis: R$\equiv$RSGD; S$\equiv$SGD; A$\equiv$ADV). (Middle panel) Euclidean distance between the optimized midpoint initialized as the mean of the two configurations in the $x$-axis with the one indicated on the left; (left panel) same as the middle panel but the distance is between the optimized midpoint and the configuration indicated on the right in the $x$-axis.}
  \label{fig:distances}
  \vskip -0.1in
\end{figure}

The top row of Fig.~\ref{fig:linearpaths} is a representative example; analogous figures for all the other networks/datasets considered in the paper are reported in the Appendix, Sec.~\ref{SI:1dpaths}. In all cases, accounting for the symmetries of the network is indeed critical to reveal these seemingly very general geometrical features.
Let us notice that while the principal effect on the barrier's height is due to the permutation symmetry, moving along the geodesic path reflects in the particular shape of the path itself and on distances along the path (see also Fig.~\ref{fig:distances}); as the train error is invariant w.r.t. the normalization procedure, large variations in barriers' height when considering this symmetry are not expected (but still possible due to the change in the path). In the case of solutions with different flatness level, a left-right asymmetry in the geodesic paths appears (even if not always with the same intensity), by coherently assigning a flatter profile towards the flatter configuration.

Notably, by sampling solutions with high flatness level using RSGD, we are able to target increasingly connected structures, even in cases like VGG in which they could appear as missing, see the black path in the lower-right panel of Fig.~\ref{fig:linearpaths}. We obtain, for example, much smaller barriers for linear mode connectivity than the ones found for the same type of network and dataset in \cite{entezari2021role}, although the barriers are still around $20\%$ of train error.

\paragraph{Optimized paths}
Our procedure to find the optimized paths is the following: we first align the two endpoint networks, then find their midpoint, optimize it for $300$ epochs with SGD in order to reach zero training error, and explore the two aligned geodesics connecting the optimized midpoint and the endpoints. The barrier in this case is defined as the highest training error encountered along the union of these two geodesics. We define the optimized path in this way \cite{garipov2018} because we seek a low-error path that is close to a straight path and also simple to find and define.

We report a representative example in Fig.~\ref{fig:optpaths} (for the other networks/datasets see Appendix Sec.~\ref{SI:1dpaths}). Again, we contrast three situations, one in which we use straight segments between non-normalized unaligned configurations, one where we remove the permutation symmetry, and one where we use the geodesic paths. The barriers are much lower than for the unoptimized paths even in the unaligned linear case, but removing the permutation symmetry has still an important effect in lowering the barriers, and in most cases (like the one shown here) it removes them entirely.
The presence of non-zero barriers in the SGD-ADV and RSGD-ADV cases in the left panel of Fig.~\ref{fig:optpaths}, which would seem counter-intuitive given the absence of barriers in the ADV-ADV case, is removed after taking symmetries into account, thus showing an example of the need of this operation in order to obtain uniform and interpretable results.
Intriguingly, in the (very high-dimensional) VGG case, the optimized barriers are smaller than $0.1\%$ in all cases (see lower-right panel of Fig.~\ref{fig:optSI}), even though the barriers in the linear connectivity remain significantly different from zero even after removing symmetries (see lower-right panel of Fig.~\ref{fig:linearpaths}).
\begin{figure}[ht]
    \begin{center}
    \includegraphics[width=1.0\linewidth]{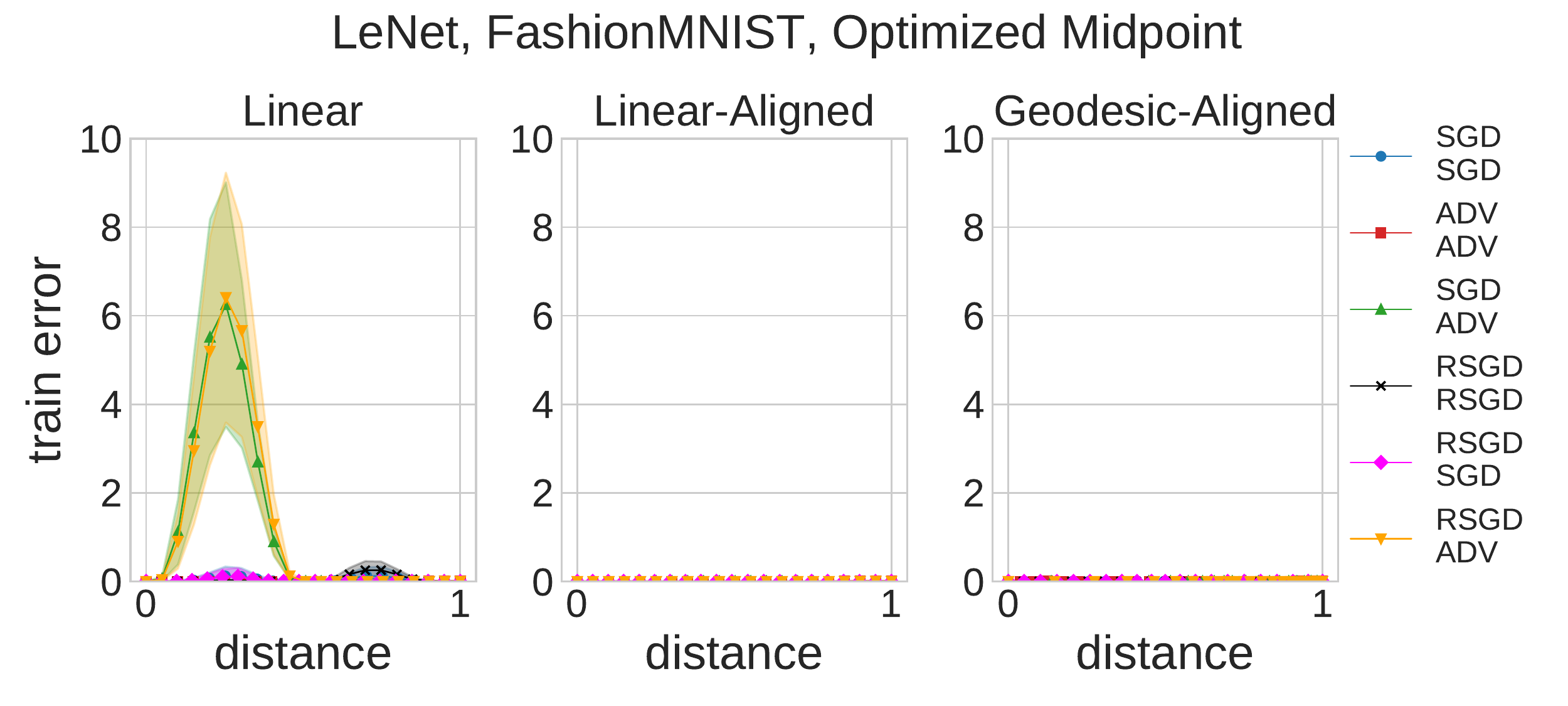}
    \caption{Removing the permutation symmetry eliminates the barriers that may appear in the single-bend optimized paths. Left to right: linear path; linear-aligned path; geodesic-aligned path. Shades are standard deviations over $5$ distinct pairs of solutions.}
    \label{fig:optpaths}
    \end{center}
    \vskip -0.2in
\end{figure}


\paragraph{Bi-dimensional visualizations}
\label{par:continuous_2dplots}
Following~\citet{garipov2018}, we study the error landscape on bi-dimensional sections of the parameters space using the Gram-Schmidt procedure. 
This procedure consists in considering the $3$ solution vectors that specify the bi-dimensional plane, defining an orthonormal basis of this plane containing these $3$ vectors, defining a Cartesian grid in this basis and evaluating the error of the networks corresponding to each of the points in the grid (see \citet{garipov2018} for details).

\begin{figure}[ht]
    \begin{center}
    \includegraphics[width=0.5\linewidth]{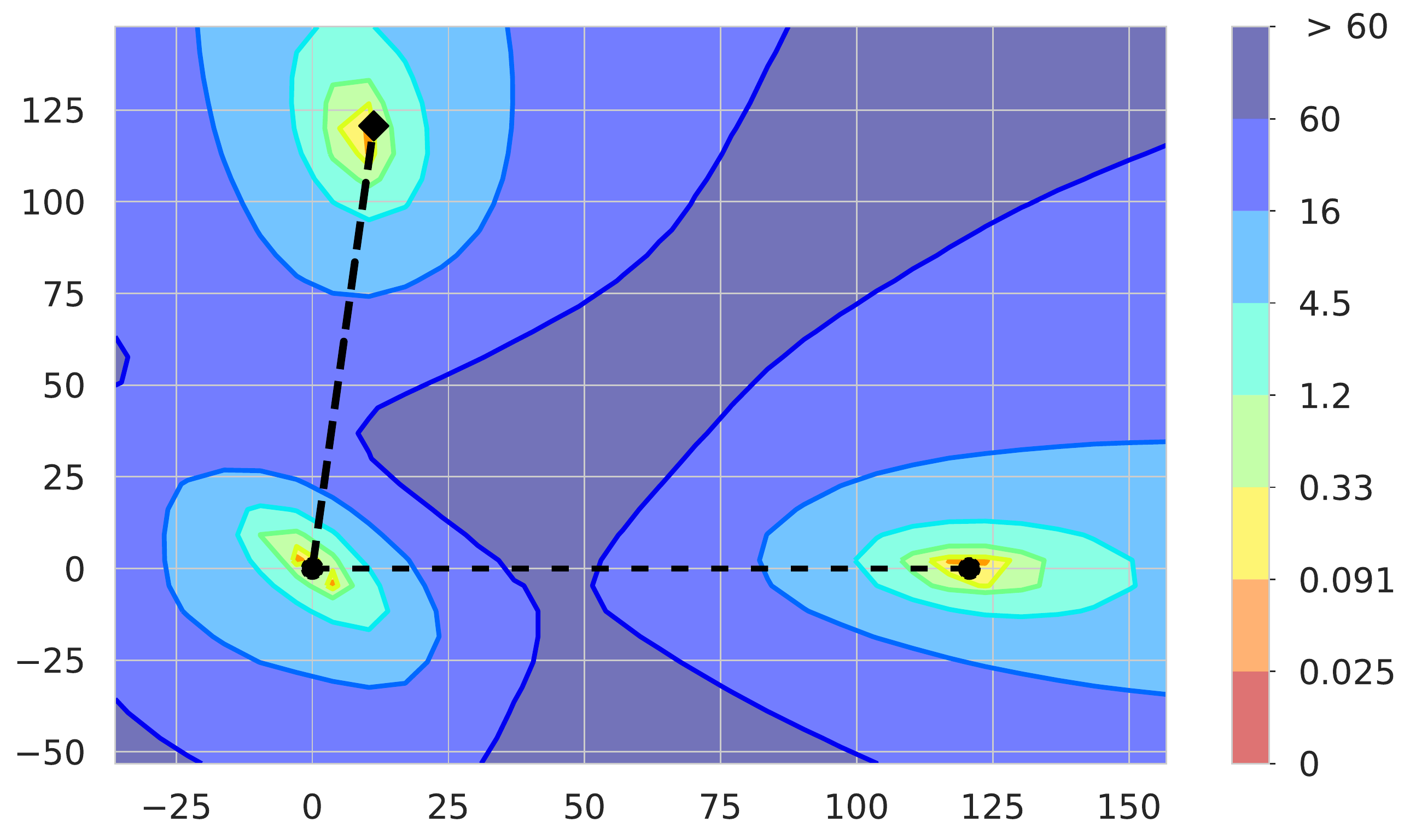}\includegraphics[width=0.5\linewidth]{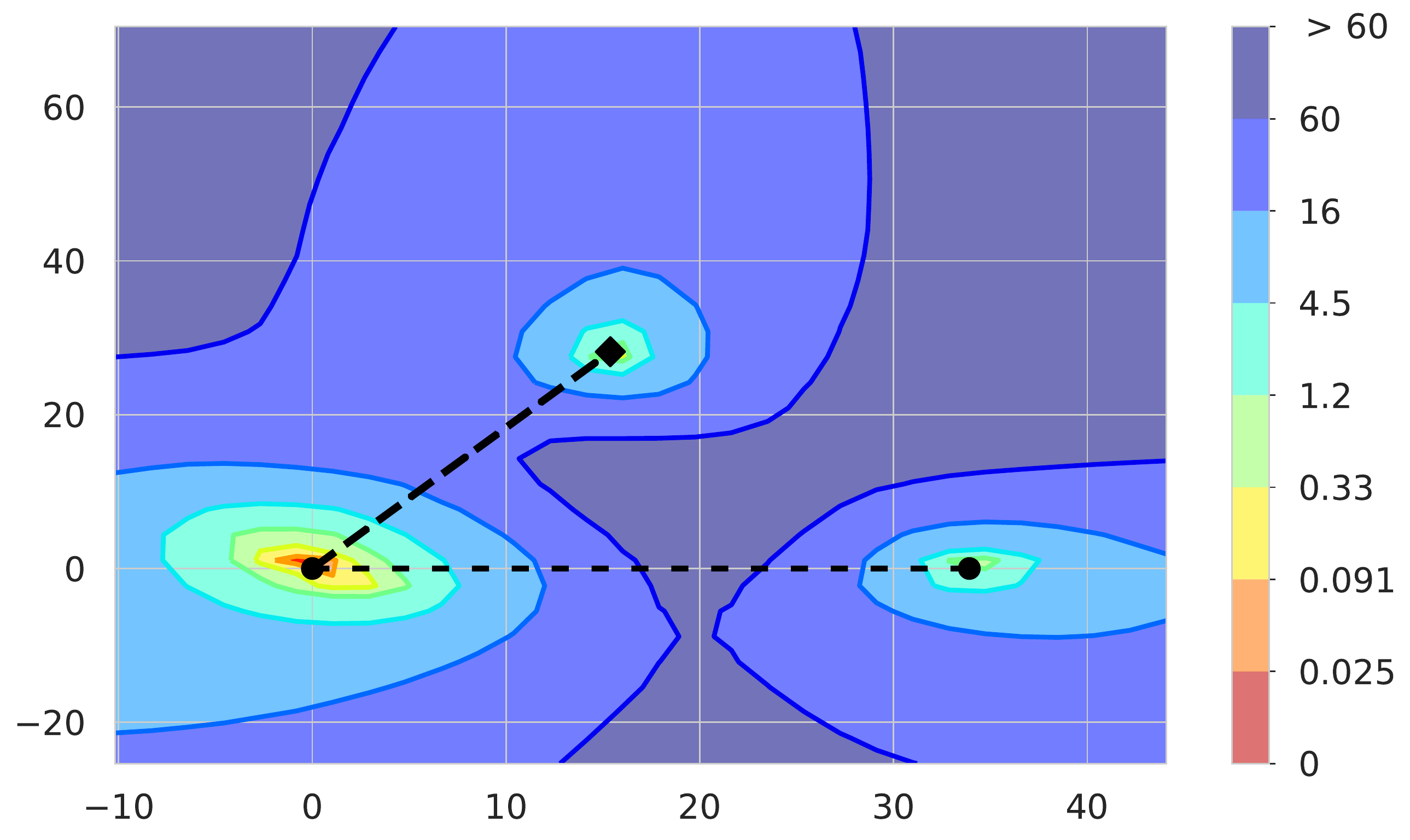}
    \includegraphics[width=0.5\linewidth]{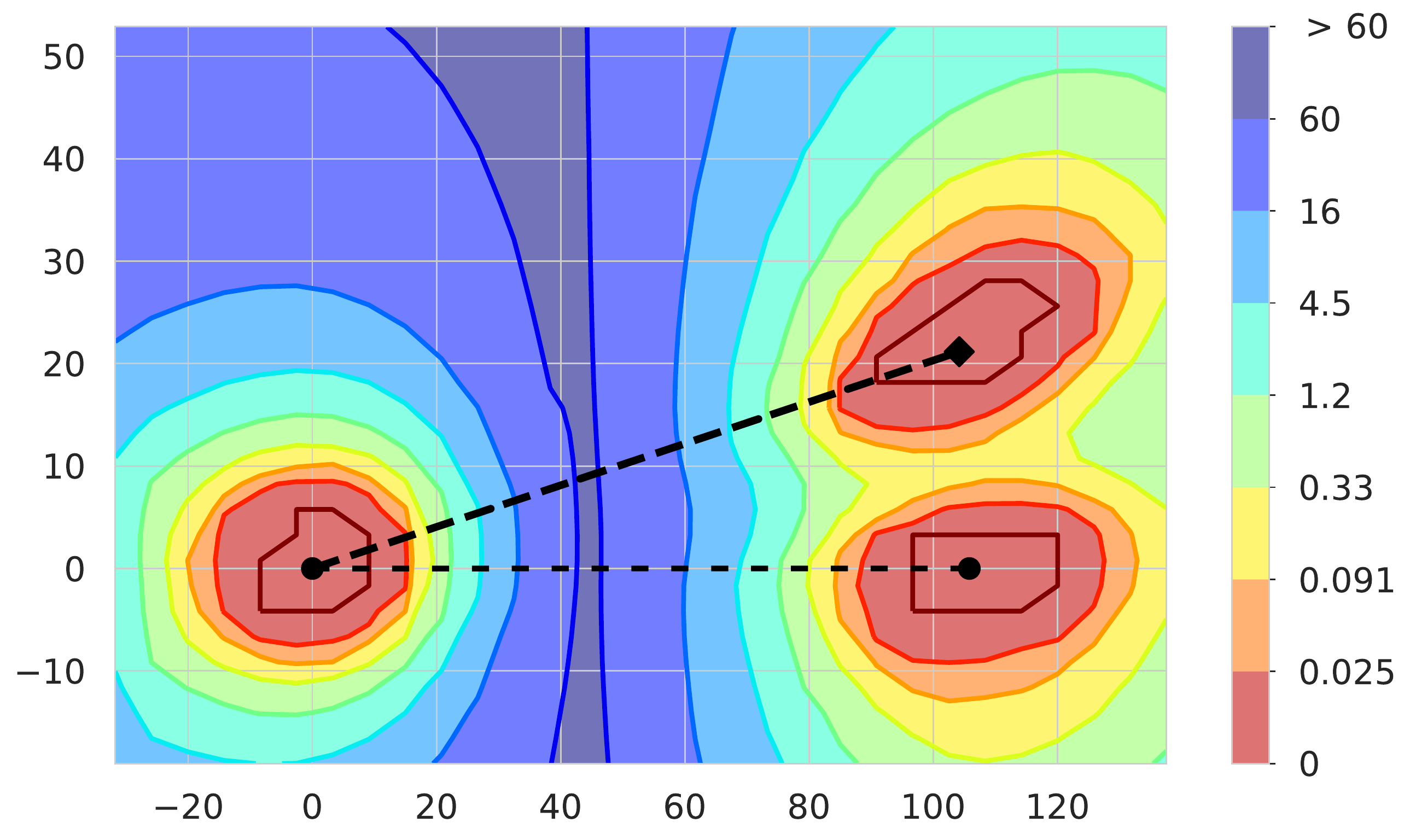}\includegraphics[width=0.5\linewidth]{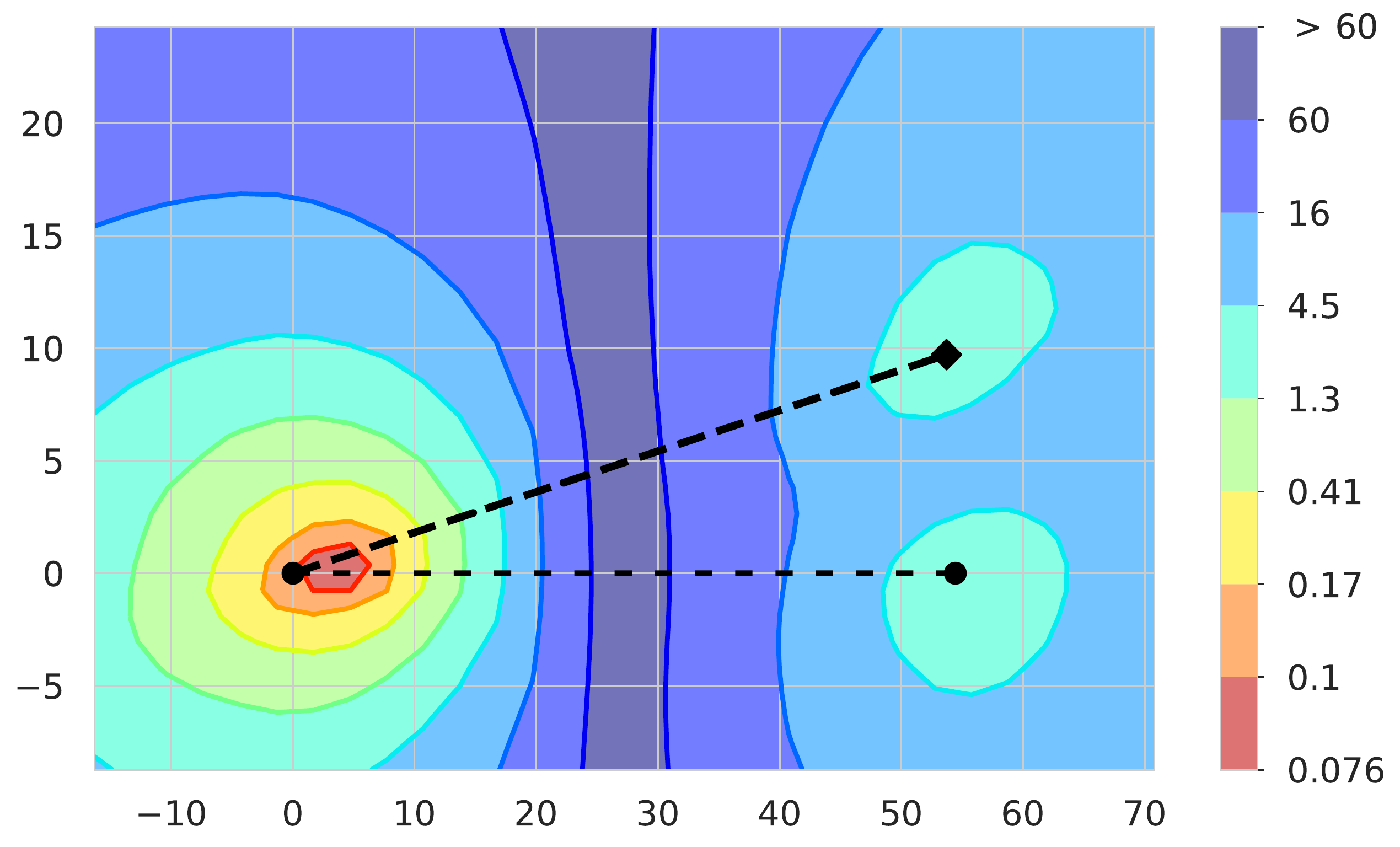}
    \caption{Bi-dimensional landscape sections obtained using the Gram-Schmidt procedure (see main text). Top row: LeNet, Fashion-MNIST. Bottom row: VGG16, CIFAR10. Left column: without normalization, lines are linear paths. Right column: with normalization, lines are distorted geodesics. In each panel the left point is RSGD, the right point unaligned ADV, the middle/top point is aligned ADV. Aligning the NNs can lower the barriers (not for VGG in this RSGD-ADV case), normalization reveals the geometry around them.
    }
    \label{fig:2dsurfacesmain}
    \end{center}
    \vskip -0.2in
\end{figure}

In Fig.~\ref{fig:2dsurfacesmain} we show the results on the planes defined by an RSGD solution (left-most point), an unaligned ADV solution (right-most point) and the corresponding aligned ADV solution (top point). We show the non-normalized (left panels) and normalized (right panels) cases. In the normalized case, the plane as a whole does not lie in normalized space (only the three chosen points do) and thus the plot suffers from some (presumably mild) distortion. Nevertheless, we can still see that accounting for the permutation symmetry alone reduces the distance between different minima and can lower the barriers between them, but also that only after normalization the expected relative sizes of the basins are revealed.

\paragraph{Distances.}
We studied the distances between pairs of solutions, categorized according to their flatness. Some representative results are reported in Fig.~\ref{fig:distances}, the full results are in Appendix Sec.~\ref{SI:distance}. When NNs are normalized and aligned, we consistently find that flatter solutions are closer to each other than sharper ones. In optimized paths, the optimized midpoints end up closer to the flatter solutions. Although our sampling is very limited, these results (together with all the previous ones) are compatible with the octopus-shaped geometrical structures described in~\cite{baldassi2021unveiling}.


\section{Neural Networks with Binary Weights}
\label{sec:binaryNNs}

In this section we present some results on the error landscape connectivity of binary NNs, in which each weight (and activation) is either $1$ or $-1$, a topic that is almost absent in literature.



We considered two main scenarios: shallow networks on synthetic datasets, for which a comparison with some theoretical results is possible, and deep networks, both MLP and convolutional NNs, on real data.



\subsection{Over-parameterized Shallow Networks on Synthetic Datasets}
\label{subsec:binary_shallow}

\begin{figure}[ht]
    \begin{center}
    \includegraphics[width=0.45\linewidth]{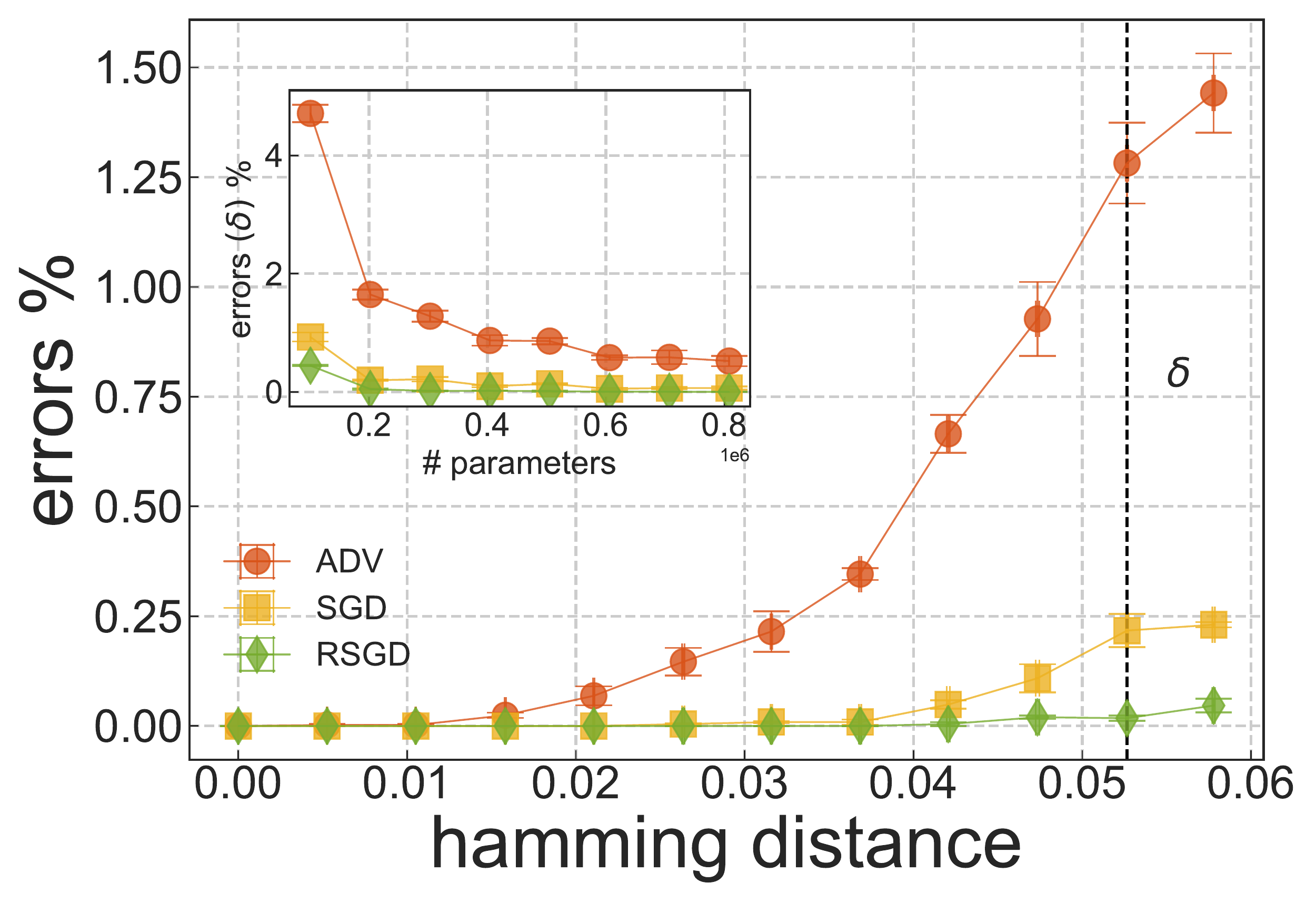} 
    \includegraphics[width=0.45\linewidth]{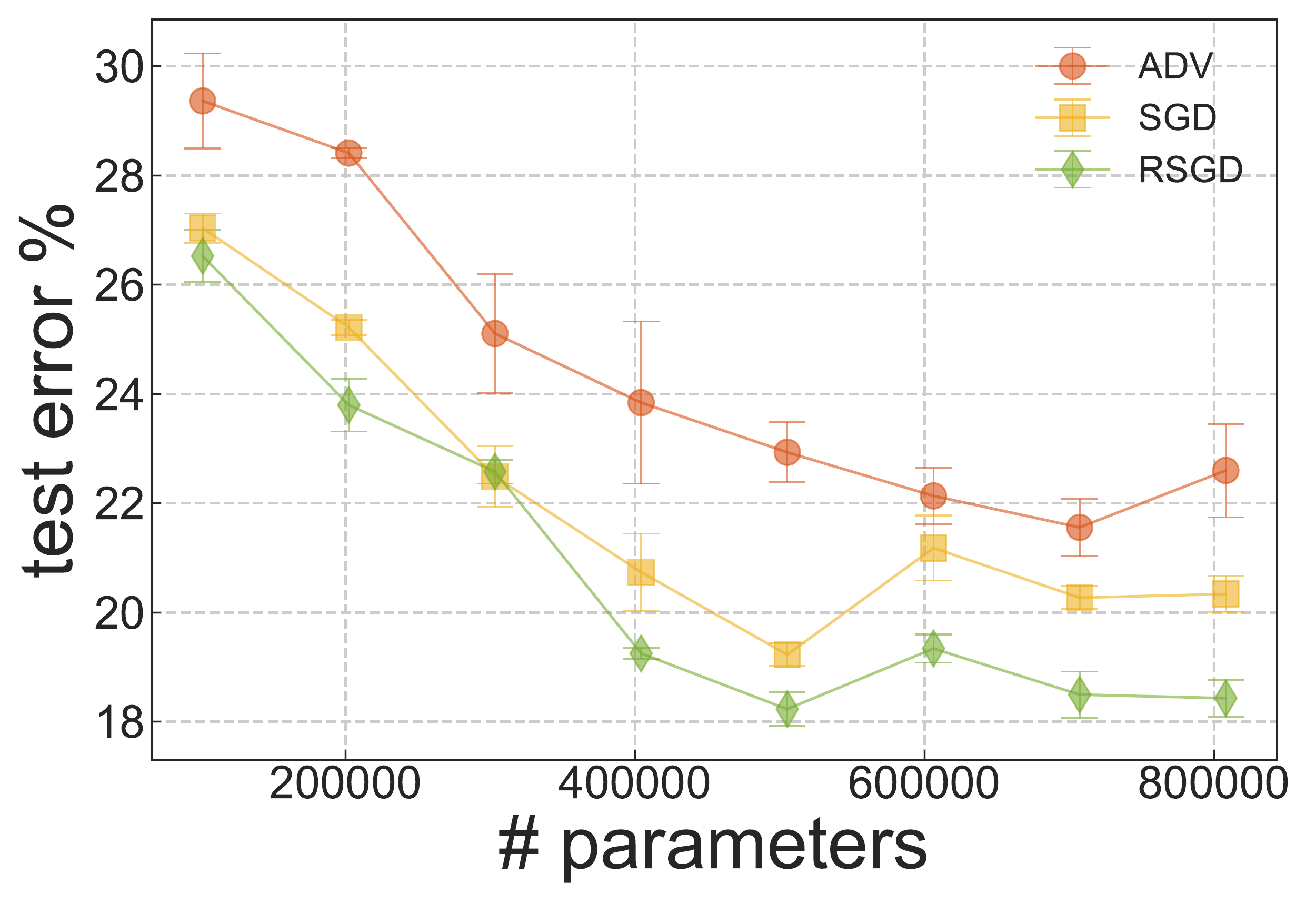} \\
    \includegraphics[width=0.45\linewidth]{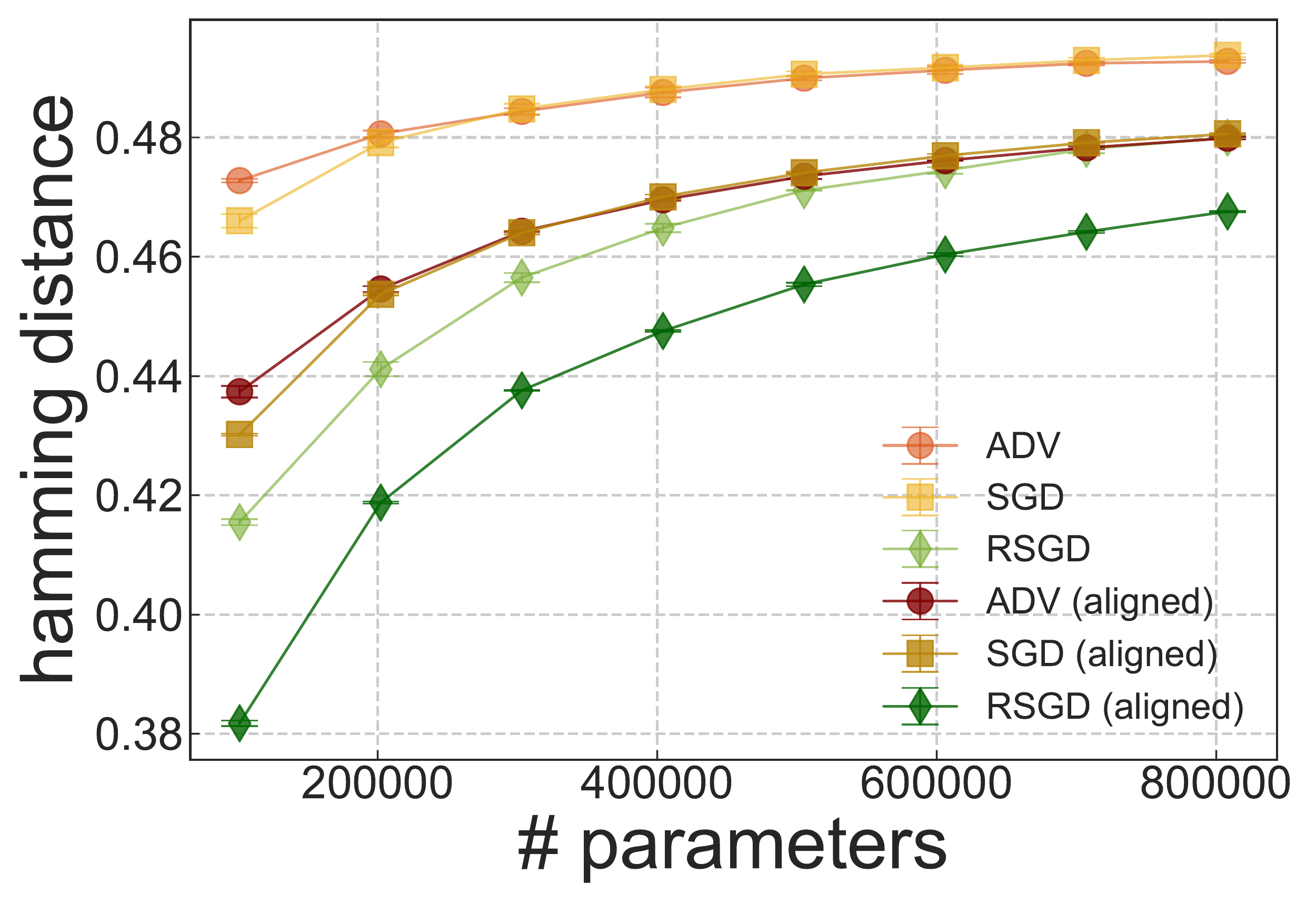}
    \includegraphics[width=0.45\linewidth]{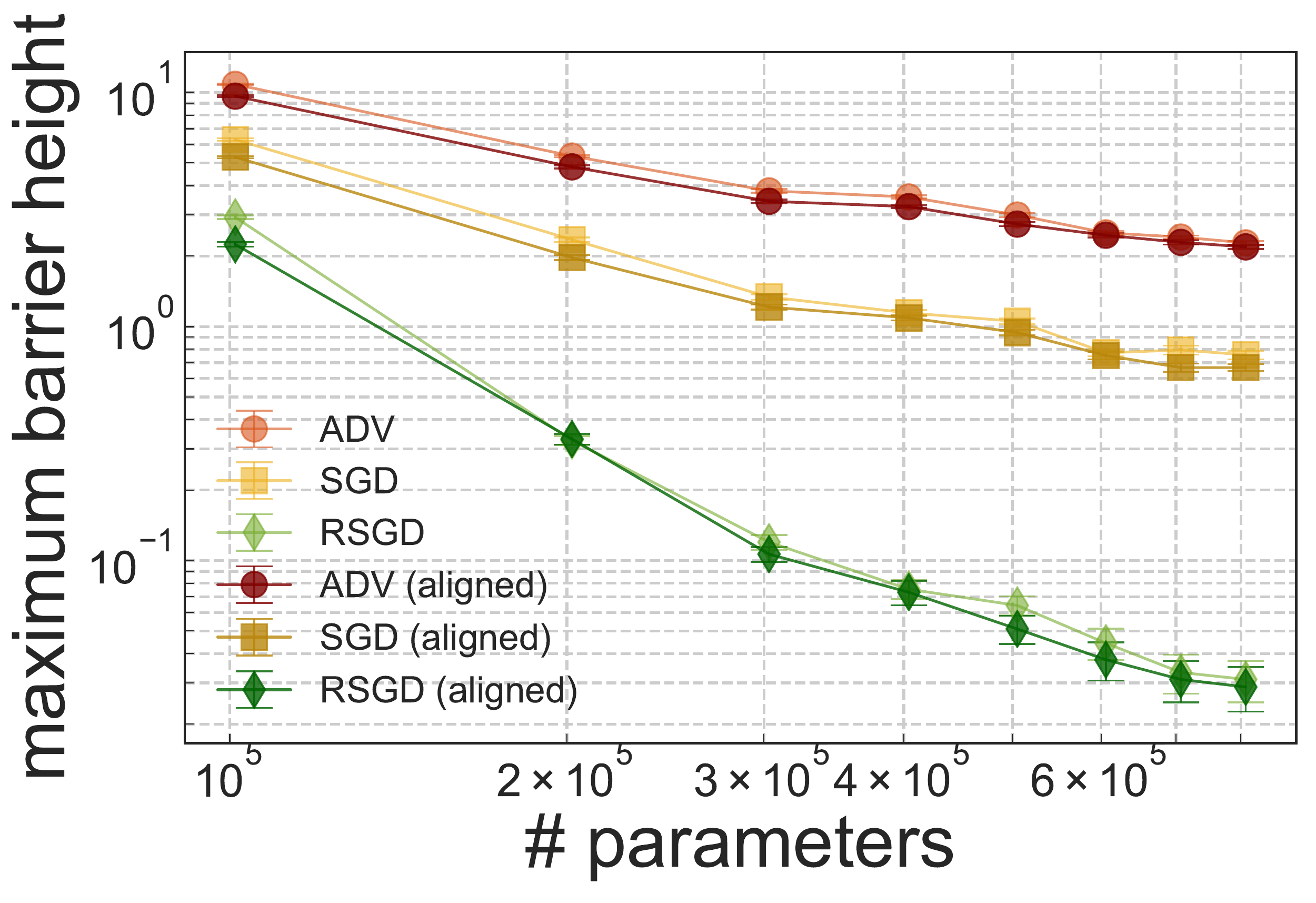}
    \caption{Fully connected binary CM trained on HMM data (see text). Top Left: Local energies for different classes of solutions. Inset: local energy at a fixed distance from a solution ($\delta$ in the main plot) as a function of the number of parameters. Top right: Test errors decrease with overparameterization and correlate with local energies. Bottom left: Average hamming distances between different solutions, before and after removal of symmetries. All distances grow with overeparameterization. Bottom right: Maximum barrier height (train error percentage) along a random shortest path connecting two solutions. Barriers go to zero with overparameterization.}
    \label{fig:binary_cm_hmm}
    \end{center}
\end{figure}

In order to bridge the gap with the theory, we performed numerical experiments on the error landscape connectivity in shallow binary architectures trained on data generated by the so called Hidden Manifold Model (HMM), also known as Random Features Model \cite{goldt2020modeling,gerace2020generalisation,baldassi2021learning}.

The HMM is a mismatched student-teacher setting. The teacher generating the labels is a simple one-layer network, whose inputs are $D$-dimensional random i.i.d. patterns. The students does not see these original patterns, but a non-linear random projection onto an $N$-dimensional space. By varying the relative size $N/D$ of the projection, the degree of overparameterization can be controlled. This arrangement aims to provide an analytically tractable model that retains some relevant features of the most common real-world vision datasets~\cite{goldt2020modeling}.

We trained both a binary perceptron and a fully connected binary committee machine (CM), i.e. a network with a single hidden layer where the weights of the output layer are fixed to $1$. Using data from a HMM, we analyzed the error landscape around solutions of different flatness, at varying levels of overparameterization (all implementation details are reported in Appendix Sec.~\ref{SI:binaryNNs}).

In this regime, the analysis of~\citet{baldassi2021learning} suggests a scenario where algorithmically accessible solutions are arranged in a connected zero-error landscape, with flatter solutions surrounded by sharper ones. Overall, the numerical findings we report here are consistent with this scenario.

In Fig.~\ref{fig:binary_cm_hmm} we report the results for the binary CM (similar results were obtained for the binary perceptron, see Appendix Sec.~\ref{SI:binaryNNs}).
As the number of parameters is increased, the flatness of all the solutions increases (while the ranking RSGD-SGD-ADV is preserved), and the generalization error is correlated with the flatness, as expected. As more parameters are added and solutions become flatter, the average maximum error along random linear paths connecting two solutions decreases. 
We observe a robust correlation between barrier heights and flatness: the flatter the solution, the lower the barrier.
However, the barriers are not significantly affected by aligning the networks. 


\subsection{Deep Architectures on Real-World Datasets}

\begin{figure}[ht]
  \centering
    \includegraphics[width=0.45\linewidth]{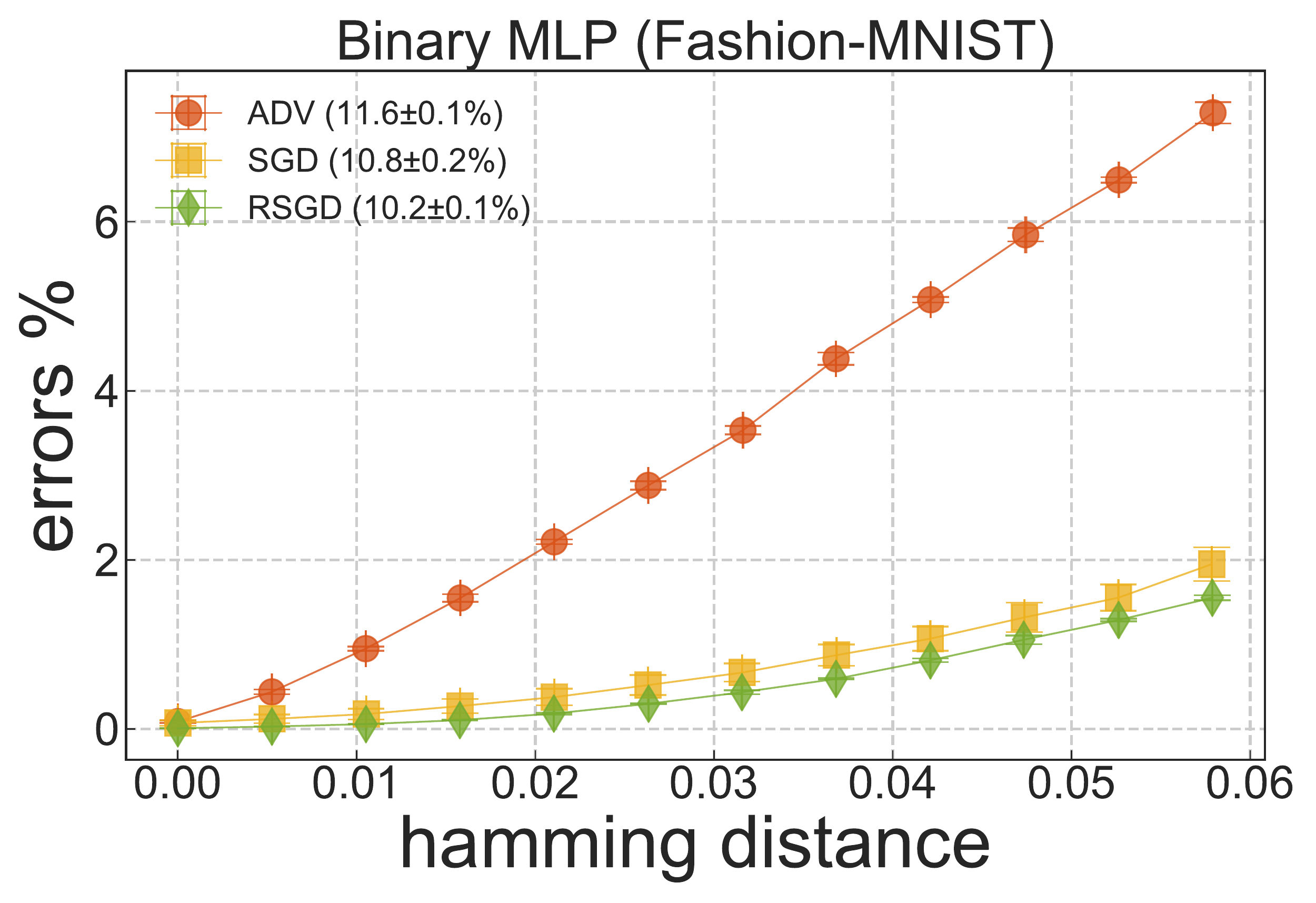}
    \includegraphics[width=0.45\linewidth]{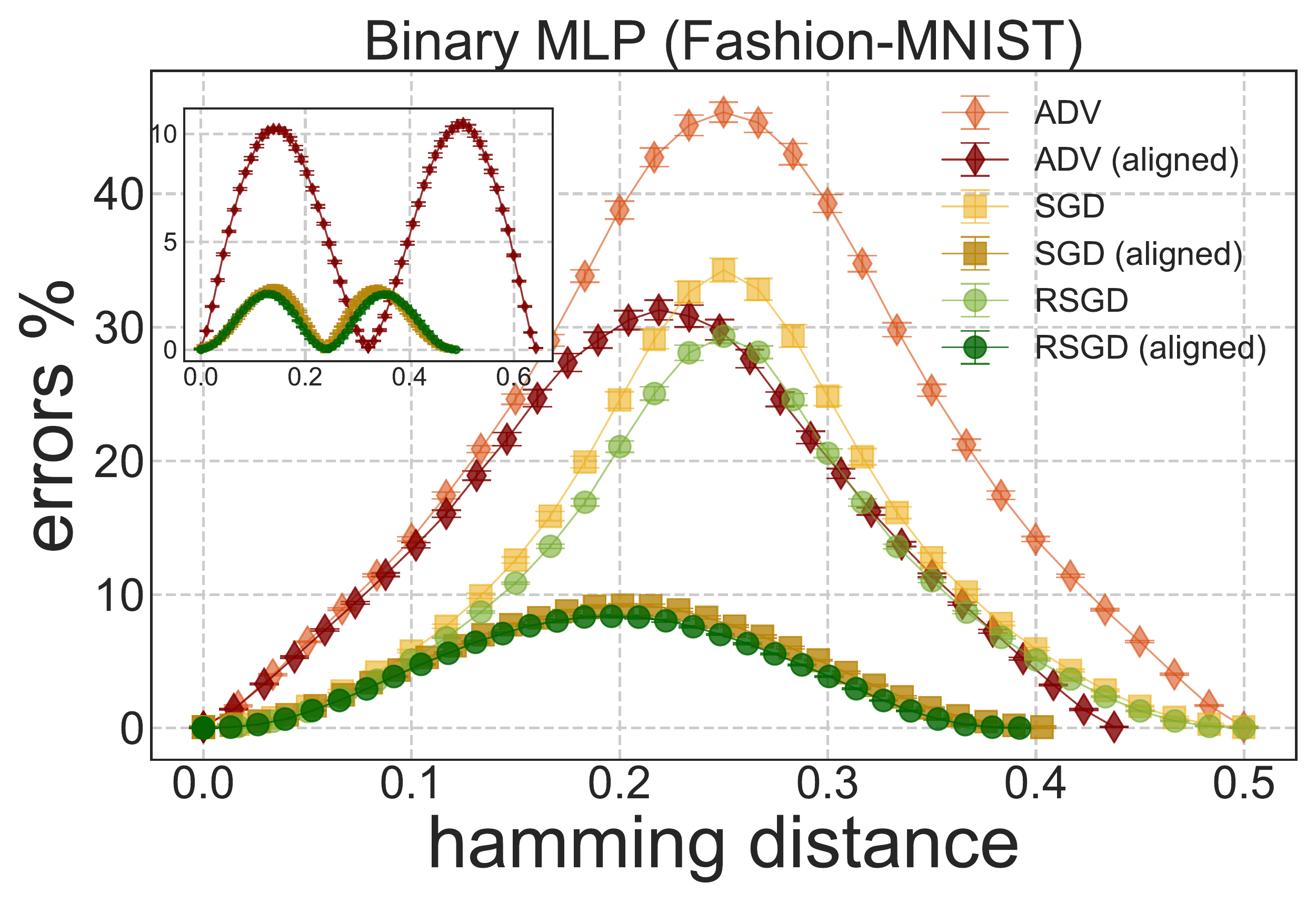} \\
    \includegraphics[width=0.45\linewidth]{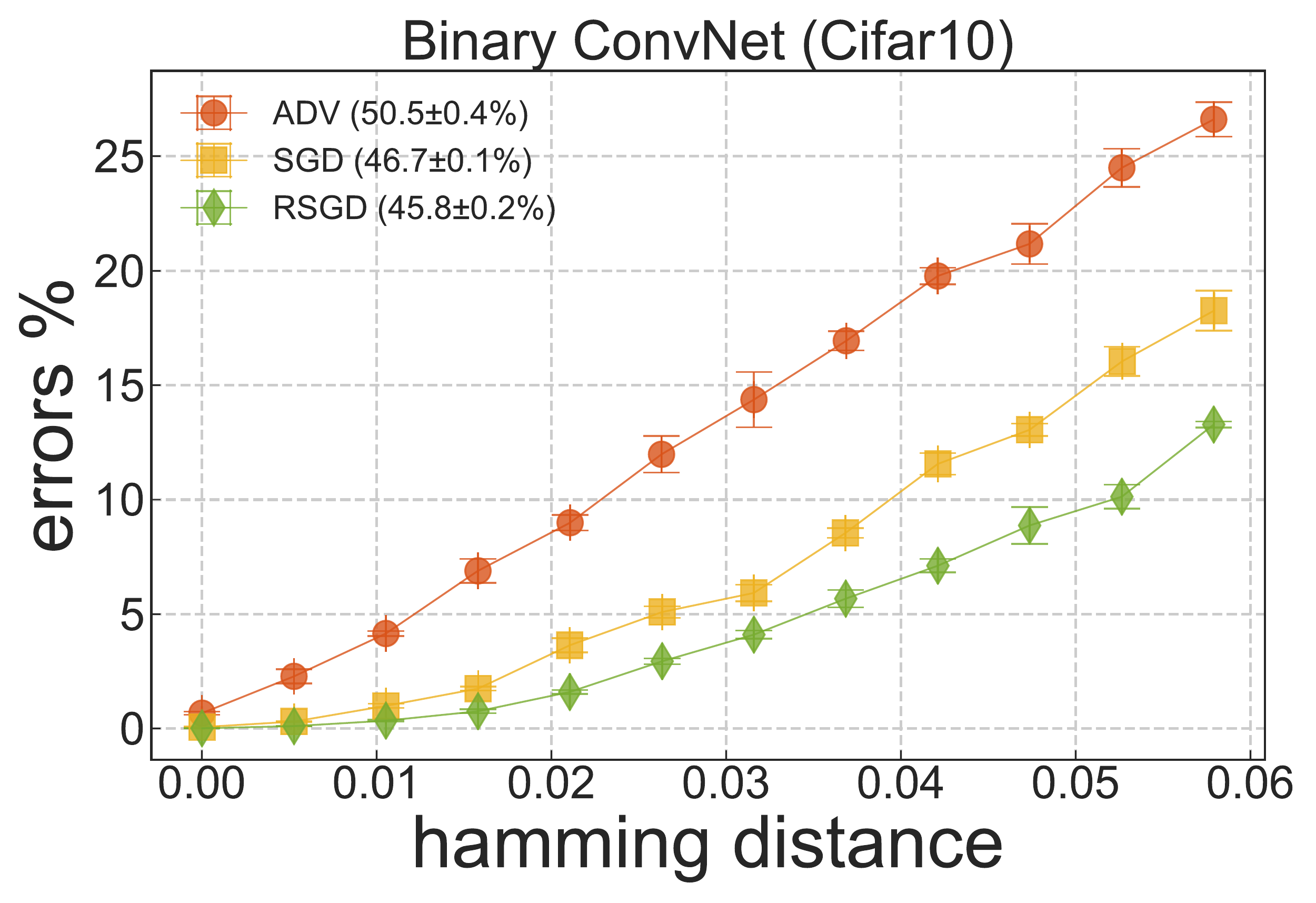}
    \includegraphics[width=0.45\linewidth]{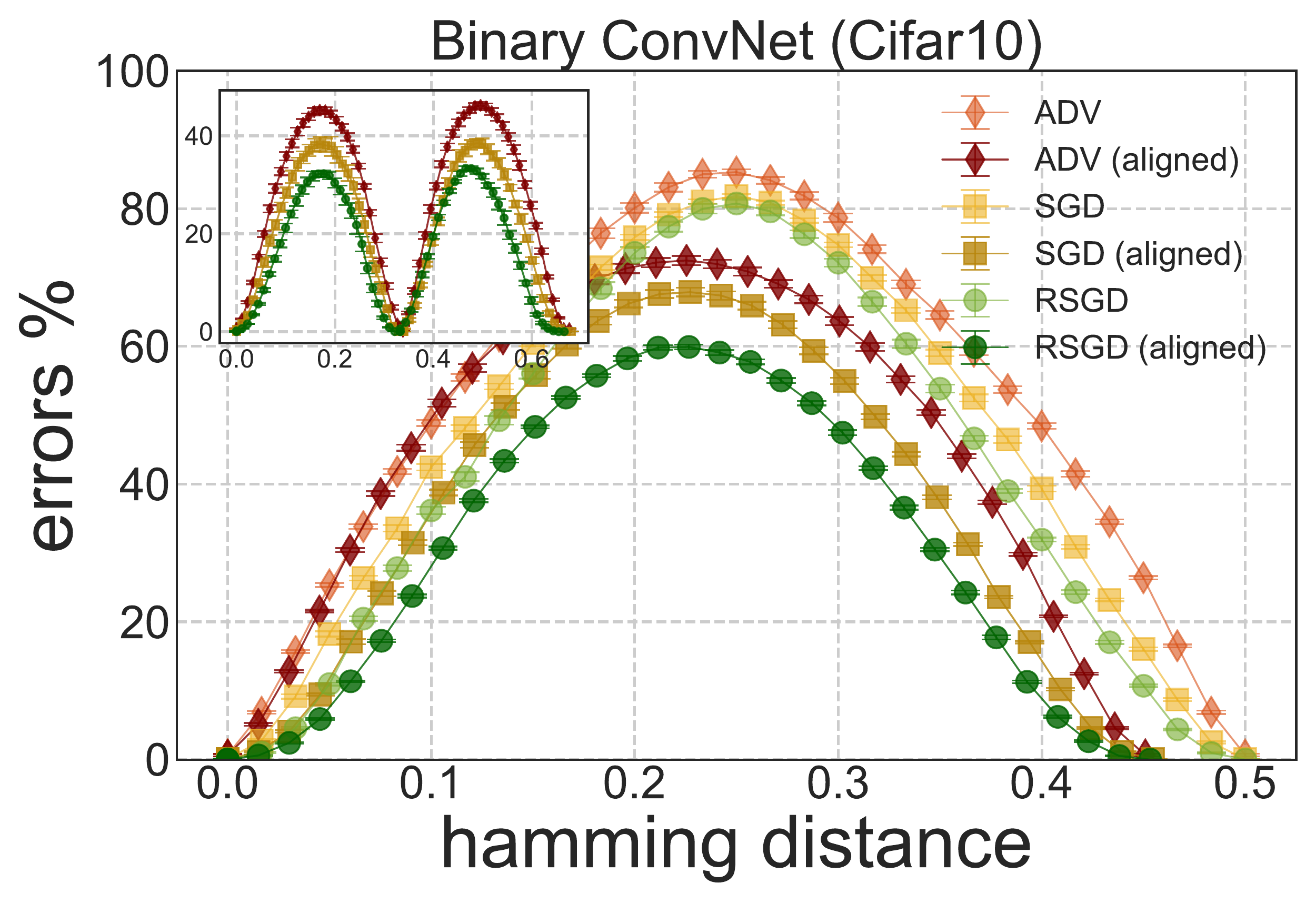}
    \caption{Three hidden layer binary MLP trained on the Fashion-MNIST dataset (top row) and binary convolutional NNs trained on CIFAR-10 (bottom row). Left: Local energies for different types of solutions (test errors are reported in the legend). Right: Train error percentage along random linear paths connecting solutions, for raw solutions (light curves) and aligned solutions (darker curves). Insets: errors along the optimized paths. The change in distance due to symmetries removal may be appreciated in the $x$-axis.}
  \label{fig:binary_deep_paths}
\end{figure}

\begin{figure}[ht]
  \centering
    \includegraphics[width=1.0\linewidth]{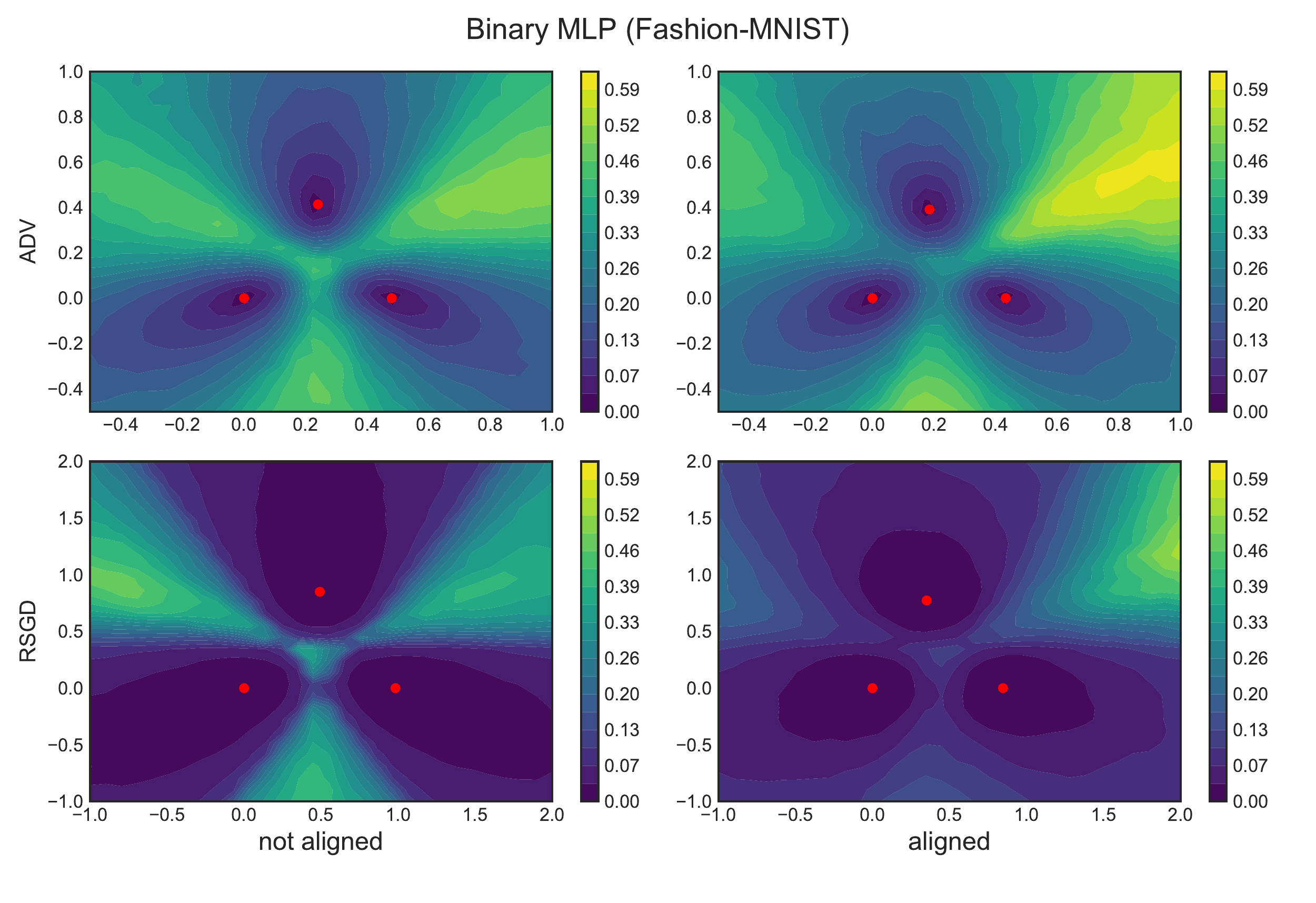}
  \caption{Train errors in the plane spanned by three solutions for a binary MLP trained on Fashion-MNIST, before (left column) and after (right column) symmetries removal. We compare ADV (top row) and RSGD (bottom row) solutions.}
  \label{fig:binarymlp_fashion_plane}
\end{figure}

We consider two deep binary NNs: a MLP with $3$-hidden-layers of $1001$ units each, trained on the Fashion-MNIST dataset, and a $5$-layers convolutional NN trained on CIFAR10 (implementation details in Appendix Sec.~\ref{SI:binaryNNs}).

For both architectures we analyzed the error landscape along random shortest paths connecting pairs of solutions with different flatness.
Results are reported in Fig.~\ref{fig:binary_deep_paths}. As expected, the average barrier height correlates with the local energies of the solutions (and in turn with test errors).
The barriers are lowered when symmetries are removed, and are lowered further when the paths are optimized (although they remain considerably large, especially for the convolutional NNs).

The effect of removing symmetries on the error landscape connectivity can be appreciated in Fig.~\ref{fig:binarymlp_fashion_plane} 
(see also Appendix Fig.~\ref{fig:binaryconv_cifar_plane} for analogous results on the convolutional network), where we projected the error landscape onto the plane spanned by the different solutions. We used the internal continuous weights used by BinaryNet (of which the actual weights are just the $\mathrm{sign}$) 
and proceeded as in the continuous case (Sec.~\ref{par:continuous_2dplots}), but
at each point we binarized the configurations in order to obtain the errors (details in Appendix Sec.~\ref{SI:binaryNNs}).
The resulting projections are a heavily distorted representation, but the effect of symmetry removal on the barriers is rather striking, especially in the case of the wider minima, revealing the presence of a connected structure.

\section{Conclusions and discussion}

We investigated numerically several features of the error landscape in a wide variety of neural networks, building on a systematic geometric approach that dictates the removal of all the symmetries in the represented functions. The methods we developed are approximate but simple, rather general and efficient, and proved to be critically important to our findings. By sampling different kinds of minima, investigating the landscape around them and the paths connecting them, we found a number of fairly robust features shared by all models. In particular, besides confirming the known connection between wide minima and generalization, our results support the conjecture of~\citet{baldassi2021learning}: that, for sufficiently overparameterized networks, wide regions of solutions are organized in a complex, octopus-shaped structure with the denser regions clustered at the center, from which progressively sharper minima branch out. Intriguingly, a similar phenomenon has been recently observed also in the—rather different—context of systems of coupled oscillators \cite{tentacles}.

Our work lies at the intersection of two lines of research that have seen significant interest lately: one on mode connectivity and the structure of neural network landscapes and the other on flat minima and their connection to generalization performance. 
In this context, our work contributes to a deeper understanding of how deep neural networks operate. This may have algorithmic implications, for example previous investigations led to enabling better ensembling schemes (see, e.g. \citet{garipov2018}), hyperparameter choices (see, e.g. \citet{foret2021sharpnessaware}) and methods for improving generalization \cite{pittorino2021}, and has the potential to also help architecture design. Work is in progress to investigate the algorithmic implications of our results.
We believe that further systematic explorations of the topics treated in this paper can produce results of great theoretical interest and significant algorithmic implications.



\section*{Acknowledgements}
FP acknowledges the European Research Council for Grant No. 834861 SO-ReCoDi.



\bibliography{references}
\bibliographystyle{icml2022}

\newpage
\appendix
\onecolumn

\section{\label{SI:contw}Neural Networks with Continuous Weights}
We report complete and additional results on the Neural Networks (NNs) with continuous weights studied in the main paper.

\subsection{Numerical details and parameters}
The training parameters for the $3$ algorithms for all the networks/datasets tested in the main paper are: (SGD) SGD with Nesterov momentum and initial learning rate $0.02$ with cosine annealing ($0.002$ for MLP on CIFAR-10); (RSGD) Replicated-SGD with Nesterov momentum and initial learning rate $0.05$ with cosine annealing, $y=5$ replicas, with an exponential schedule on the interaction parameter $\gamma_t=\gamma_0(1+\gamma_1)^t$ with $\gamma_0$ and $\gamma_1$ automatically chosen (see \citet{pittorino2021} for details on this algorithm); (ADV) for the configurations obtained with the adversarial initialization, we use vanilla SGD and initial learning rate $0.02$ with cosine annealing (see \citet{bad_minima_sgd} for details on the generation of this initialization: we replicate one time the dataset with random labels, i.e. $R=1$, and we zero-out a $10\%$ random fraction of the pixels in each image). We train all networks with batch-size $128$ and for $300$ epochs in order to reach training errors $<1\%$.
Neither data augmentation nor $\ell_2$ regularization are used in our experiments.

\subsection{\label{SI:1dpaths}One-Dimensional Paths}

\begin{figure}[H]
    \begin{center}
    \includegraphics[width=0.5\linewidth]{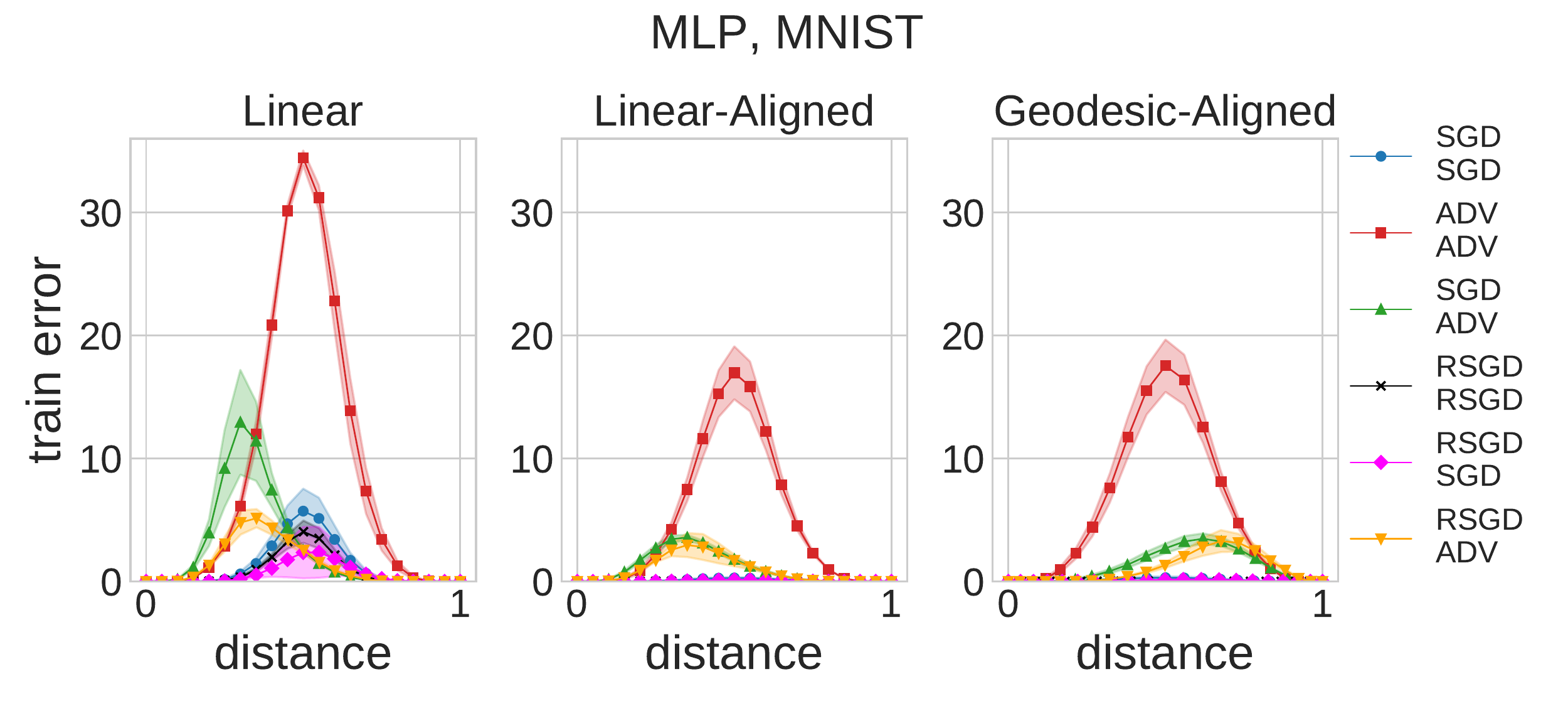}\includegraphics[width=0.5\linewidth]{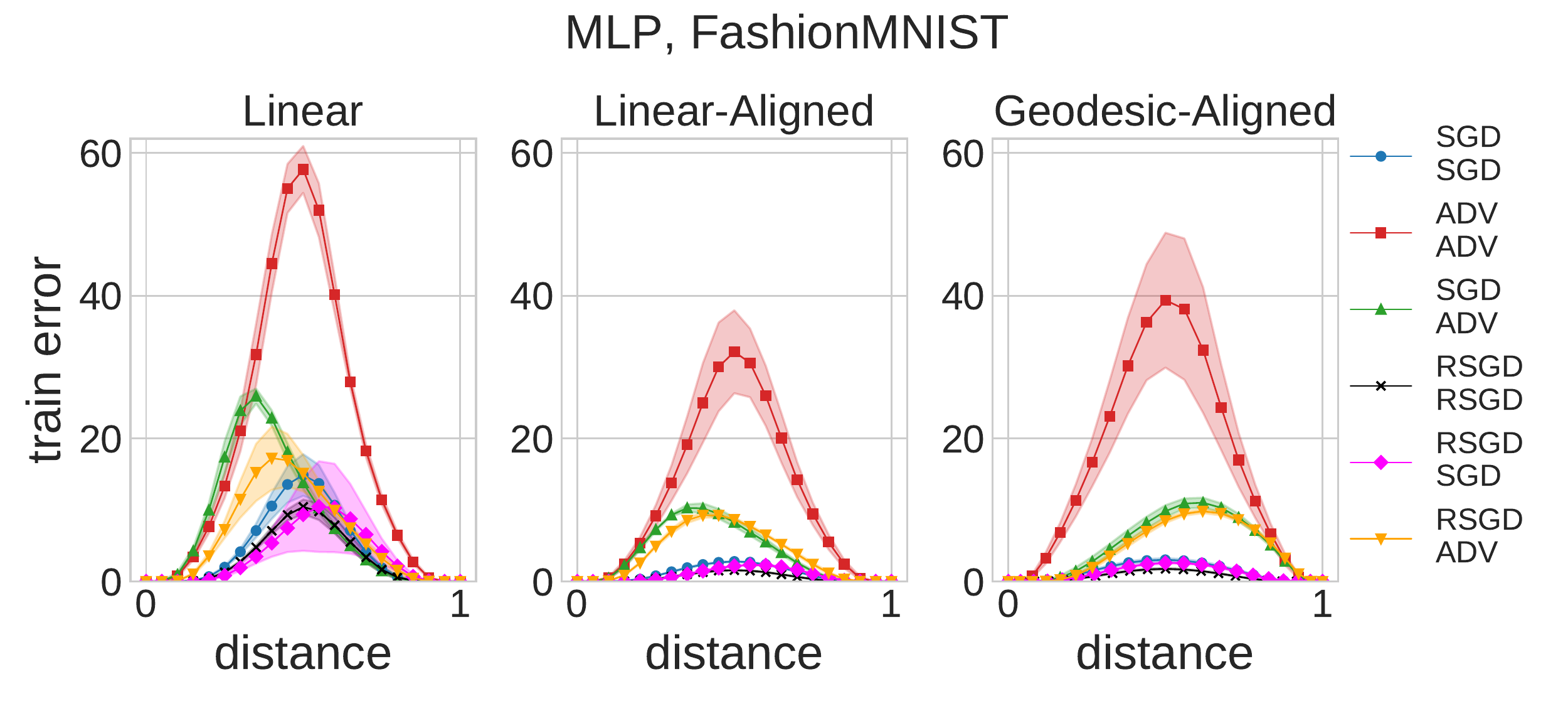}
    \includegraphics[width=0.5\linewidth]{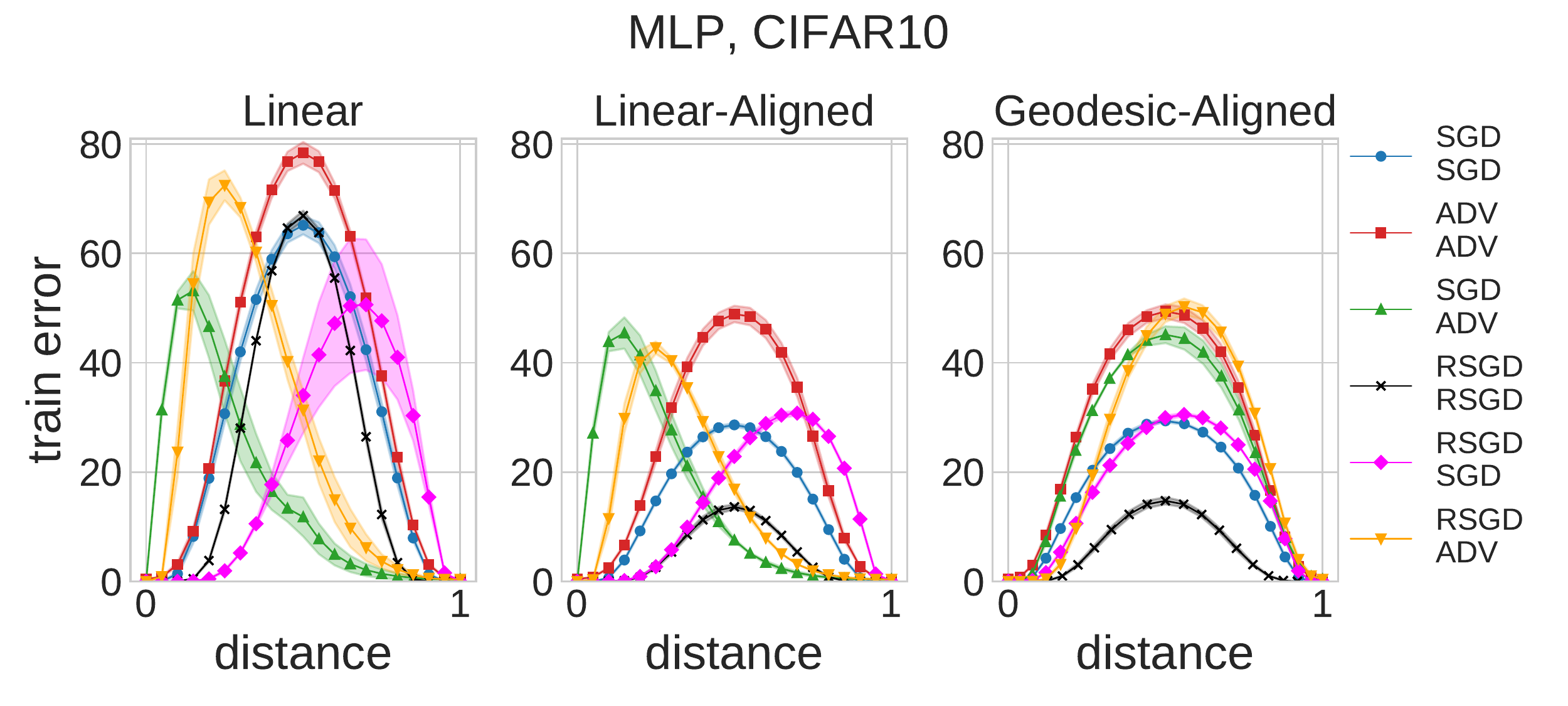}\includegraphics[width=0.5\linewidth]{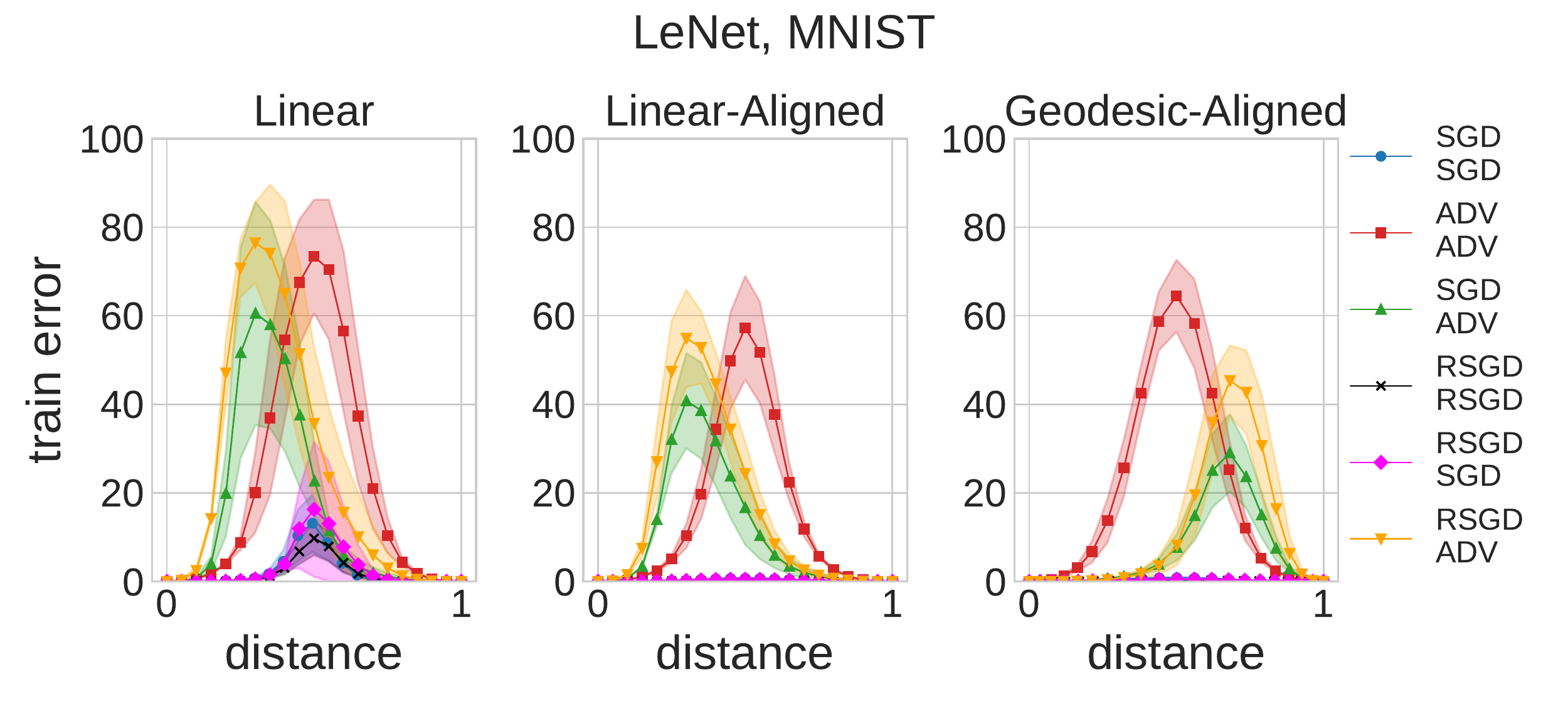}
    \includegraphics[width=0.5\linewidth]{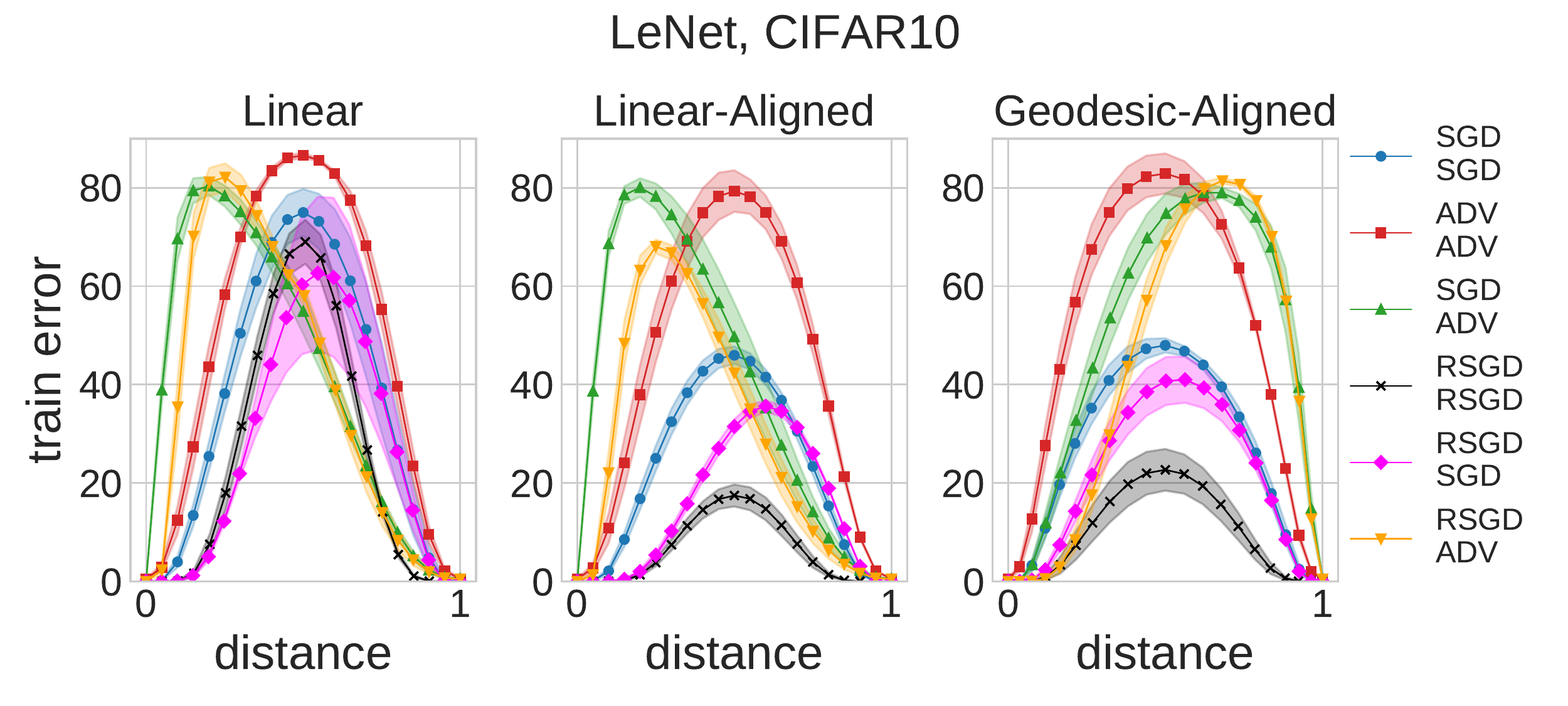}\includegraphics[width=0.5\linewidth]{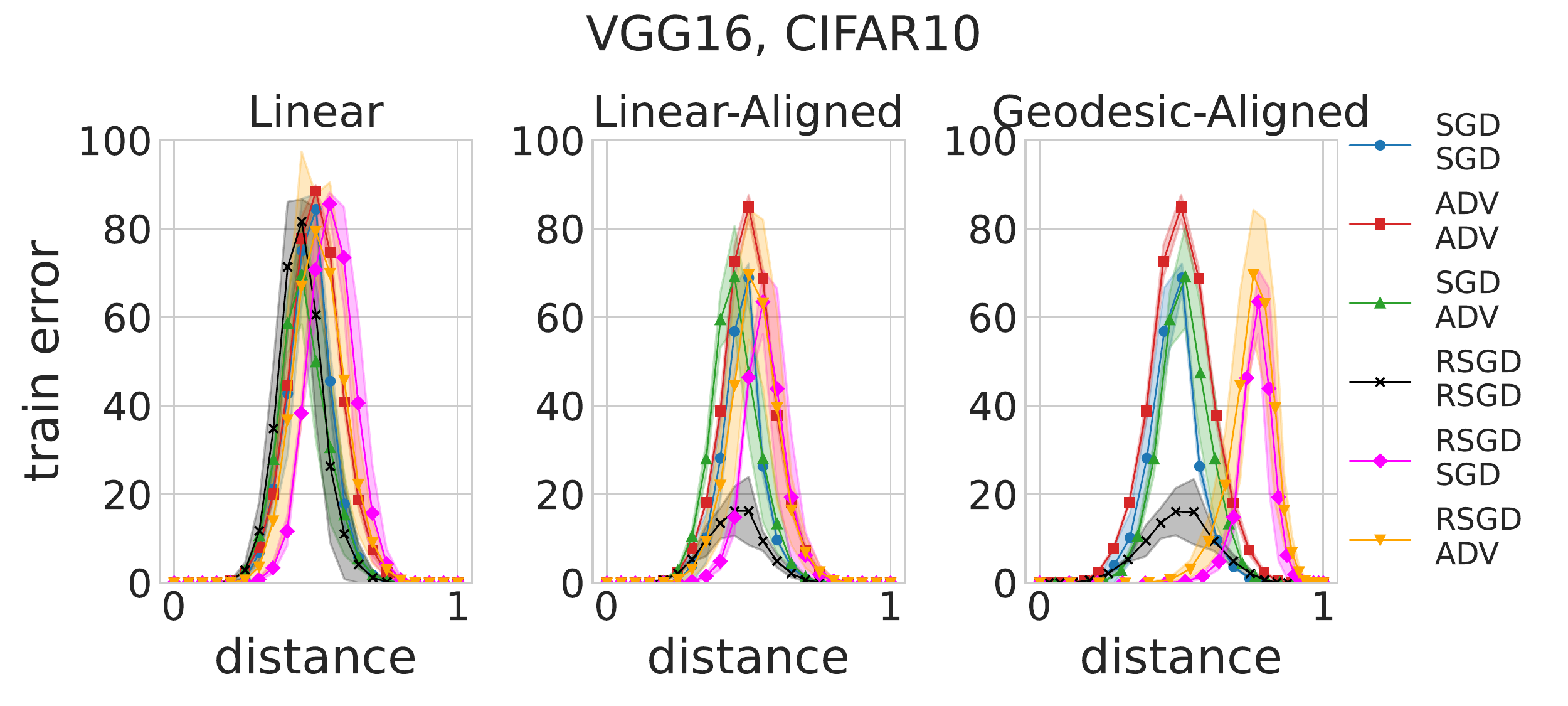}
    \caption{Error landscape along one-dimensional paths connecting minima of different flatness (minima appear on left/right following the up/down order of the legend). For each network/dataset the comparison highlights the general effect of the symmetries. The geodesic-aligned paths represents the final result of our analysis.}
    \label{fig:linearSI}
    \end{center}
\end{figure}

\begin{figure}[H]
    \vskip 0.2in
    \begin{center}
    \includegraphics[width=0.5\linewidth]{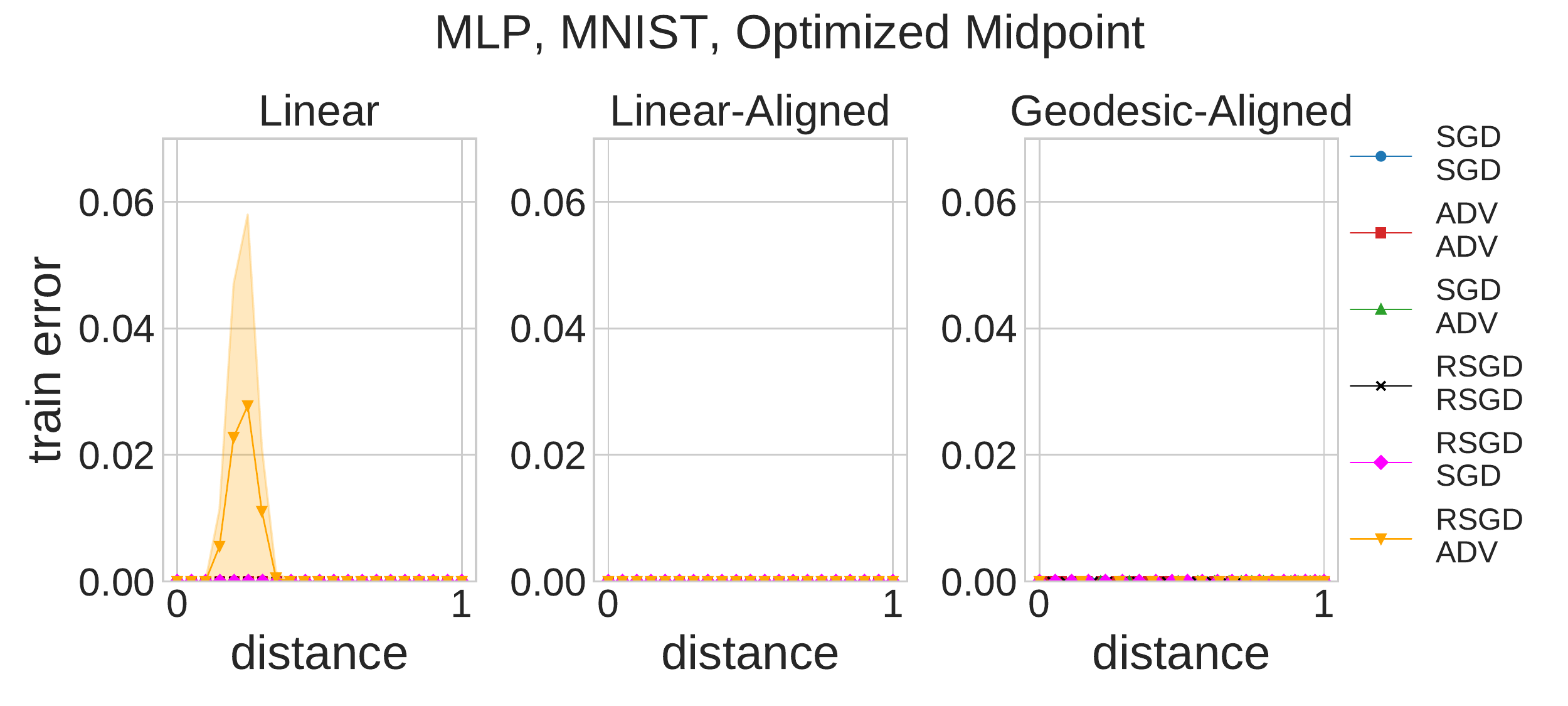}\includegraphics[width=0.5\linewidth]{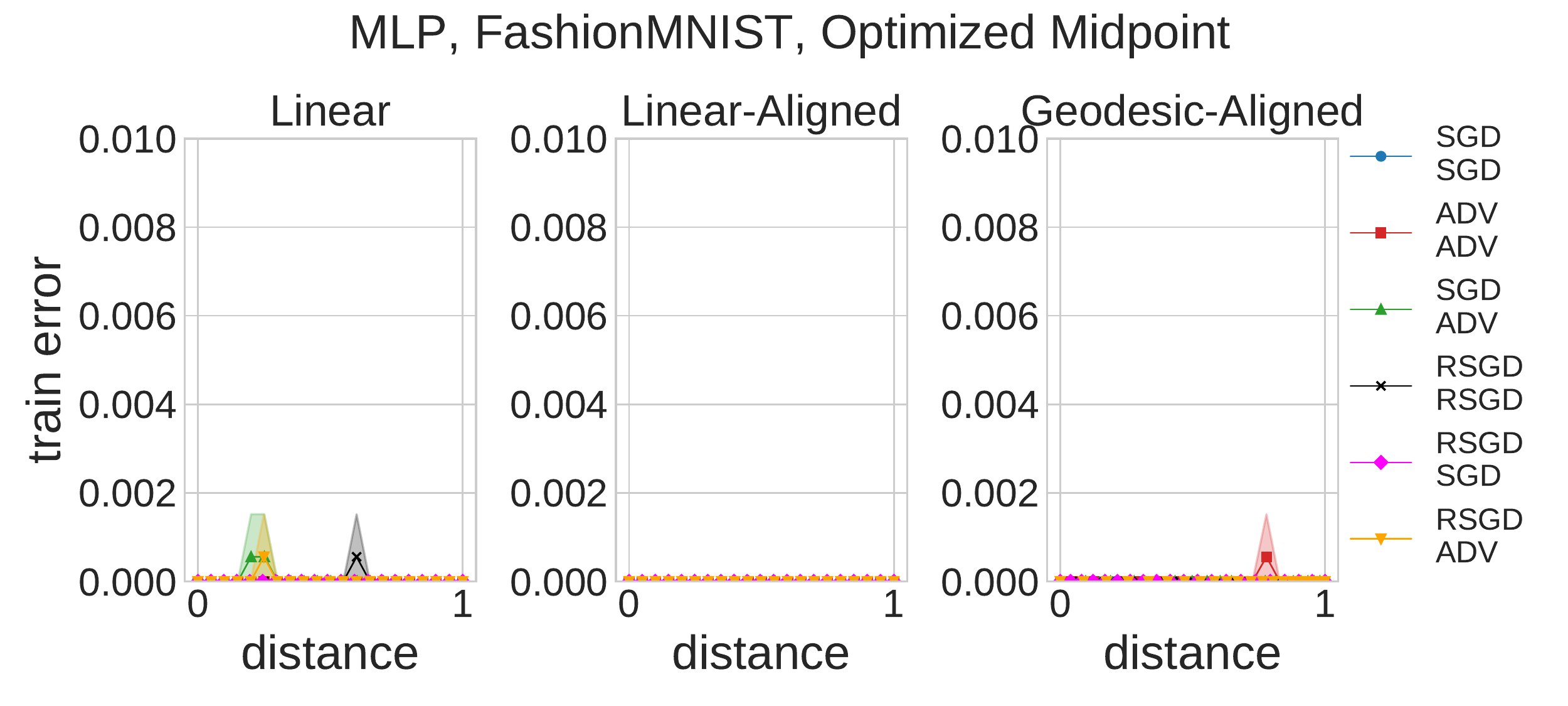}
    \includegraphics[width=0.5\linewidth]{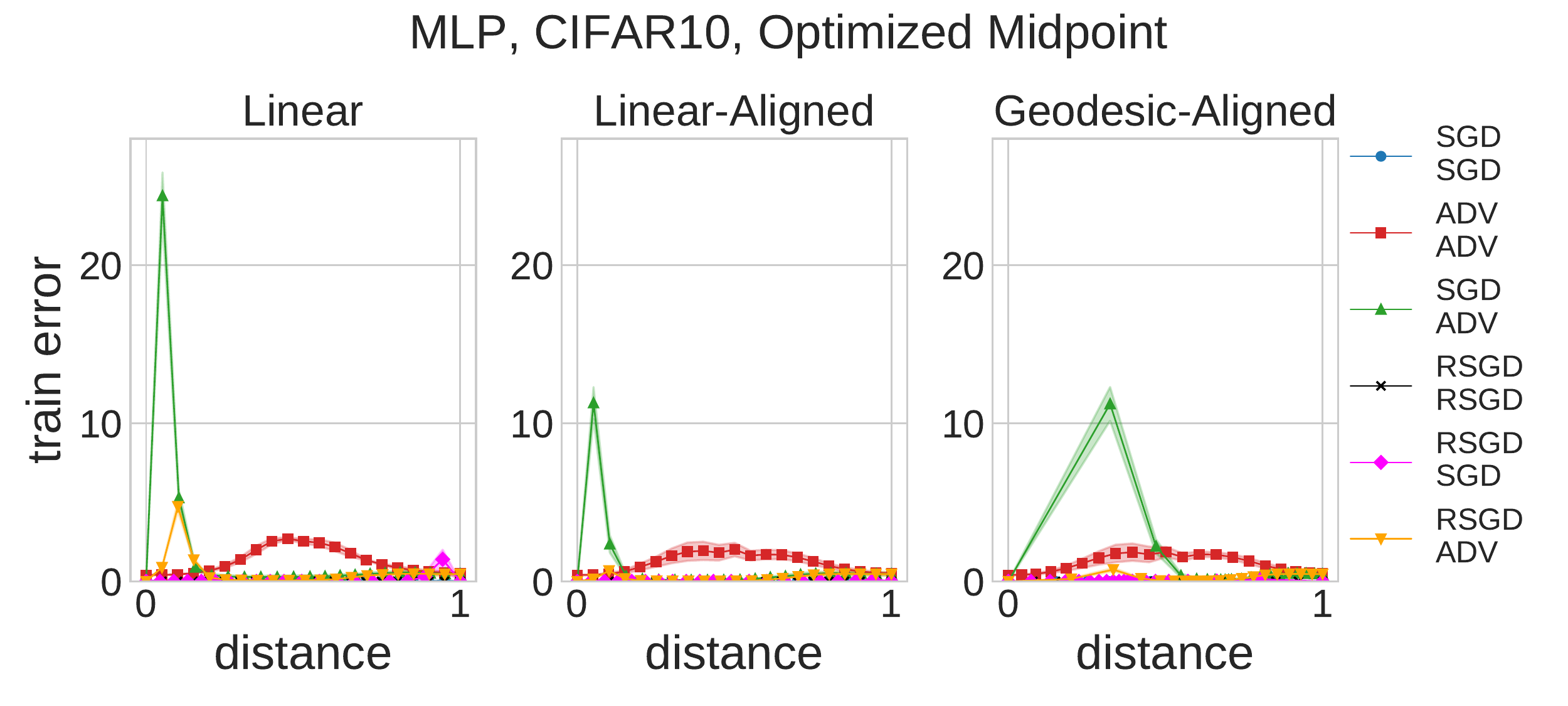}\includegraphics[width=0.5\linewidth]{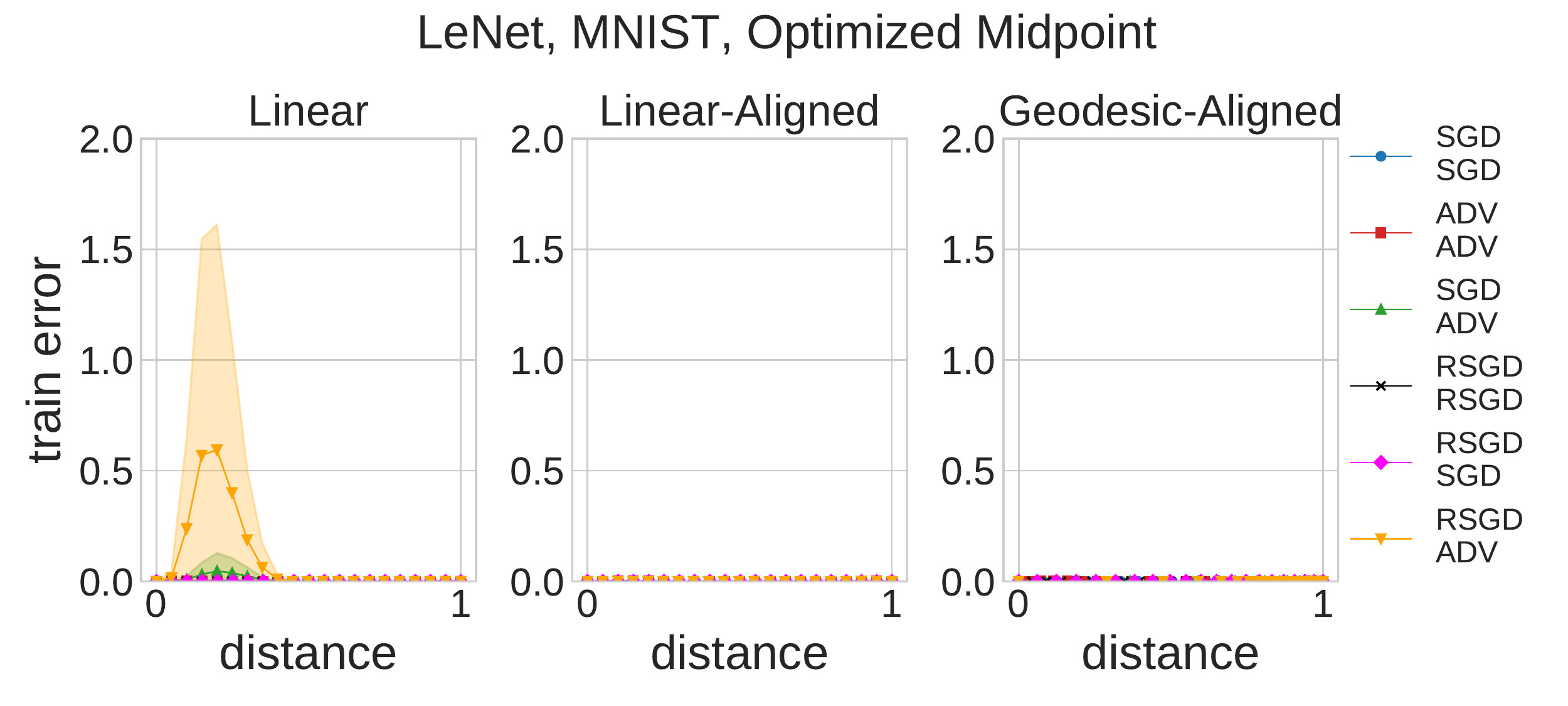}
    \includegraphics[width=0.5\linewidth]{good_figures/paths1d/confonti/curves1d_epoch600_testlenet_fashion-mnist_confronto.pdf}\includegraphics[width=0.5\linewidth]{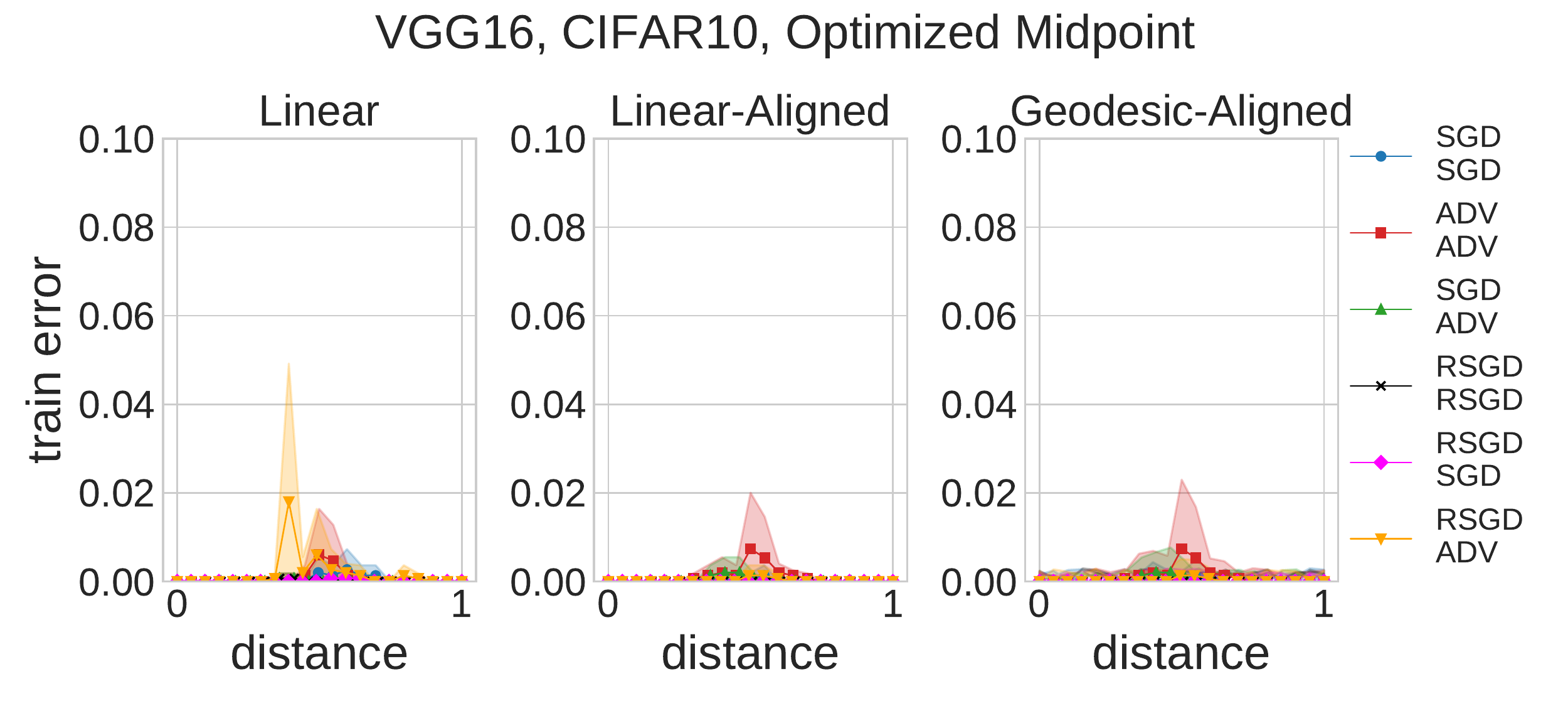}
    \caption{Removing the permutation symmetry eliminates the barriers that may appear in the single-bend optimized paths. For each network/dataset, left to right: linear path; linear-aligned path; geodesic-aligned path. For these optimized paths with one bend, the midpoint is optimized with batch-size $128$ for $300$ epochs with Nesterov momentum and initial learning rate $0.02$ with cosine annealing in order to reach training error equal or almost equal to zero.}
    \label{fig:optSI}
    \end{center}
    \vskip -0.2in
\end{figure}

We add here further results on the comparison among Linear, Linear-Aligned and Geodetic-Aligned one-dimensional paths on the networks/dataset with continuous weights studied in the main paper. We report in Fig.~\ref{fig:linearSI} the paths without optimizing the midpoint, while in Fig.~\ref{fig:optSI} the results obtained by optimizing it (the single-bend optimized paths).
\vspace{10cm}

\subsection{\label{SI:bidim}Bi-Dimensional Visualization}

\begin{figure}[H]
    \vskip 0.2in
    \begin{center}
    \includegraphics[width=0.25\linewidth]{good_figures/paths2d/train_loss_lin.pdf}\includegraphics[width=0.25\linewidth]{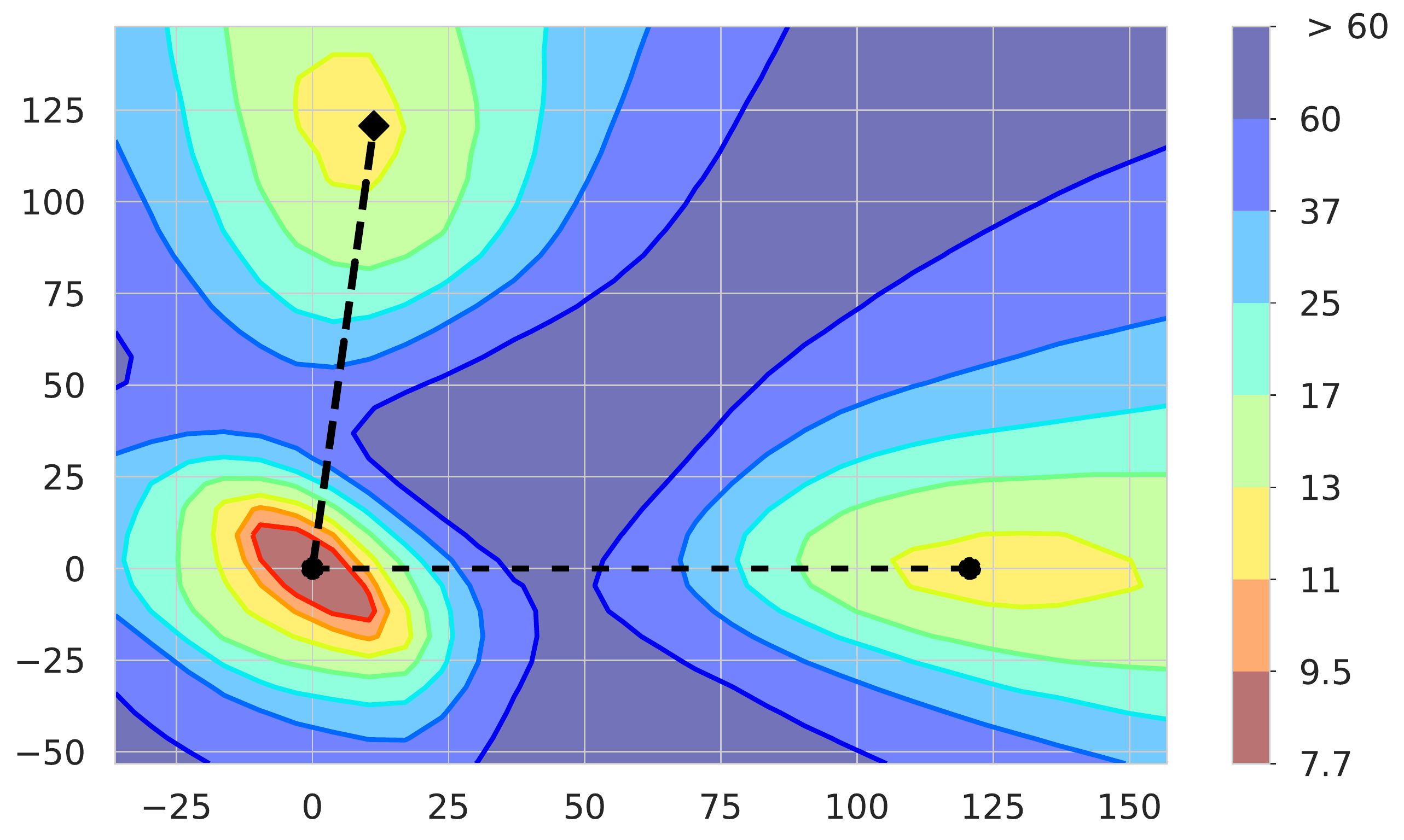}\includegraphics[width=0.25\linewidth]{good_figures/paths2d/train_loss_lin_norm.pdf}\includegraphics[width=0.25\linewidth]{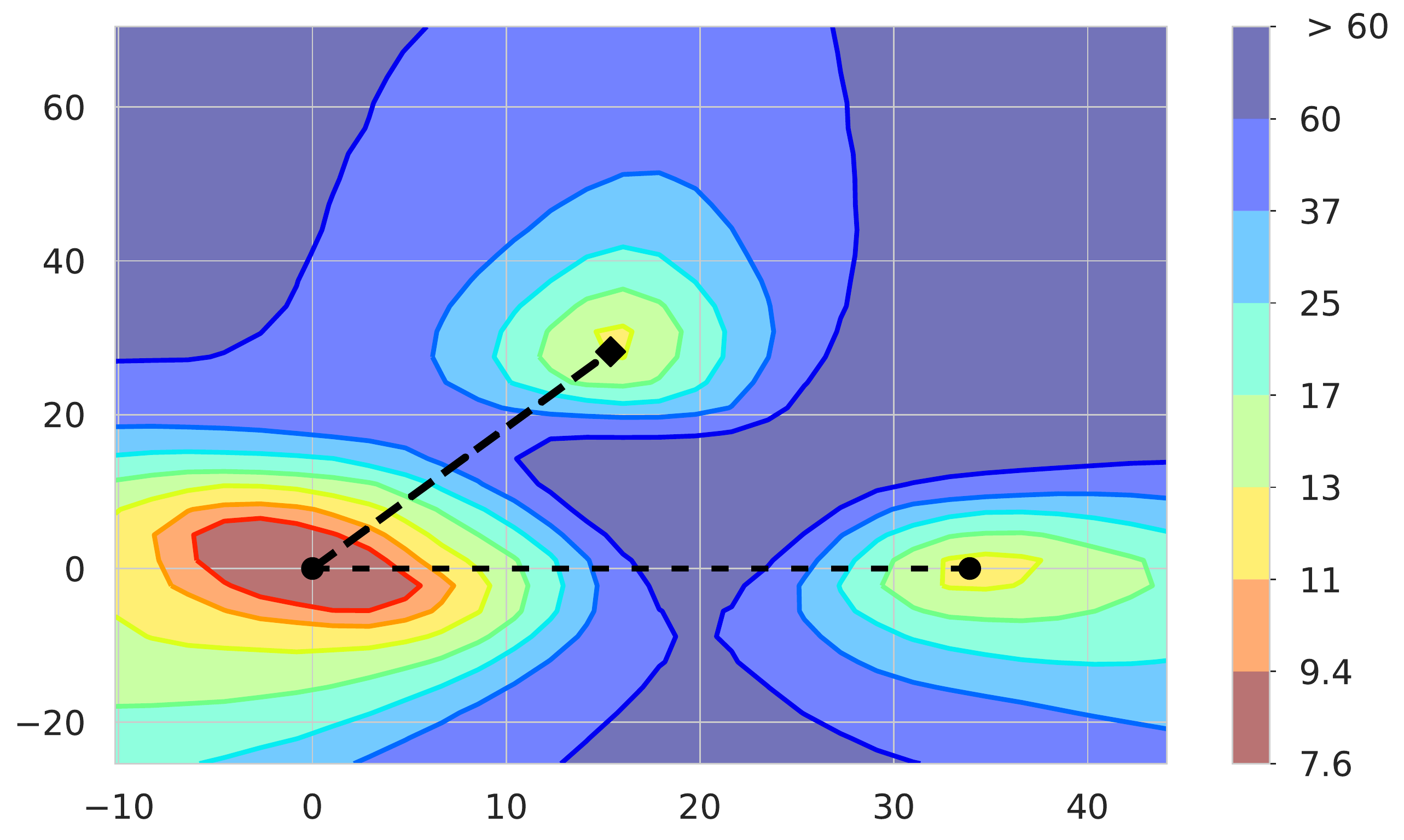}
    \caption{Bi-dimensional sections, LeNet on Fashion-MNIST. Train errors (panels 1 and 3) and test errors (panels 2 and 4). Comparison of RSGD (left points), unaligned ADV (right points), aligned ADV (top points). Panels 1 and 2: without normalization. Panels 3 and 4: with normalization. Dashed lines represent linear (panels 1 and 2) and geodesic paths (panels 3 and 4).}
    \label{figSI:2dsurfaces-1}
    \end{center}
    \vskip -0.2in
\end{figure}

\begin{figure}[H]
    \vskip 0.2in
    \begin{center}
    \includegraphics[width=0.33\linewidth]{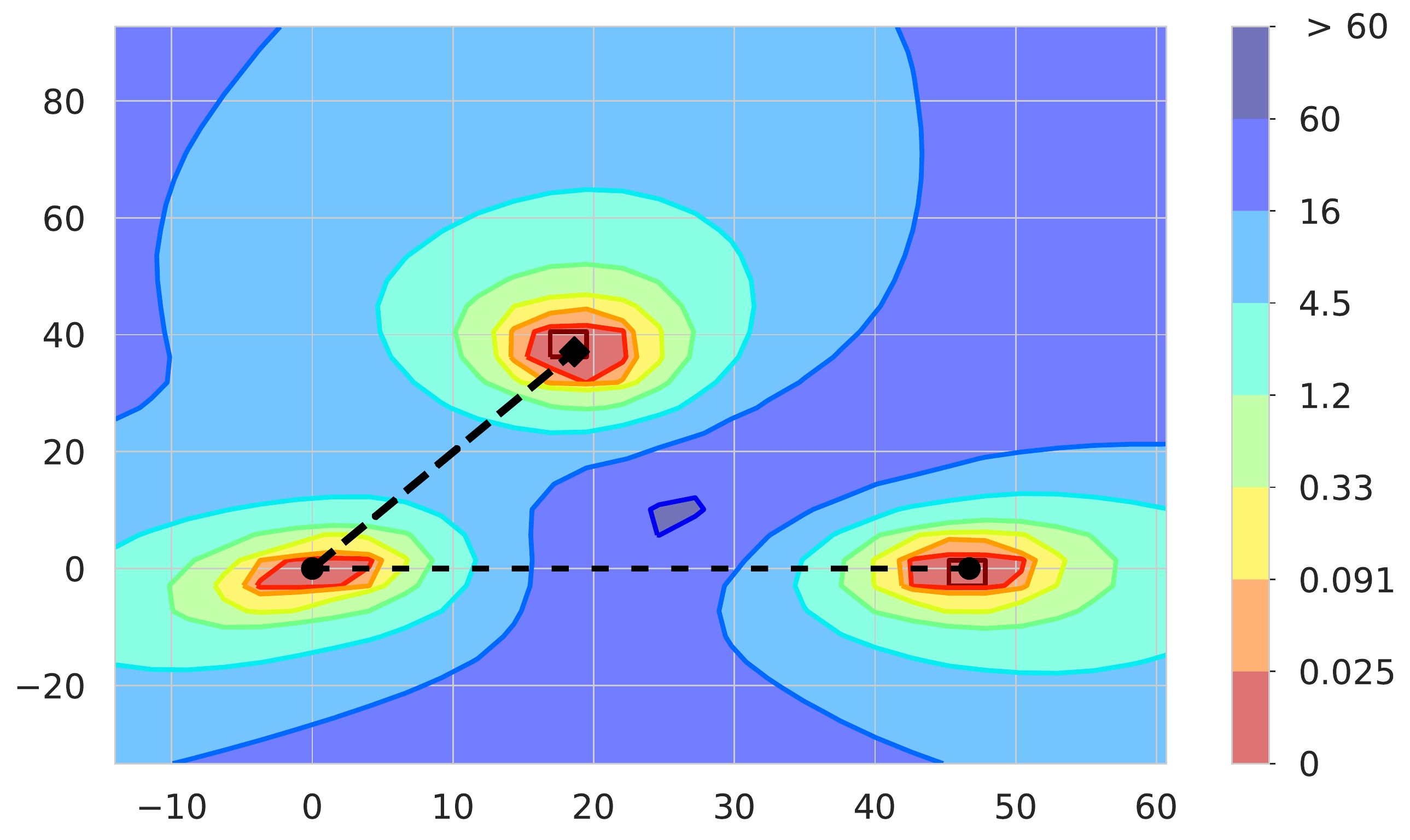}\includegraphics[width=0.33\linewidth]{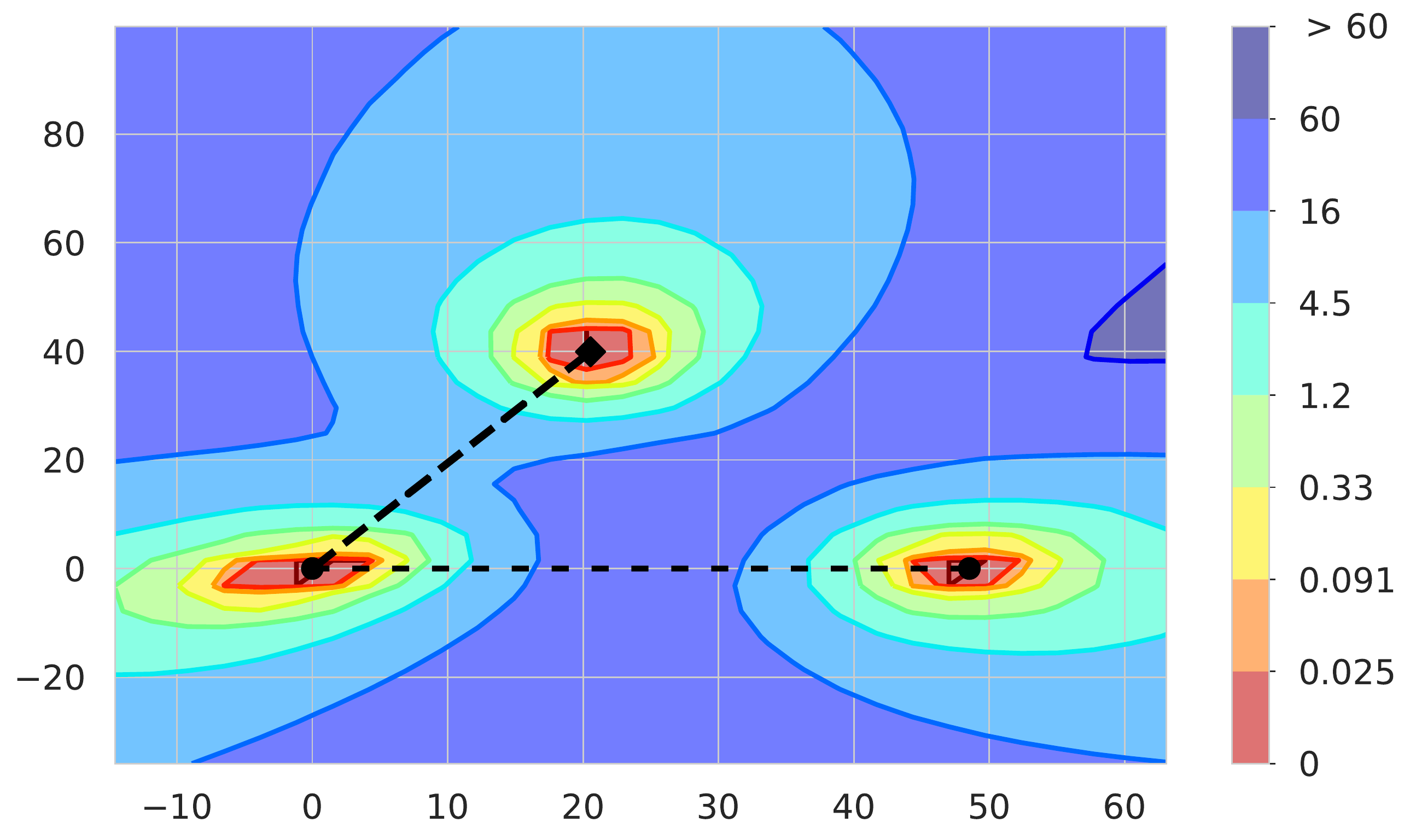}\includegraphics[width=0.33\linewidth]{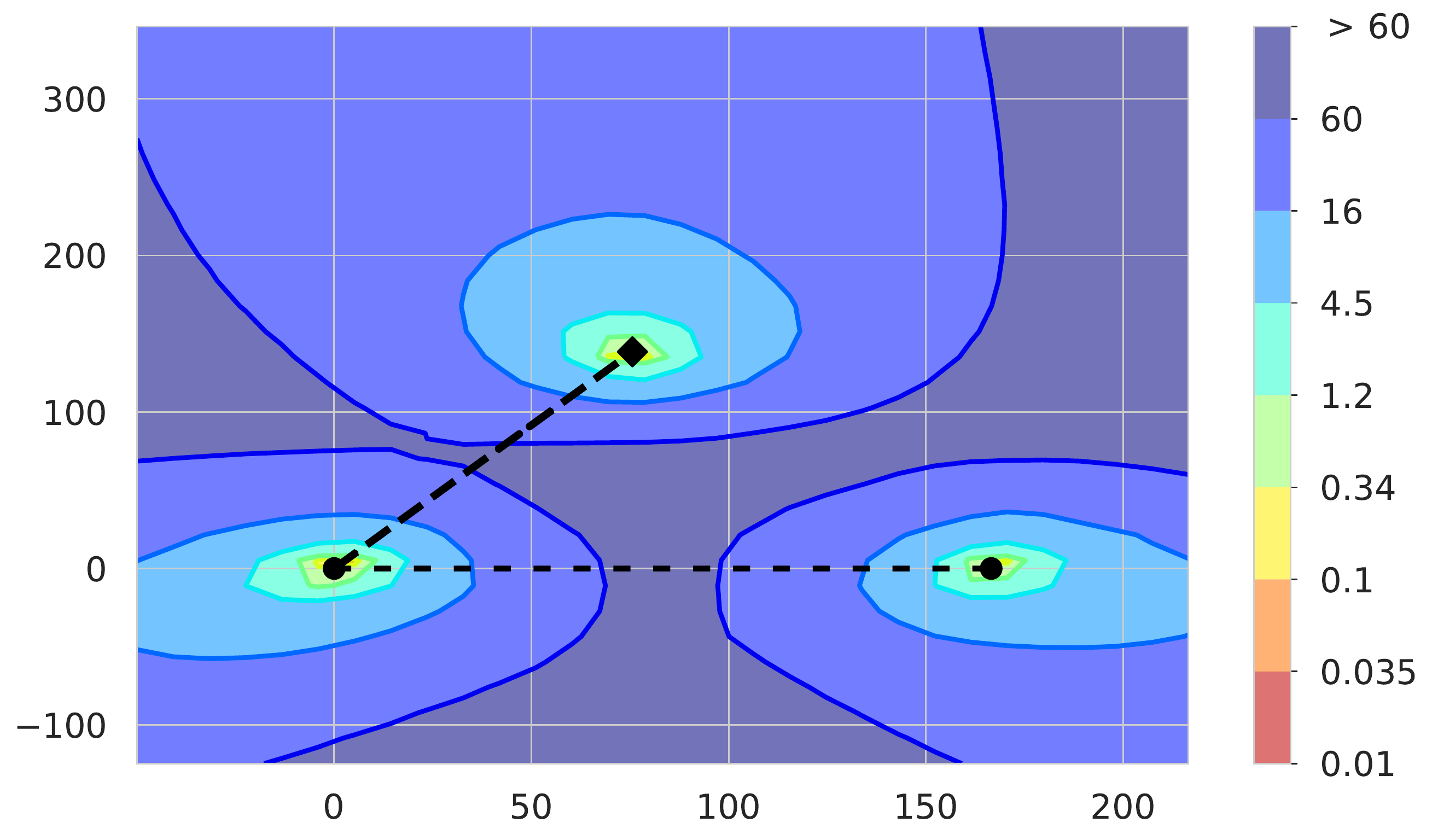}
    \includegraphics[width=0.33\linewidth]{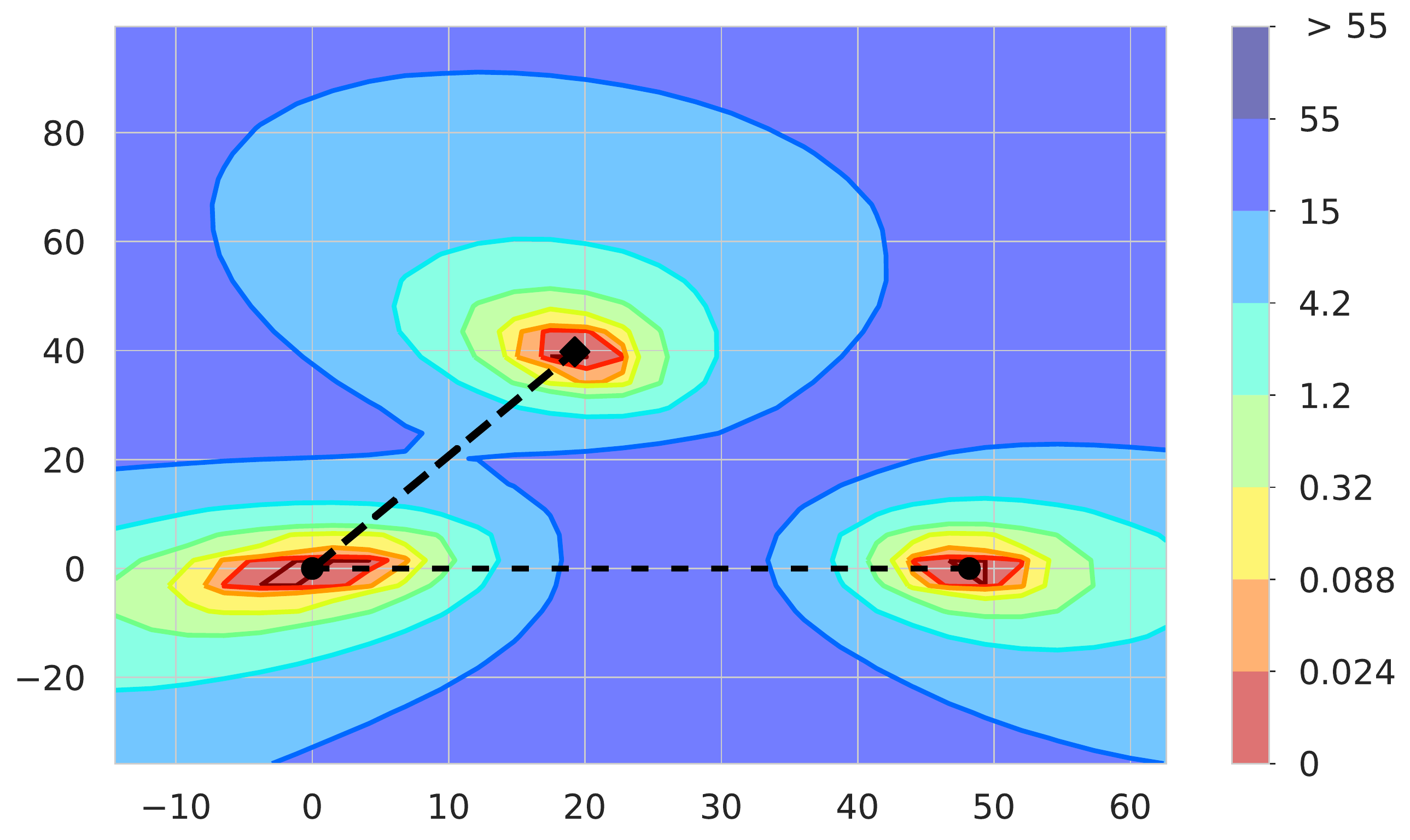}\includegraphics[width=0.33\linewidth]{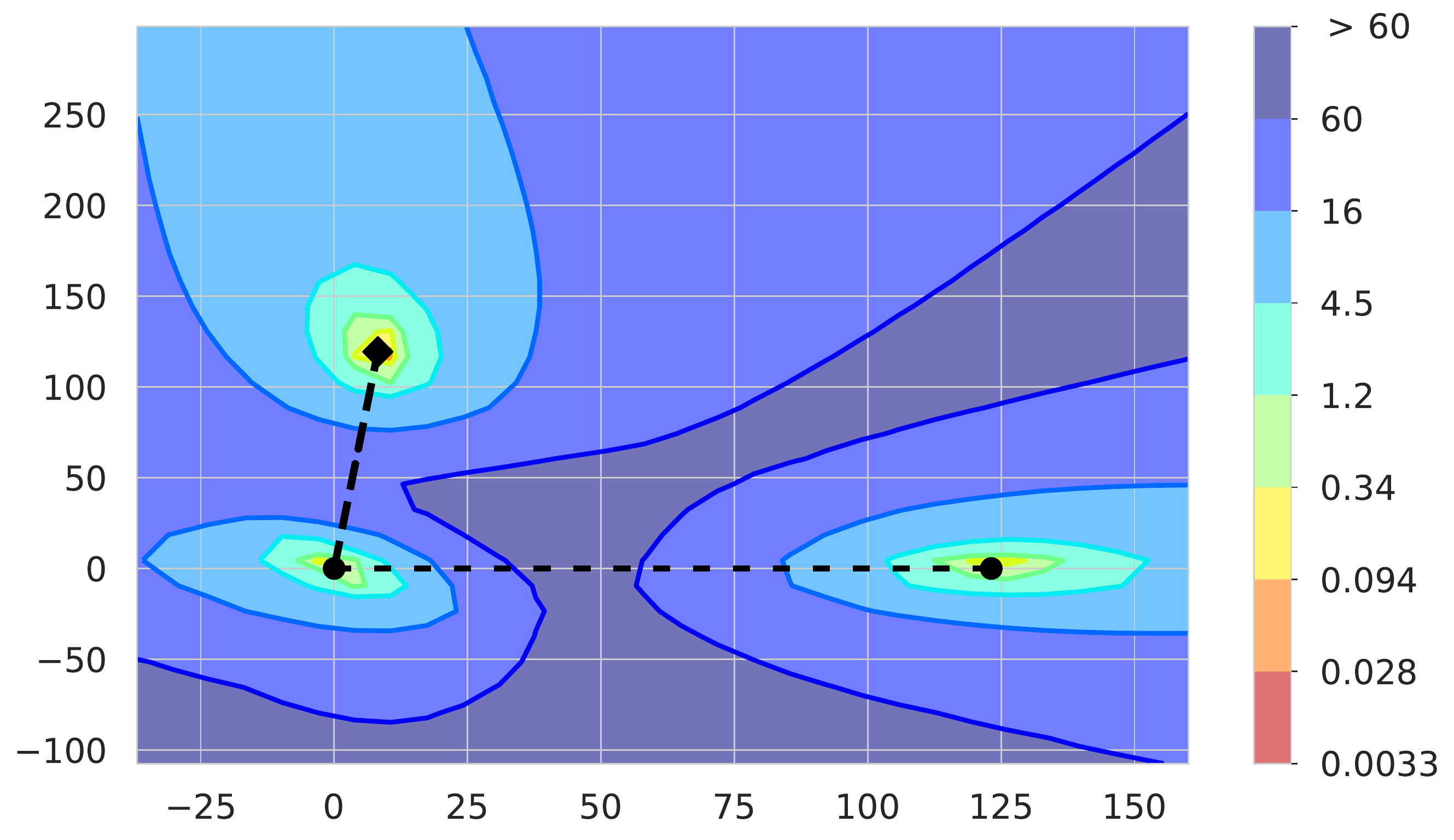}\includegraphics[width=0.33\linewidth]{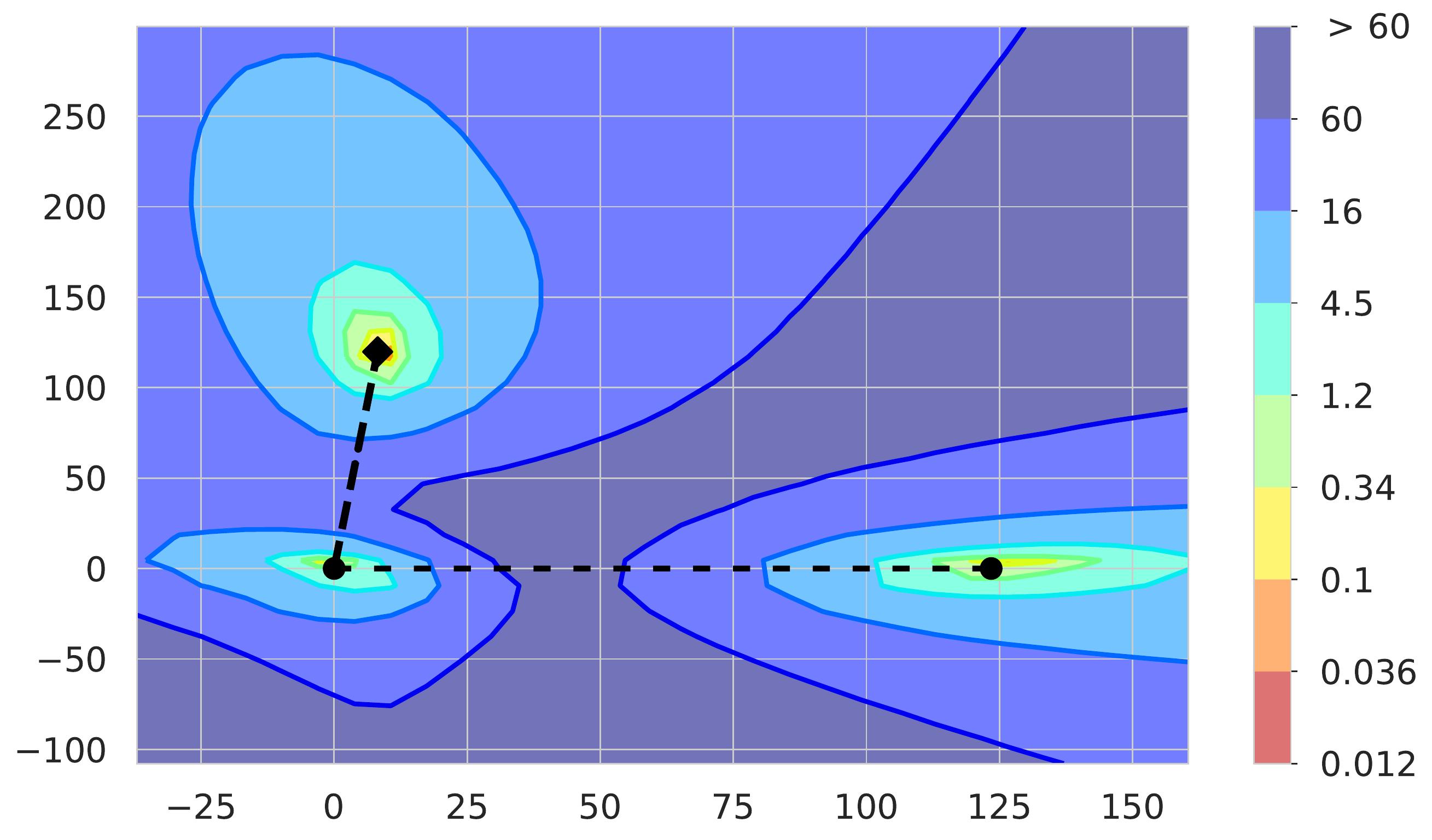}
    \caption{Bi-dimensional sections of the train error, LeNet on Fashion-MNIST, without normalization. 
    Top points are the aligned version of the right point w.r.t. the left point.
    Top row: (left) left-right points: RSGD-RSGD; (middle) left-right points: SGD-SGD; (right) left-right points: ADV-ADV. 
    Bottom row: (left) left-right points: RSGD-SGD; (middle) left-right points: RSGD-ADV; (right) left-right points: SGD-ADV. 
    Aligning the NNs lowers the barriers and in some cases reveals that solutions lie in closer and connected basins. 
    Dashed lines represent linear paths.}
    \label{figSI:2dsurfaces-2}
    \end{center}
    \vskip -0.2in
\end{figure}

\begin{figure}[H]
    \vskip 0.2in
    \begin{center}
    \includegraphics[width=0.33\linewidth]{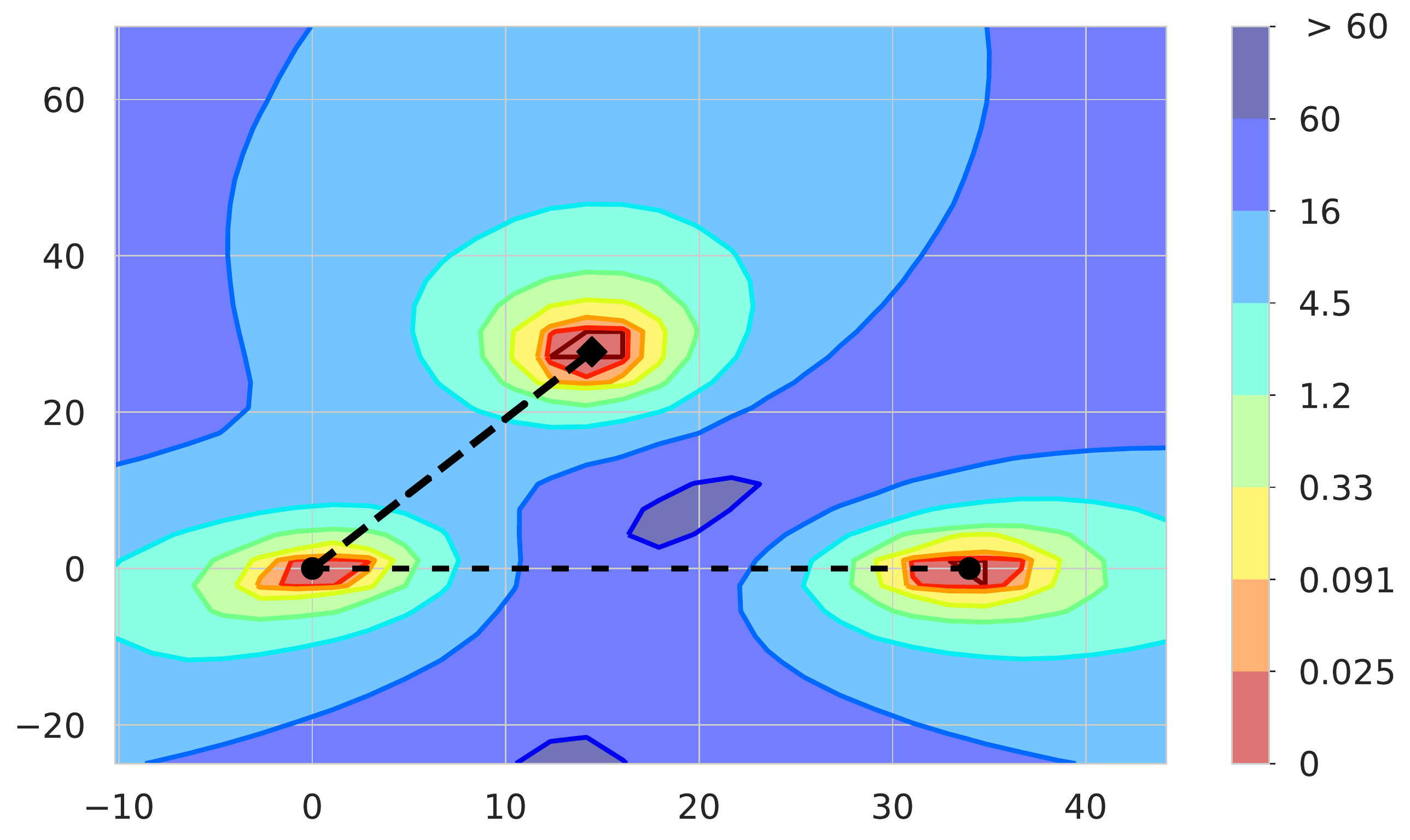}\includegraphics[width=0.33\linewidth]{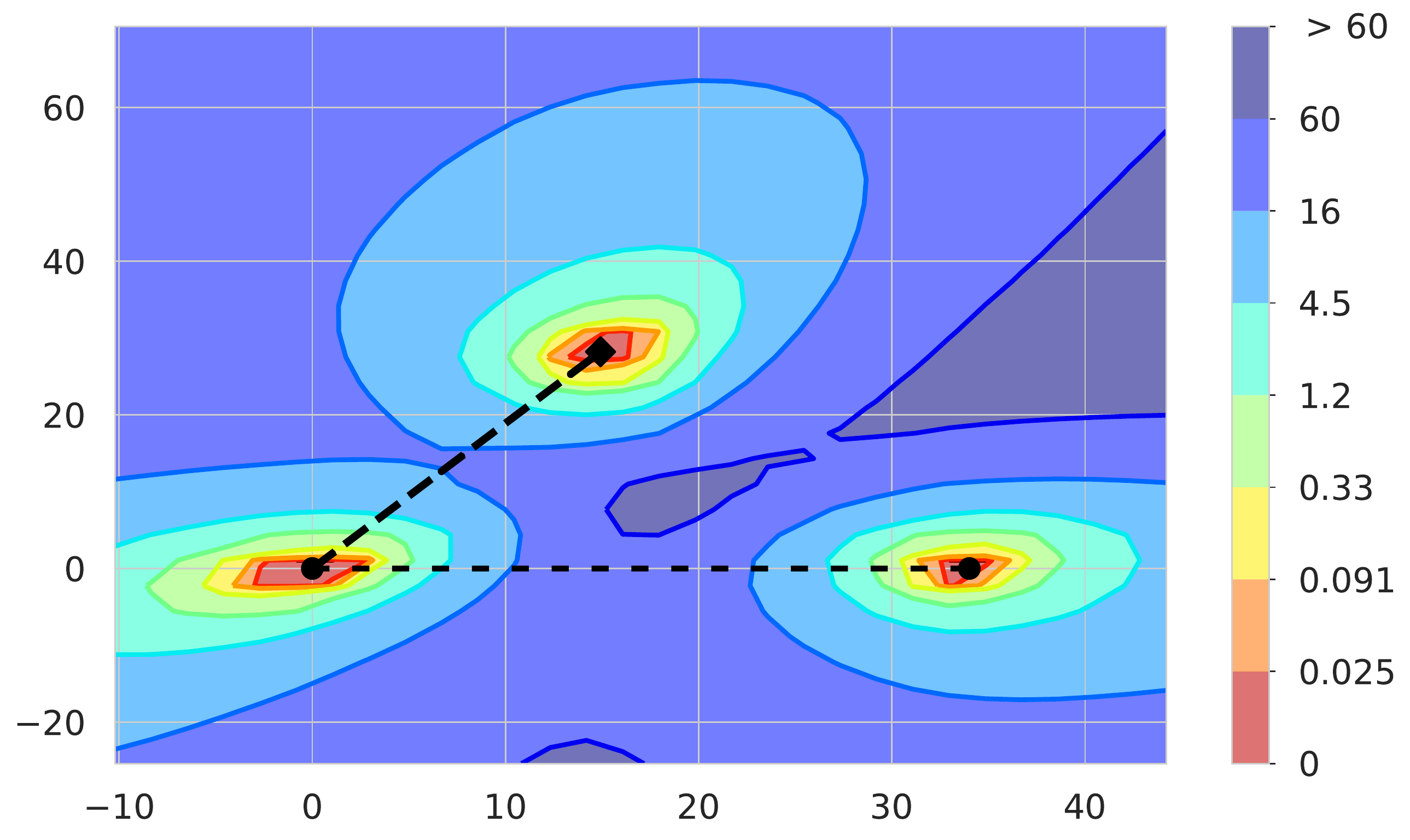}\includegraphics[width=0.33\linewidth]{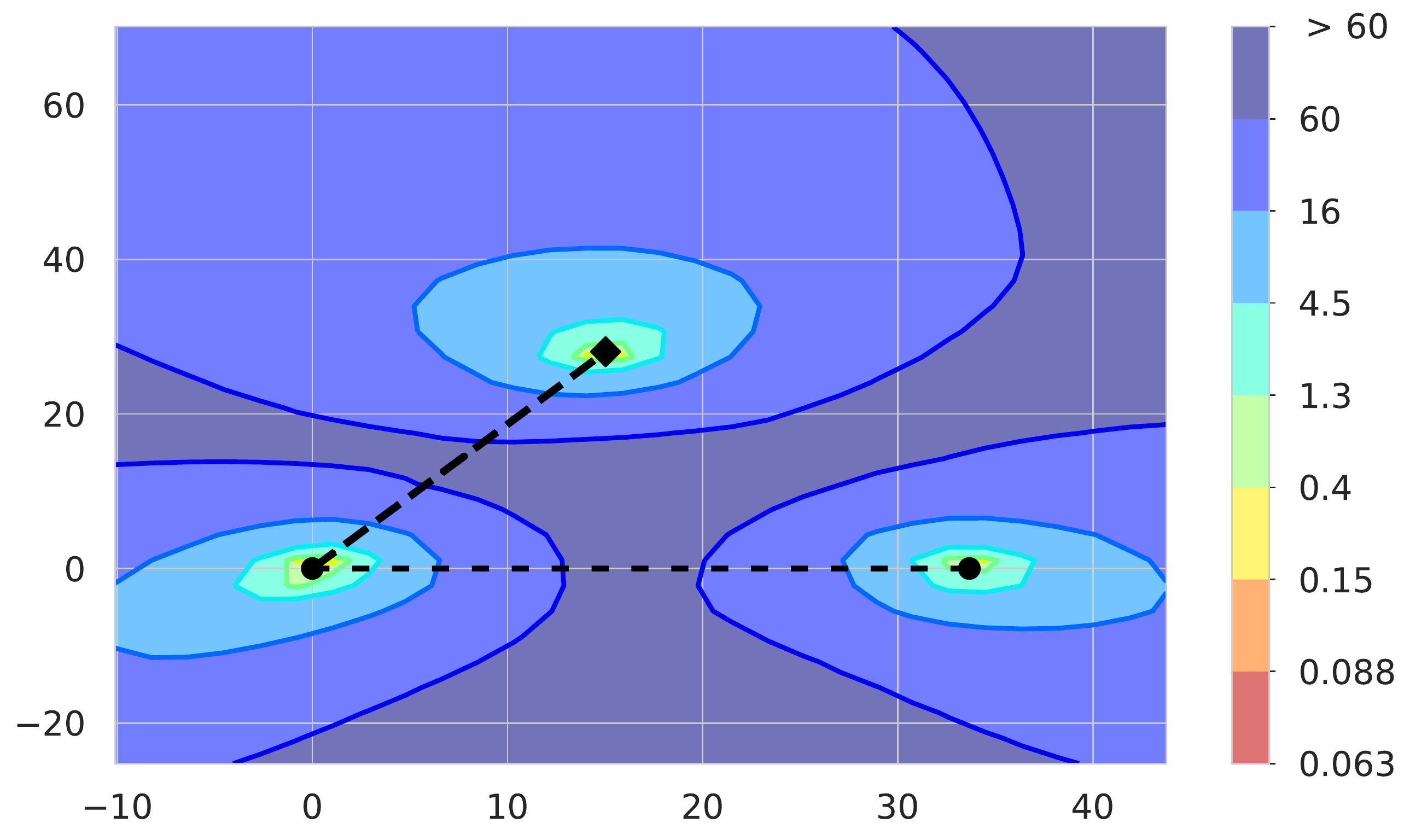}
    \includegraphics[width=0.33\linewidth]{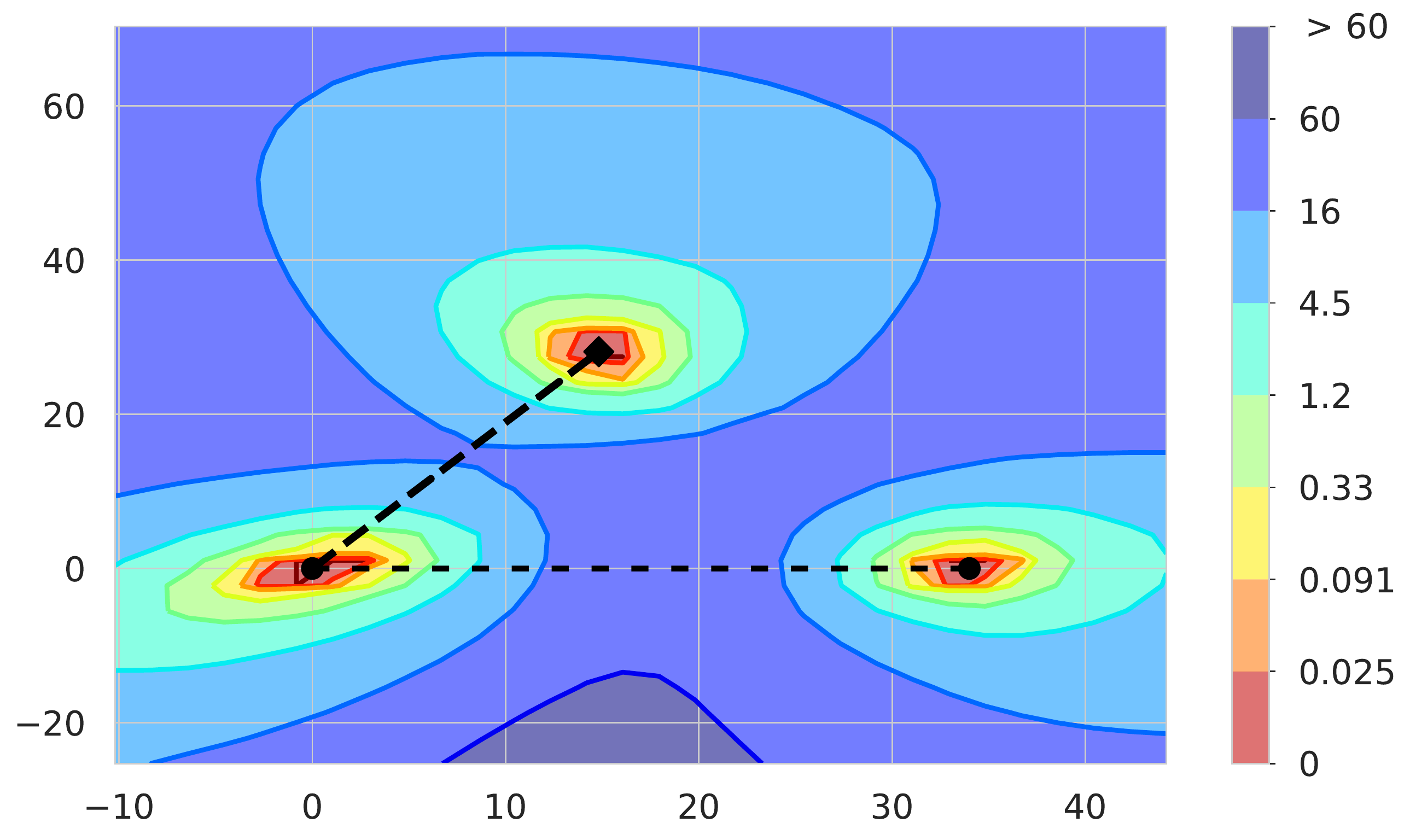}\includegraphics[width=0.33\linewidth]{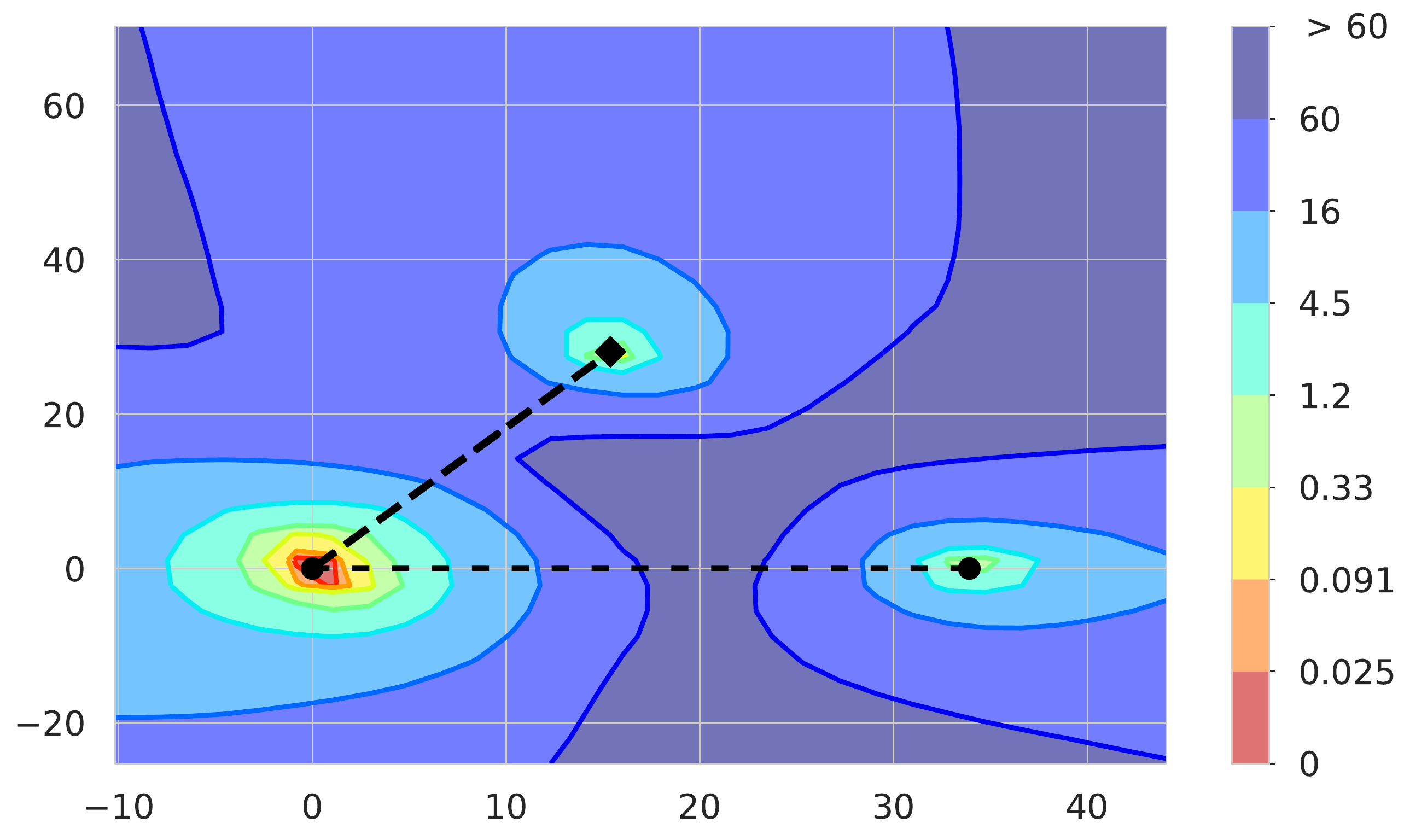}\includegraphics[width=0.33\linewidth]{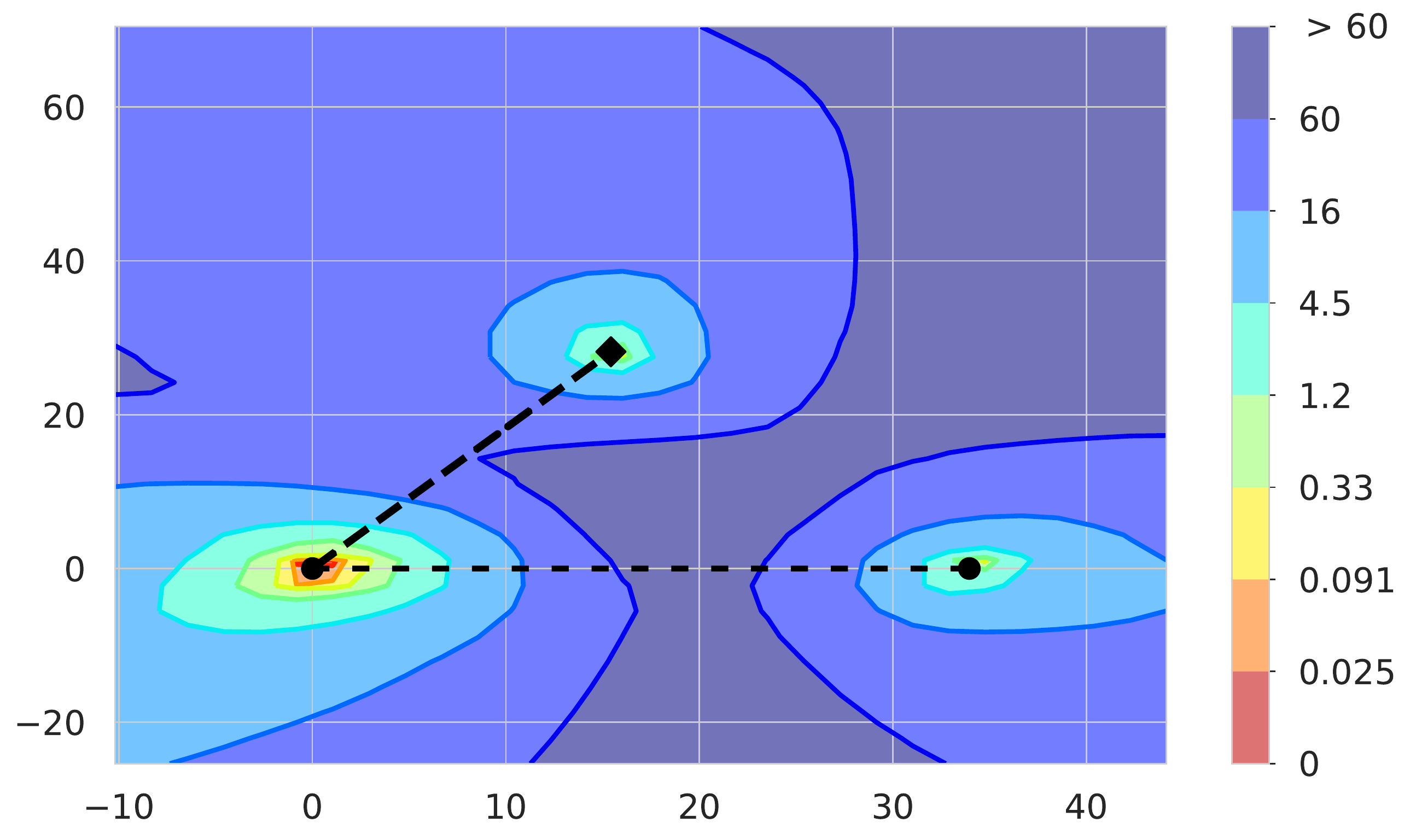}
    \caption{Bi-dimensional sections of the train error, LeNet on Fashion-MNIST, with normalization. 
    Top points are always the aligned version of the right point w.r.t. the left point.
    Top row: (left) left-right points: RSGD-RSGD; (middle) left-right points: SGD-SGD; (right) left-right points: ADV-ADV. 
    Bottom row: (left) left-right points: RSGD-SGD; (middle) left-right points: RSGD-ADV; (right) left-right points: SGD-ADV. 
    Normalization reveals the geometry around the solutions. 
    Dashed lines represent distorted geodesic paths.}
    \label{figSI:2dsurfaces-3}
    \end{center}
    \vskip -0.2in
\end{figure}

\begin{figure}[H]
    \vskip 0.2in
    \begin{center}
    \includegraphics[width=0.33\linewidth]{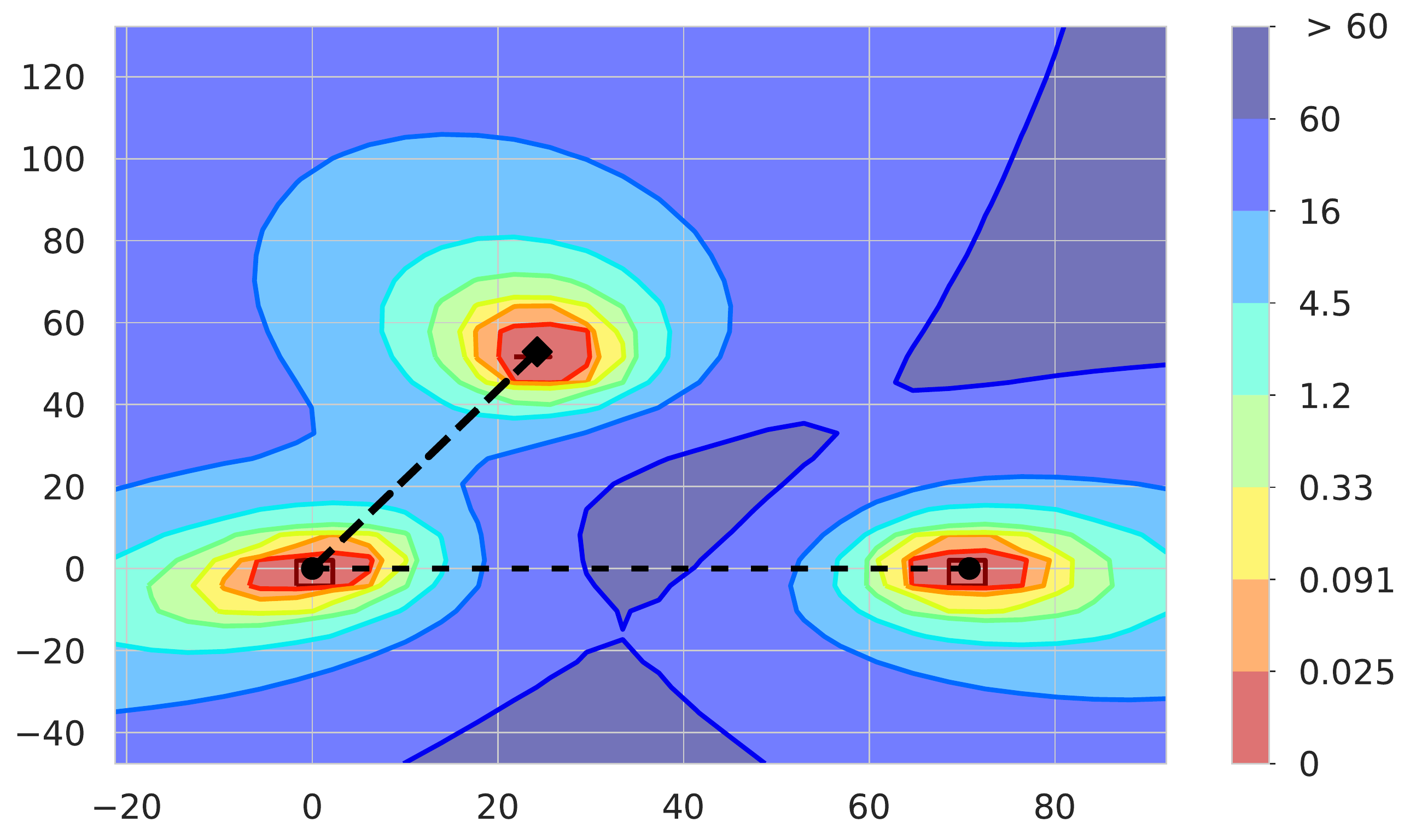}\includegraphics[width=0.33\linewidth]{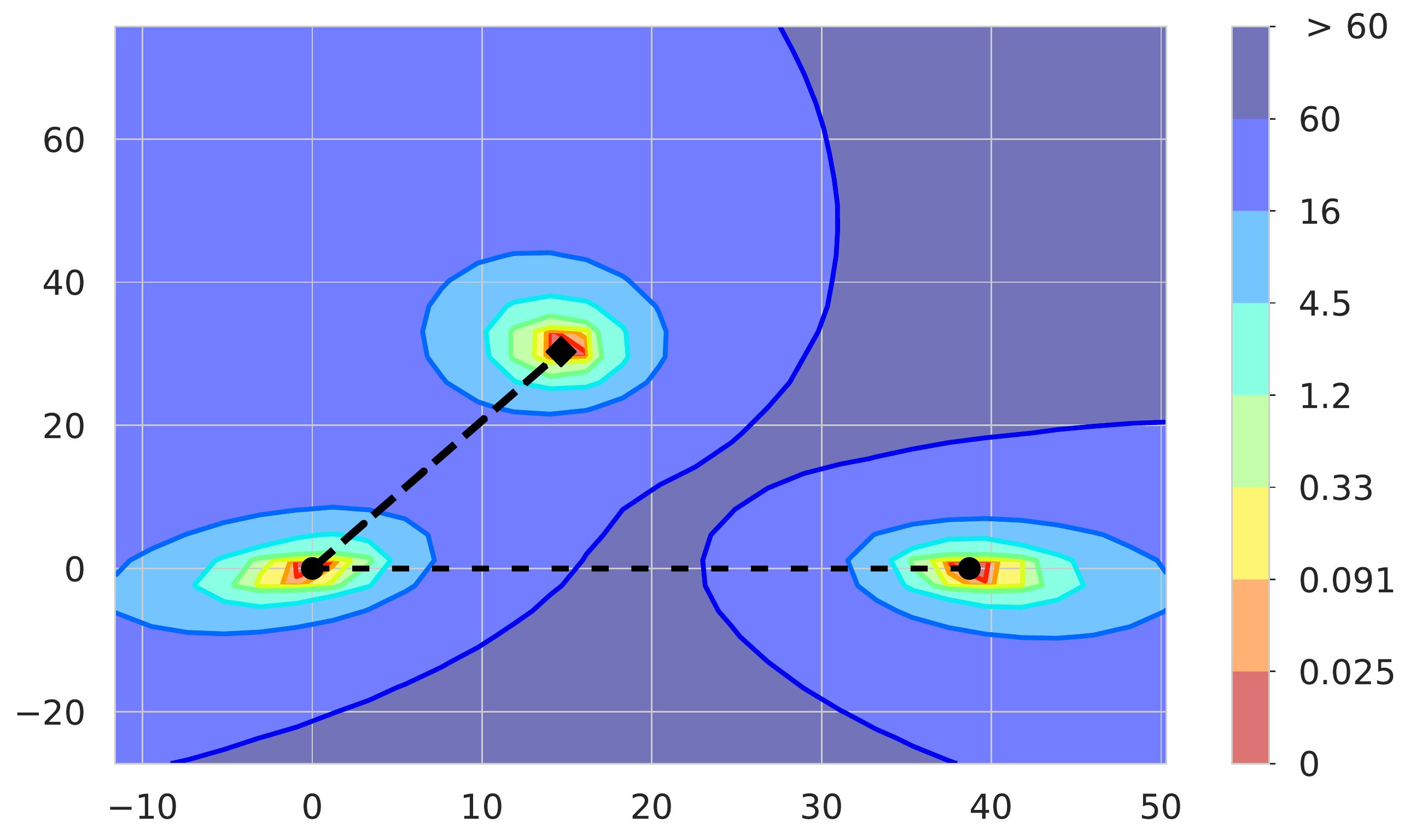}\includegraphics[width=0.33\linewidth]{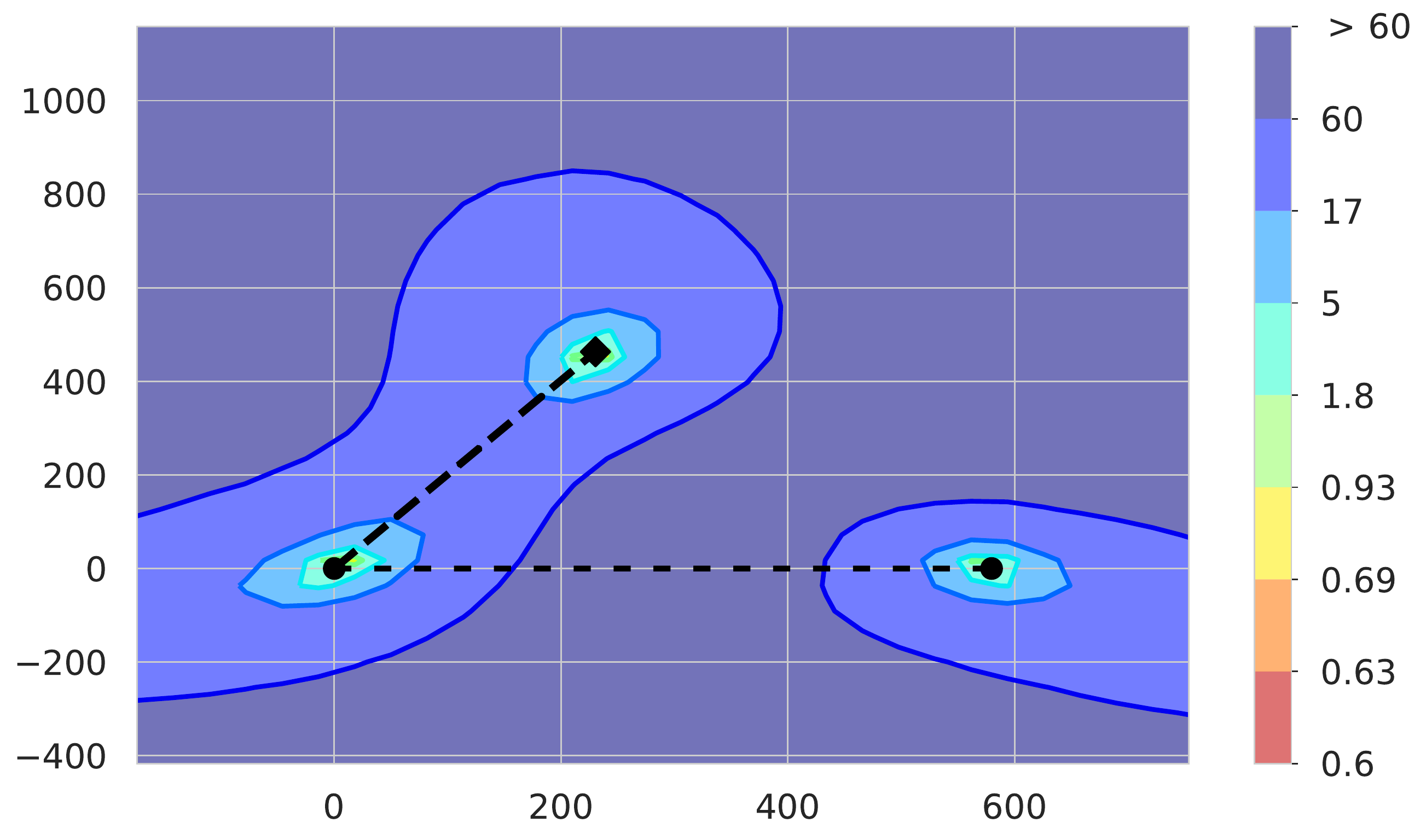}
    \includegraphics[width=0.33\linewidth]{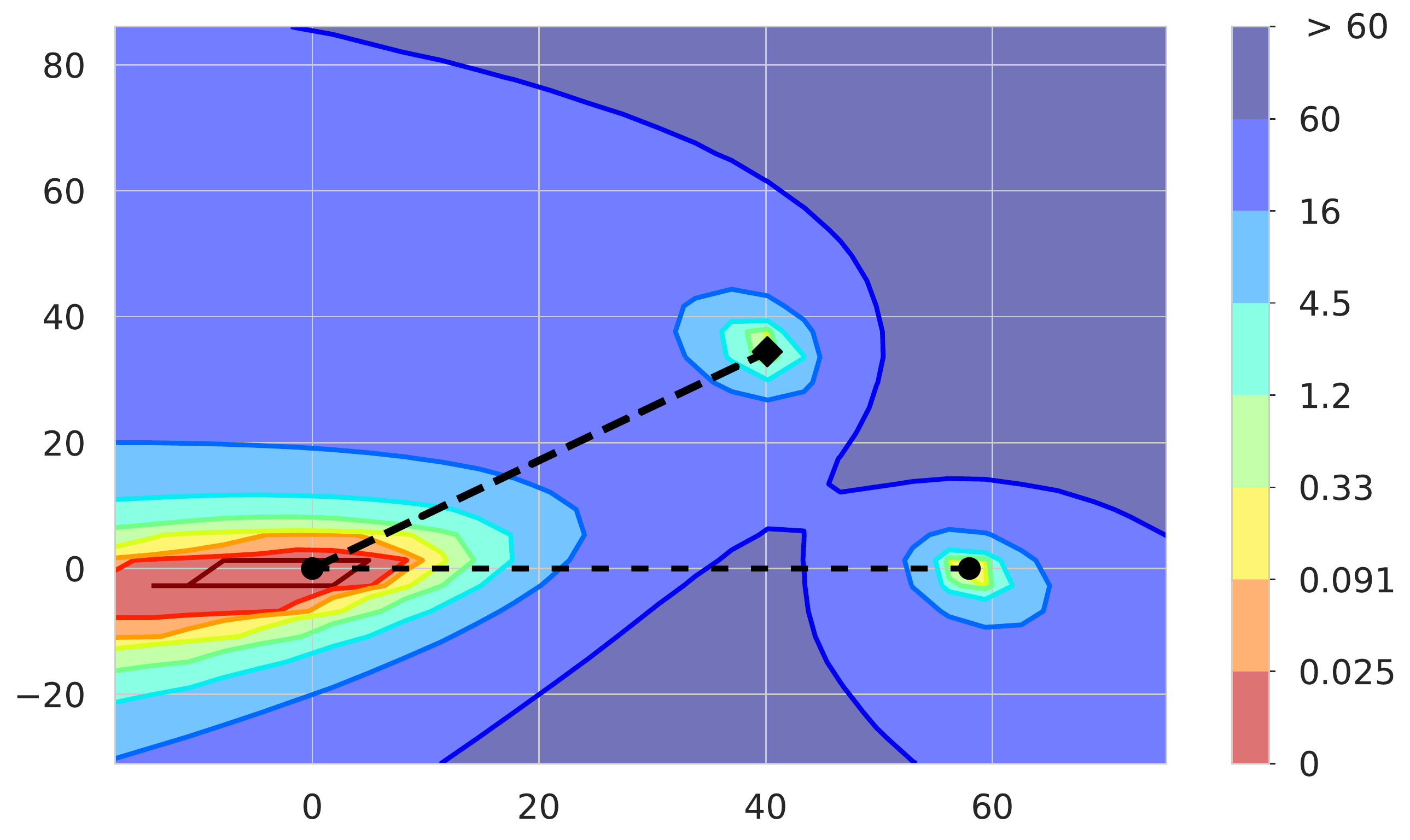}\includegraphics[width=0.33\linewidth]{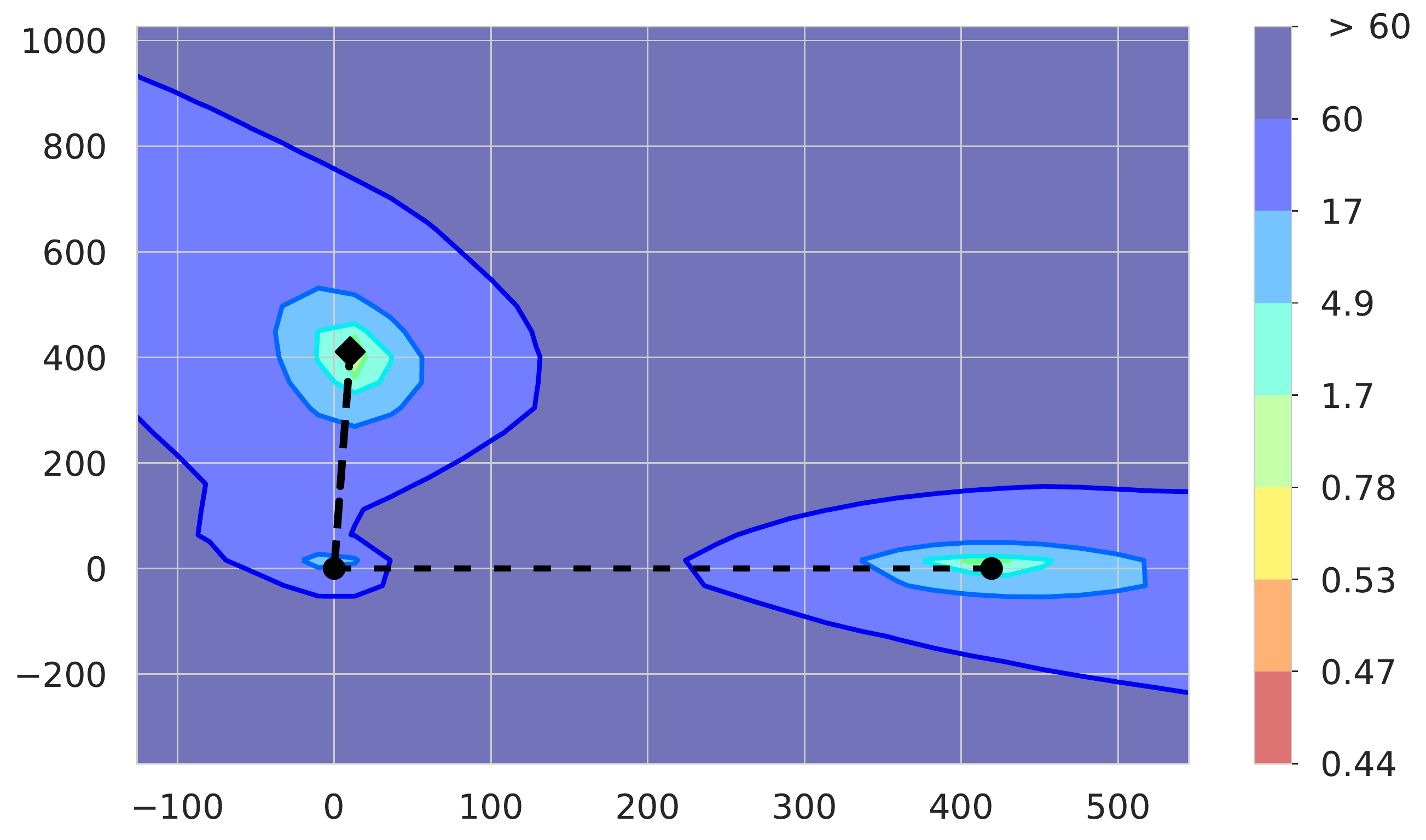}\includegraphics[width=0.33\linewidth]{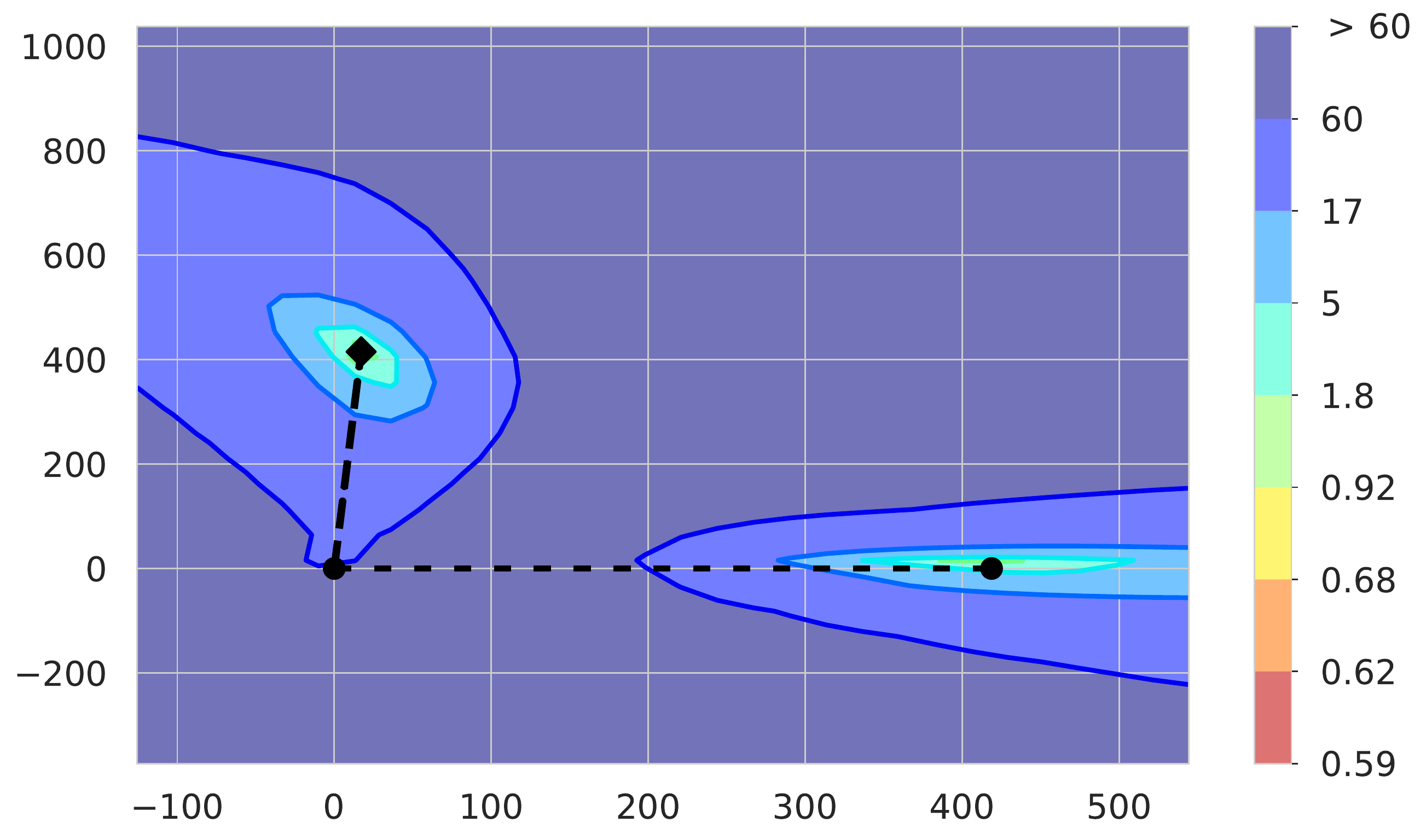}
    \caption{Bi-dimensional sections of the train error, MLP on CIFAR-10, without normalization. 
    Top points are the aligned version of the right point w.r.t. the left point.
    Top row: (left) left-right points: RSGD-RSGD; (middle) left-right points: SGD-SGD; (right) left-right points: ADV-ADV. 
    Bottom row: (left) left-right points: RSGD-SGD; (middle) left-right points: RSGD-ADV; (right) left-right points: SGD-ADV. 
    Aligning the NNs lowers the barriers and in some cases reveals that solutions lie in closer and connected basins. 
    Dashed lines represent linear paths.}
    \label{figSI:2dsurfaces-4}
    \end{center}
    \vskip -0.2in
\end{figure}

\begin{figure}[H]
    \vskip 0.2in
    \begin{center}
    \includegraphics[width=0.33\linewidth]{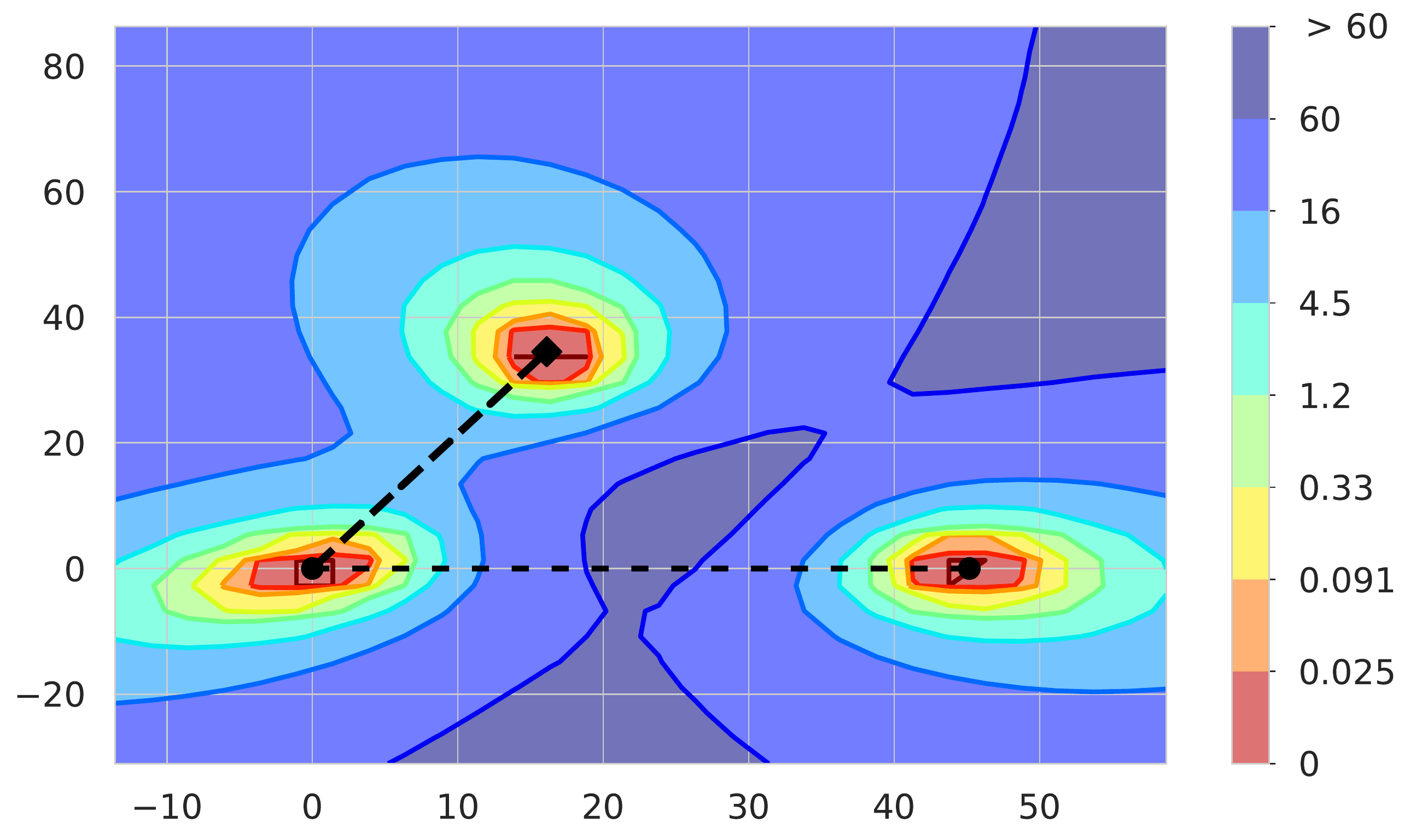}\includegraphics[width=0.33\linewidth]{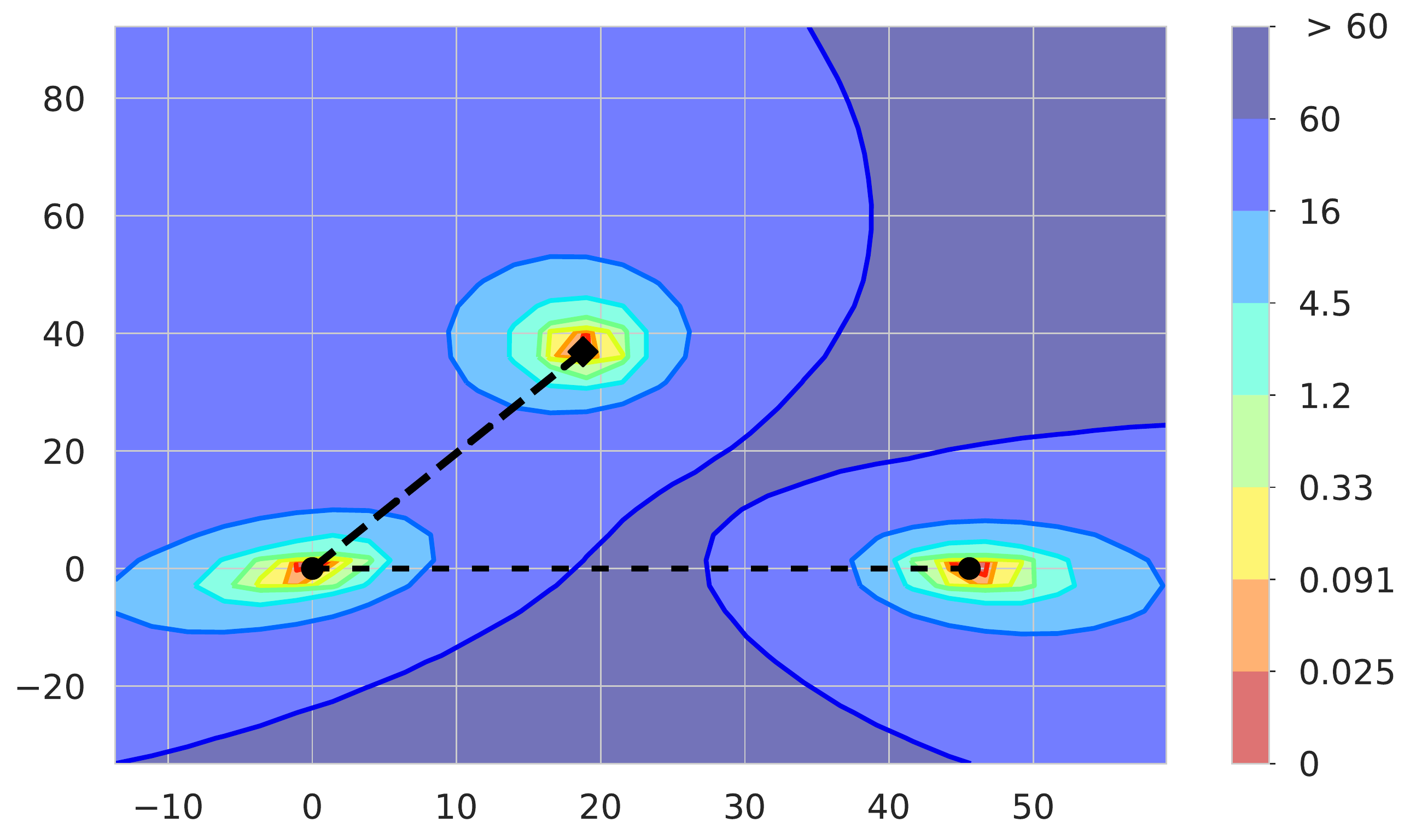}\includegraphics[width=0.33\linewidth]{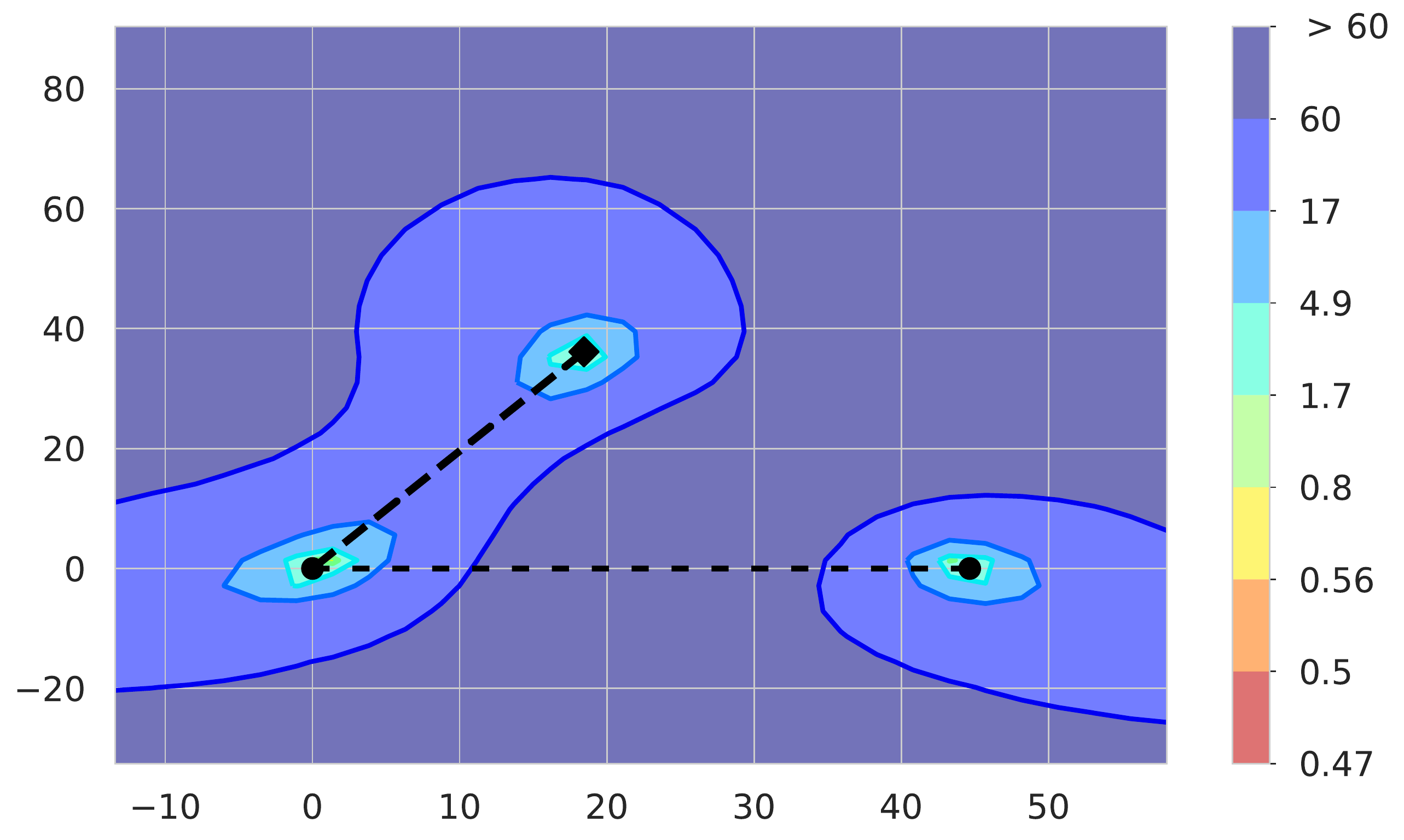}
    \includegraphics[width=0.33\linewidth]{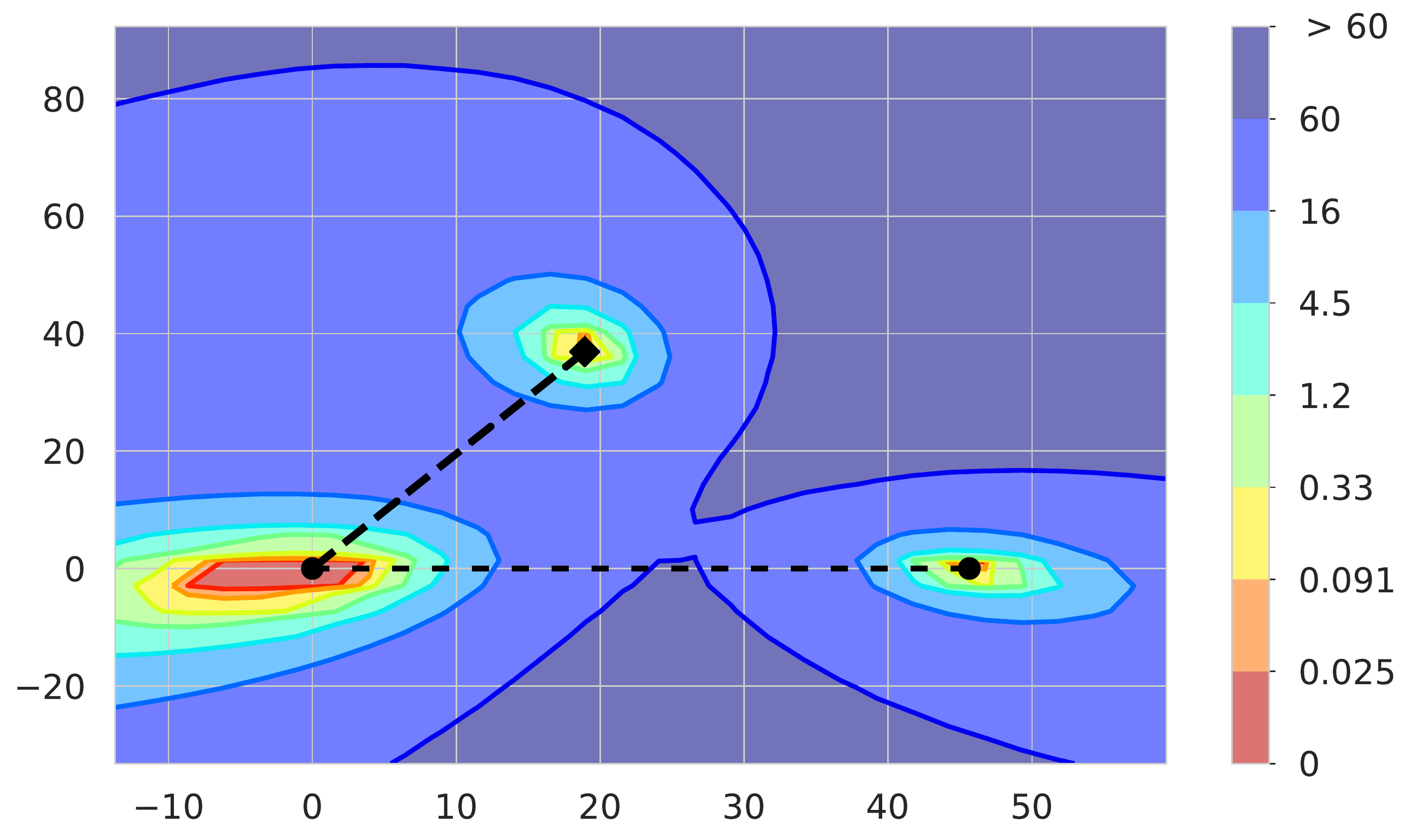}\includegraphics[width=0.33\linewidth]{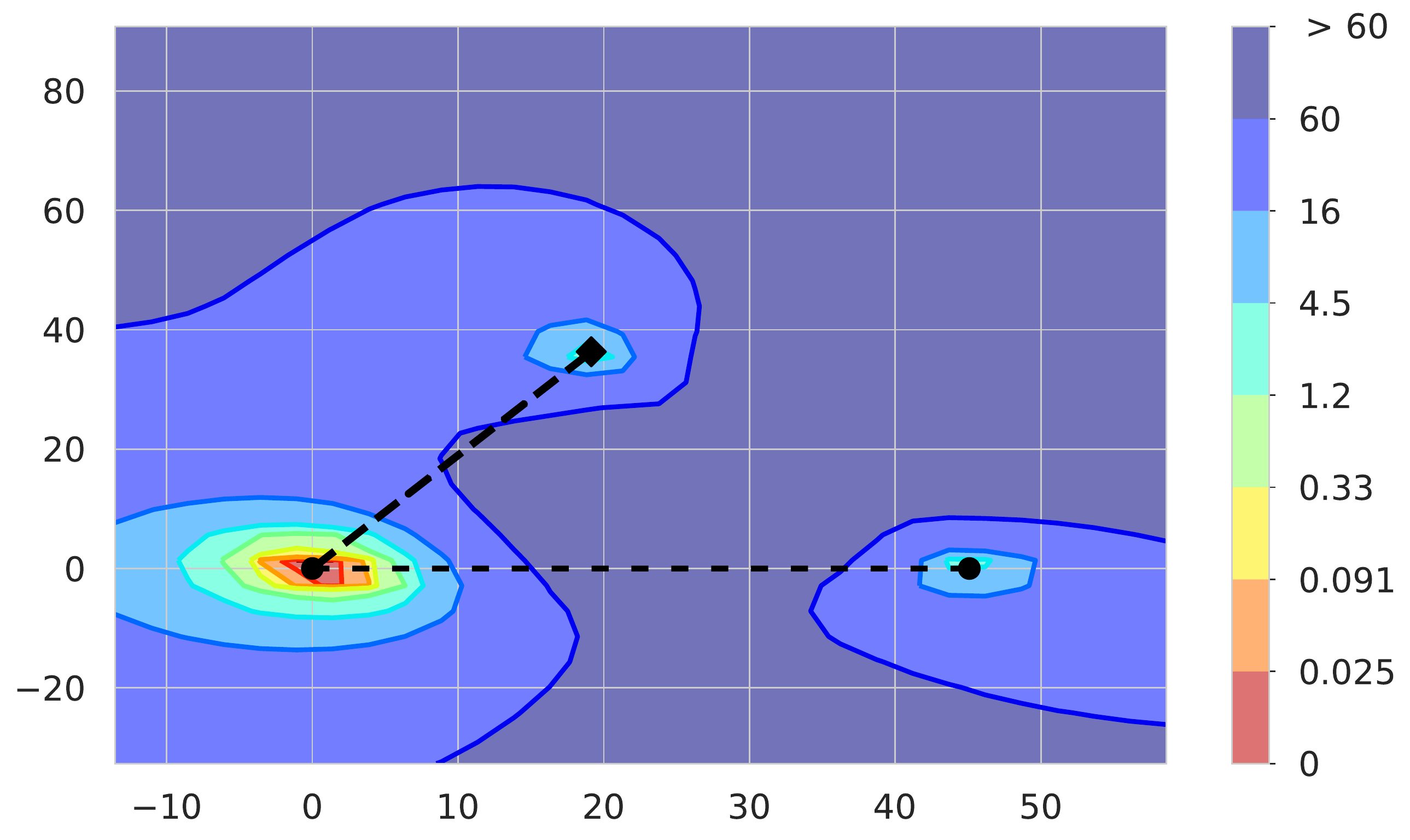}\includegraphics[width=0.33\linewidth]{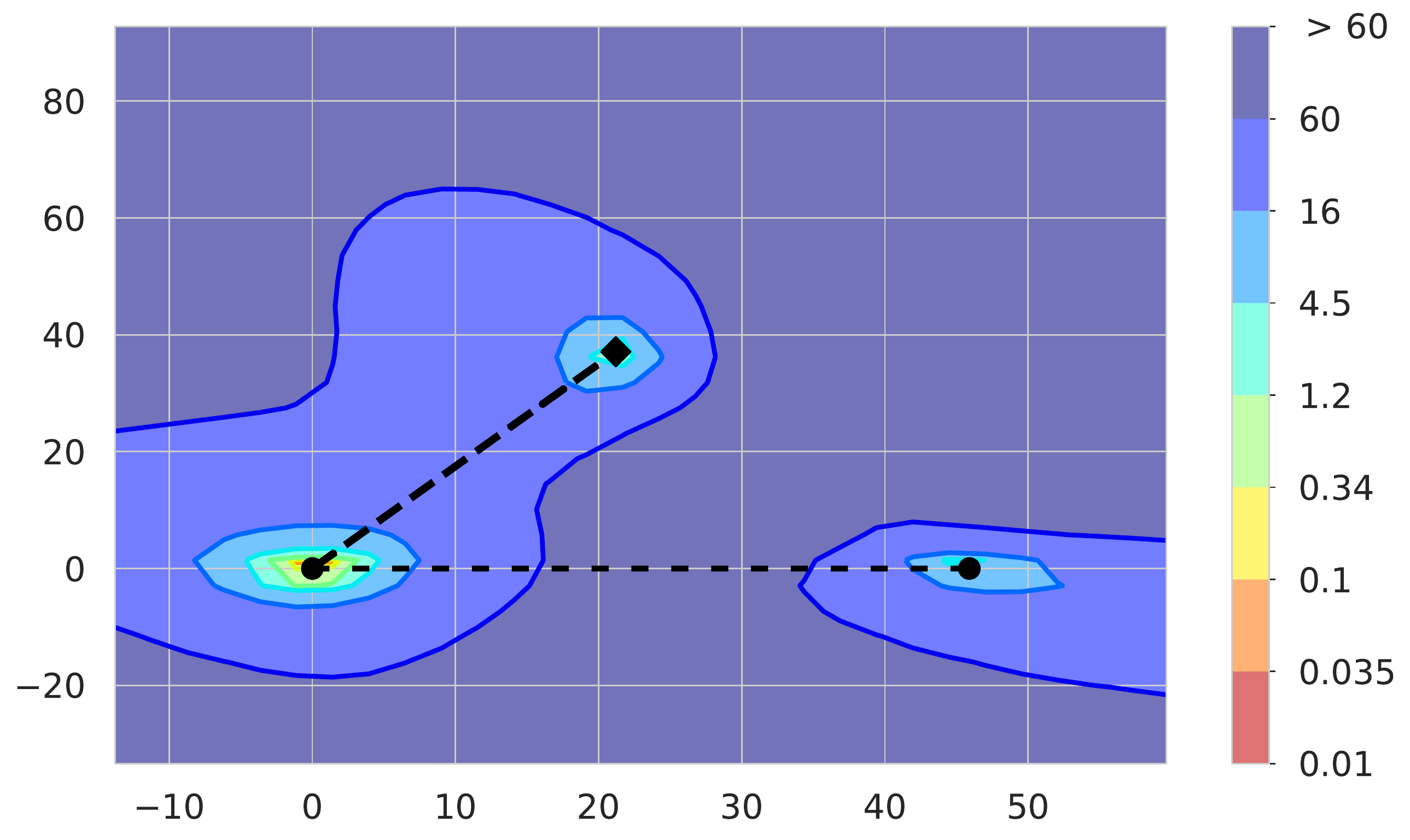}
    \caption{Bi-dimensional sections of the train error, MLP on CIFAR-10, with normalization. 
    Top points are always the aligned version of the right point w.r.t. the left point.
    Top row: (left) left-right points: RSGD-RSGD; (middle) left-right points: SGD-SGD; (right) left-right points: ADV-ADV. 
    Bottom row: (left) left-right points: RSGD-SGD; (middle) left-right points: RSGD-ADV; (right) left-right points: SGD-ADV. 
    Normalization reveals the geometry around the solutions. 
    Dashed lines represent distorted geodesic paths.}
    \label{figSI:2dsurfaces-5}
    \end{center}
    \vskip -0.2in
\end{figure}

\begin{figure}[H]
    \vskip 0.2in
    \begin{center}
    \includegraphics[width=0.33\linewidth]{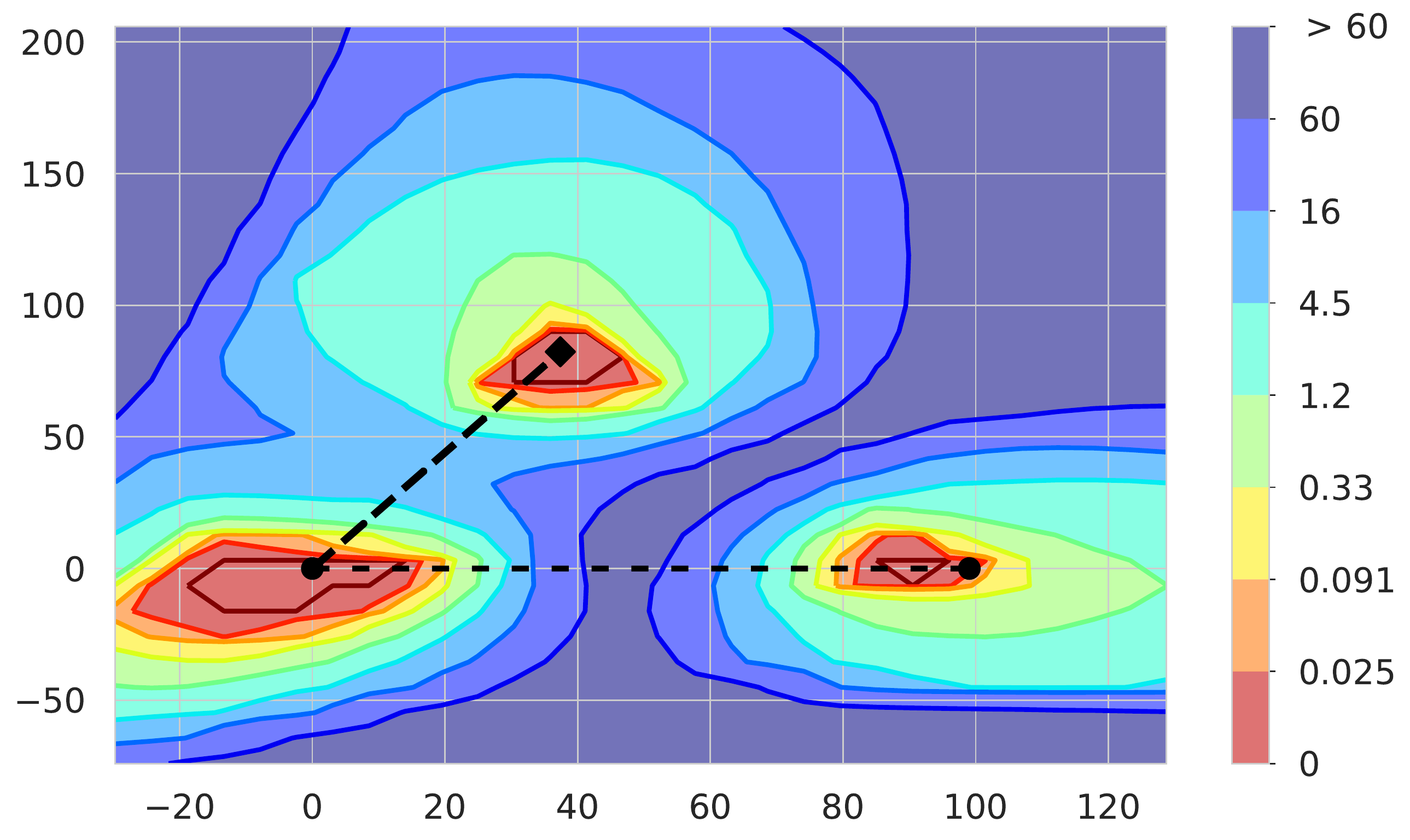}\includegraphics[width=0.33\linewidth]{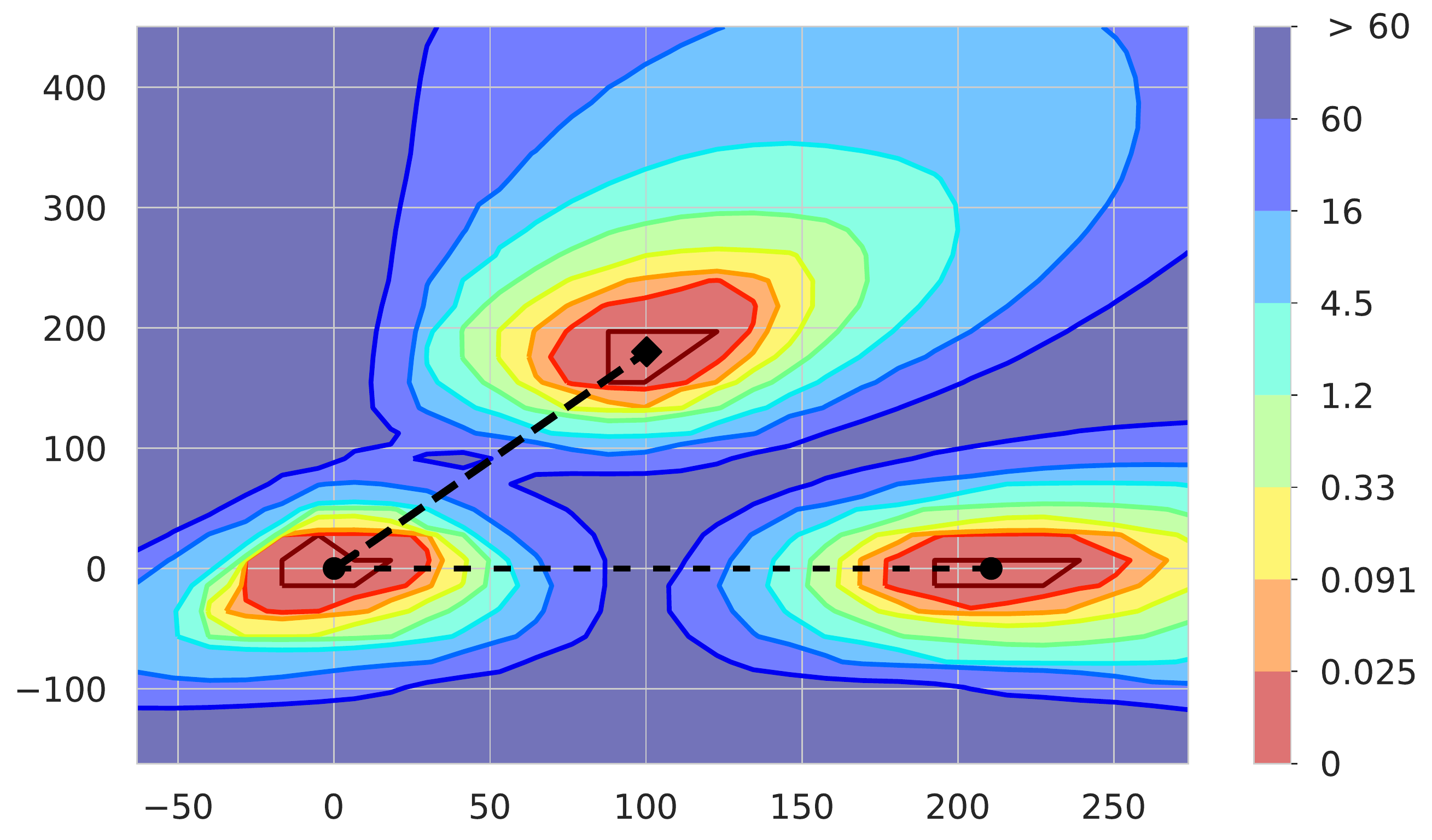}\includegraphics[width=0.33\linewidth]{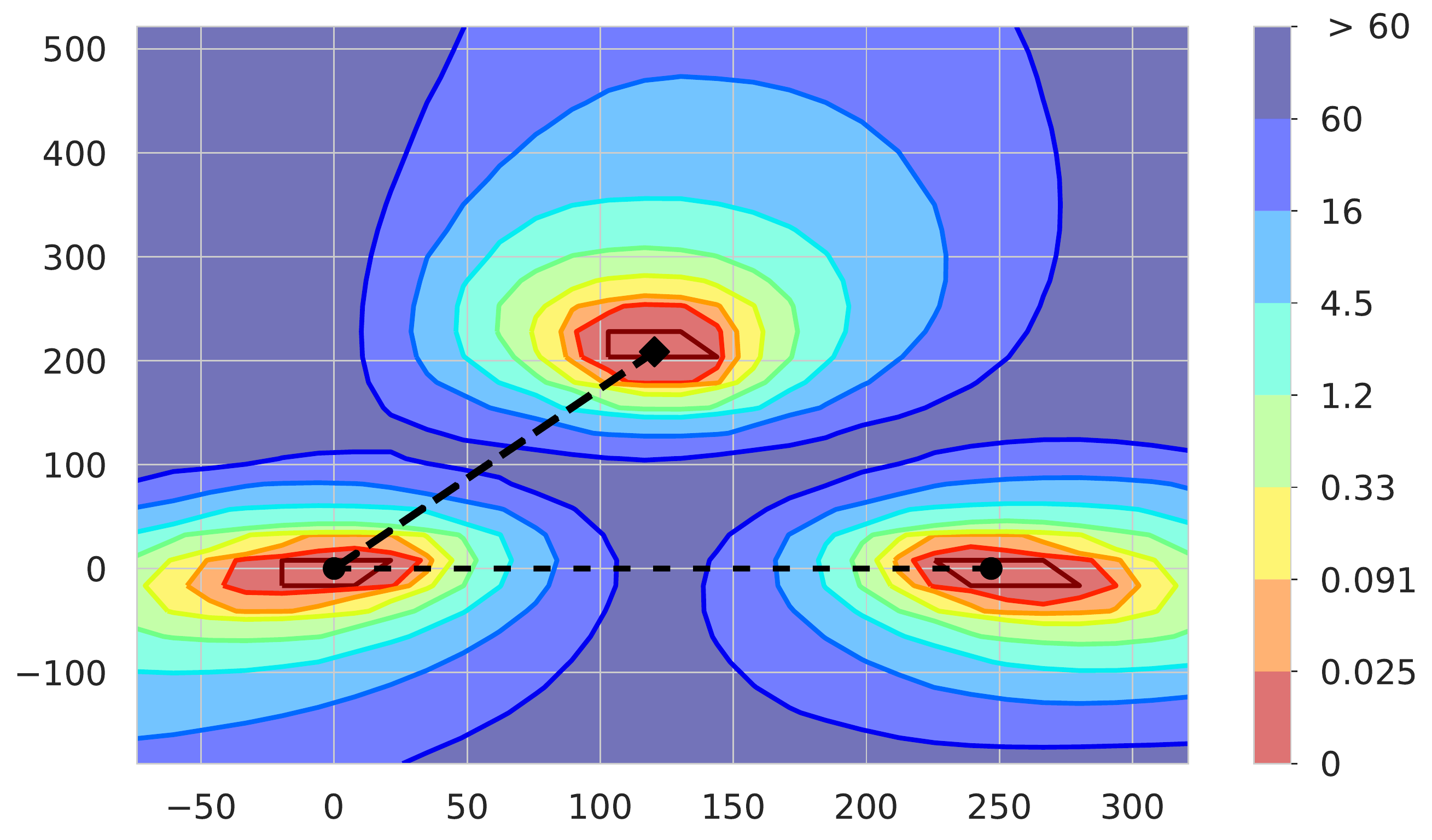}
    \caption{Bi-dimensional sections of the train error, effect of the permutation symmetry on VGG16 trained on CIFAR-10. 
    Top points are the aligned version of the right point w.r.t. the left point.
    (Left) left-right points: RSGD-RSGD; (Middle) left-right points: SGD-SGD; (right) left-right points: ADV-ADV. 
    Aligning the NNs lowers the barriers and in some cases reveals that solutions lie in closer and connected basins (for RSGD-RSGD and to a lesser extent for SGD-SGD), while it is not sufficient in other cases (ADV-ADV). 
    Dashed lines represent linear paths.}
    \label{figSI:2dsurfaces-6}
    \end{center}
    \vskip -0.2in
\end{figure}

In this section we report bi-dimensional visualizations for some of the networks/datasets explored in the main paper.
In Fig.~\ref{figSI:2dsurfaces-1} we compare train and test errors for LeNet on Fashion-MNIST (RSGD-ADV); In Figs.~\ref{figSI:2dsurfaces-2} and \ref{figSI:2dsurfaces-3} we report visualizations for all pairs of LeNet on Fashion-MNIST (without and with normalization respectively); In Figs.~\ref{figSI:2dsurfaces-4} and \ref{figSI:2dsurfaces-5} we report visualizations for all pairs of MLP on CIFAR-10 (without and with normalization respectively); in Fig.~\ref{figSI:2dsurfaces-6} we show the effects of permutation symmetry removal on VGG16 on CIFAR-10 for solutions of varying sharpness.


\subsection{Distances\label{SI:distance}}

\begin{figure}[H]
    \vskip 0.2in
    \begin{center}
    \includegraphics[width=0.5\linewidth]{good_figures/distances/distances_MLP_MNIST.pdf}\includegraphics[width=0.5\linewidth]{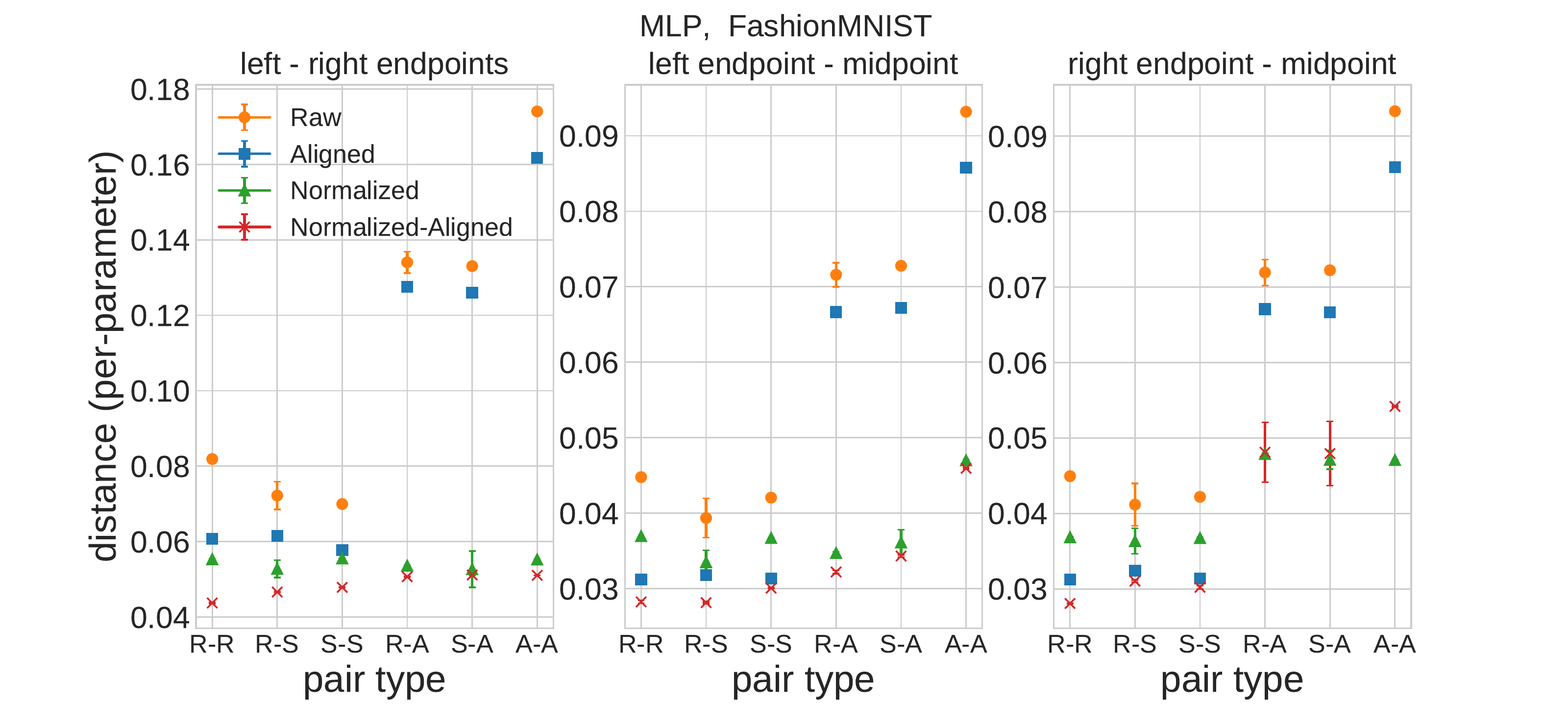}
    \includegraphics[width=0.5\linewidth]{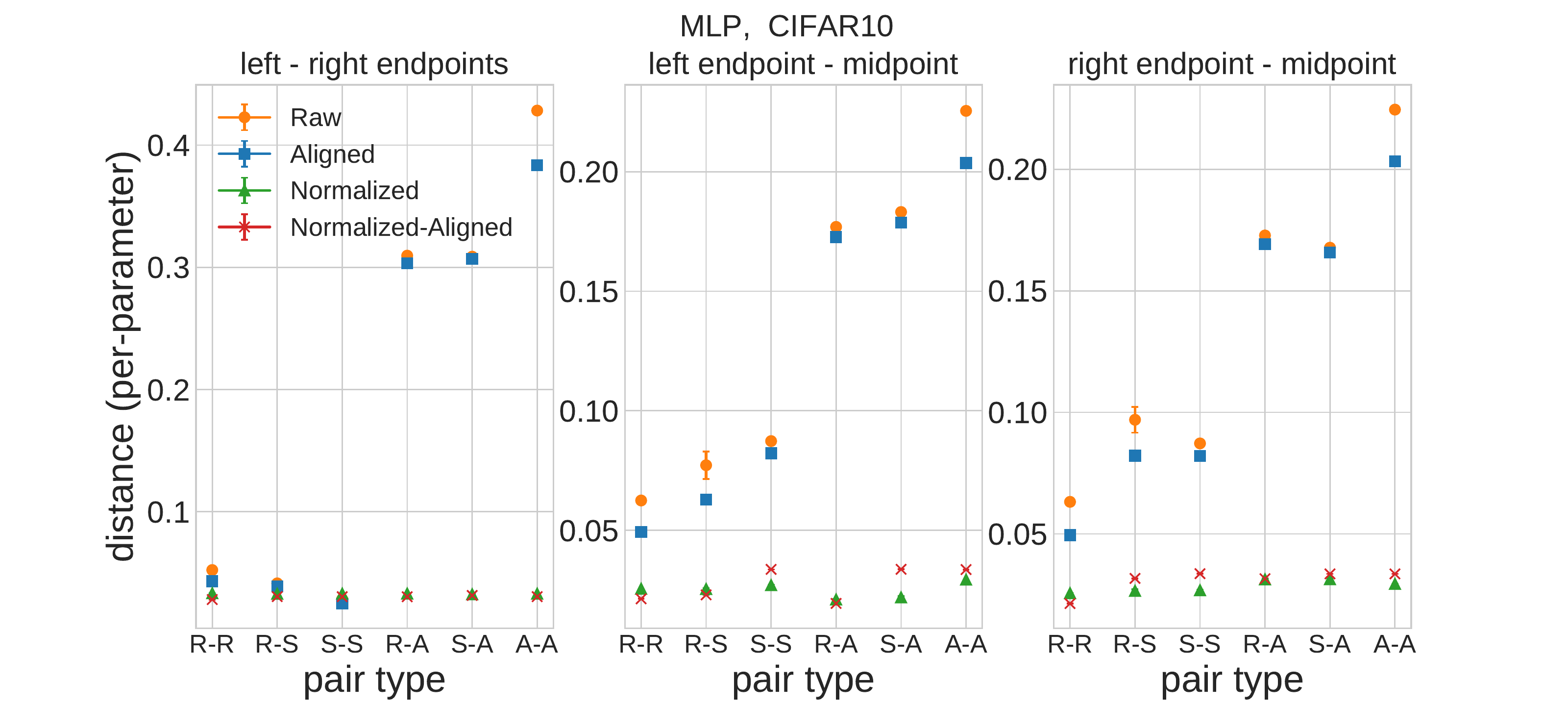}\includegraphics[width=0.5\linewidth]{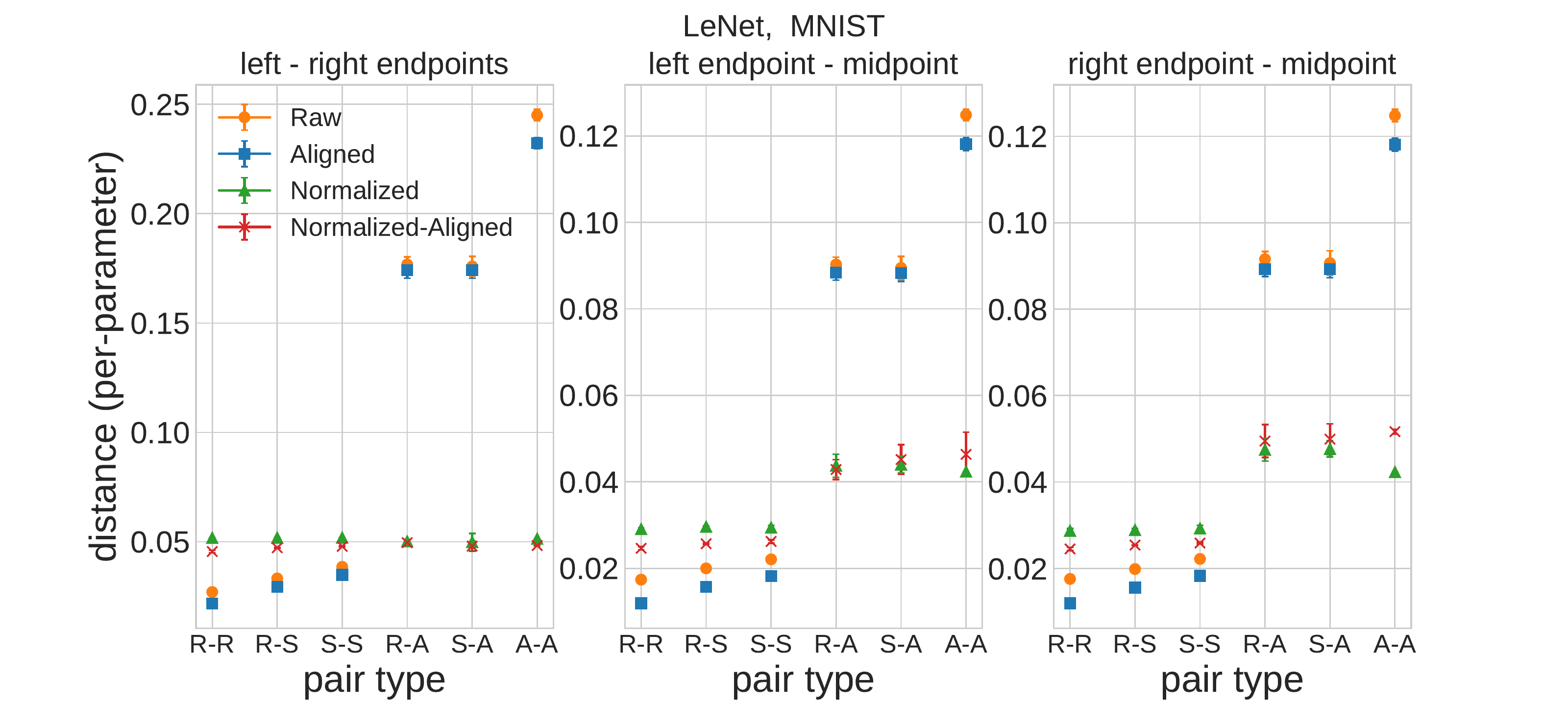}
    \includegraphics[width=0.5\linewidth]{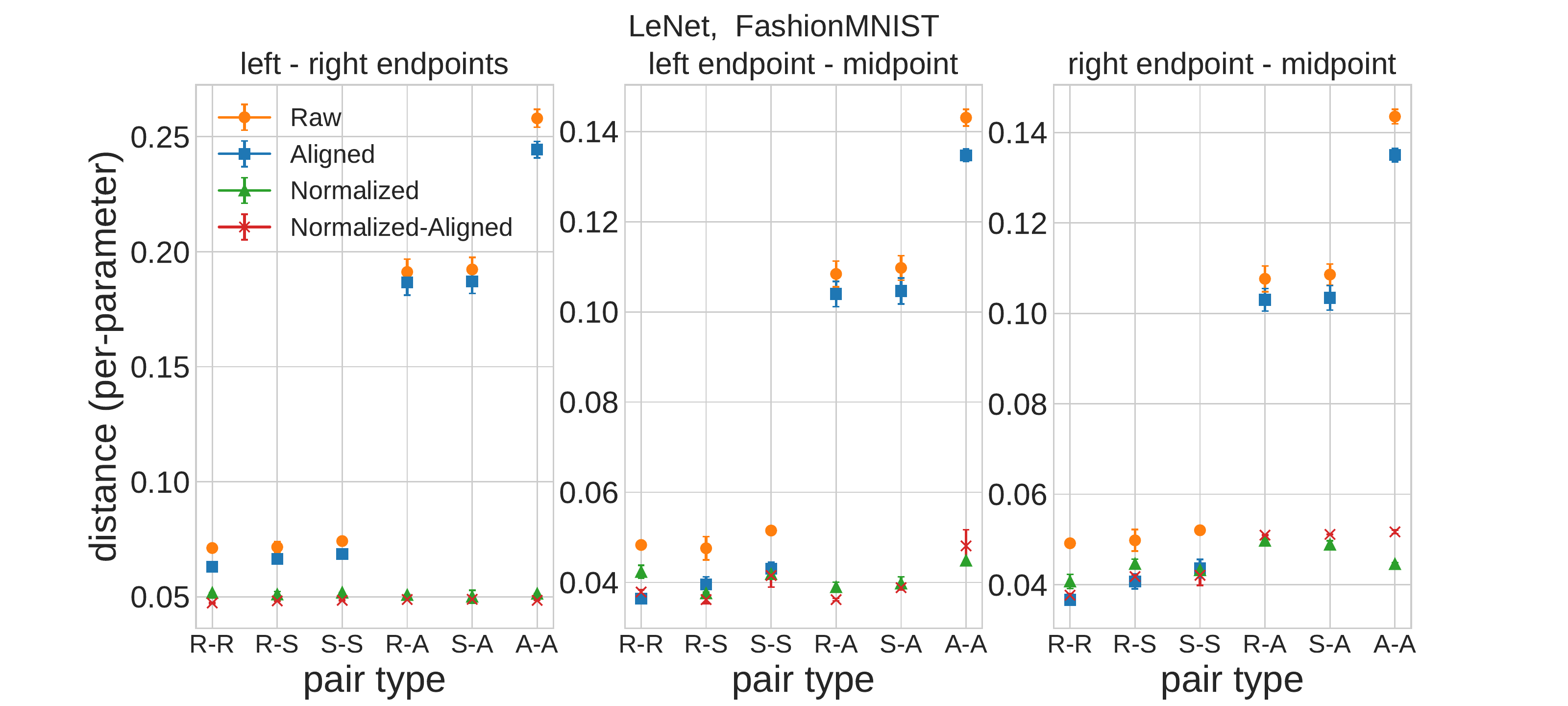}\includegraphics[width=0.5\linewidth]{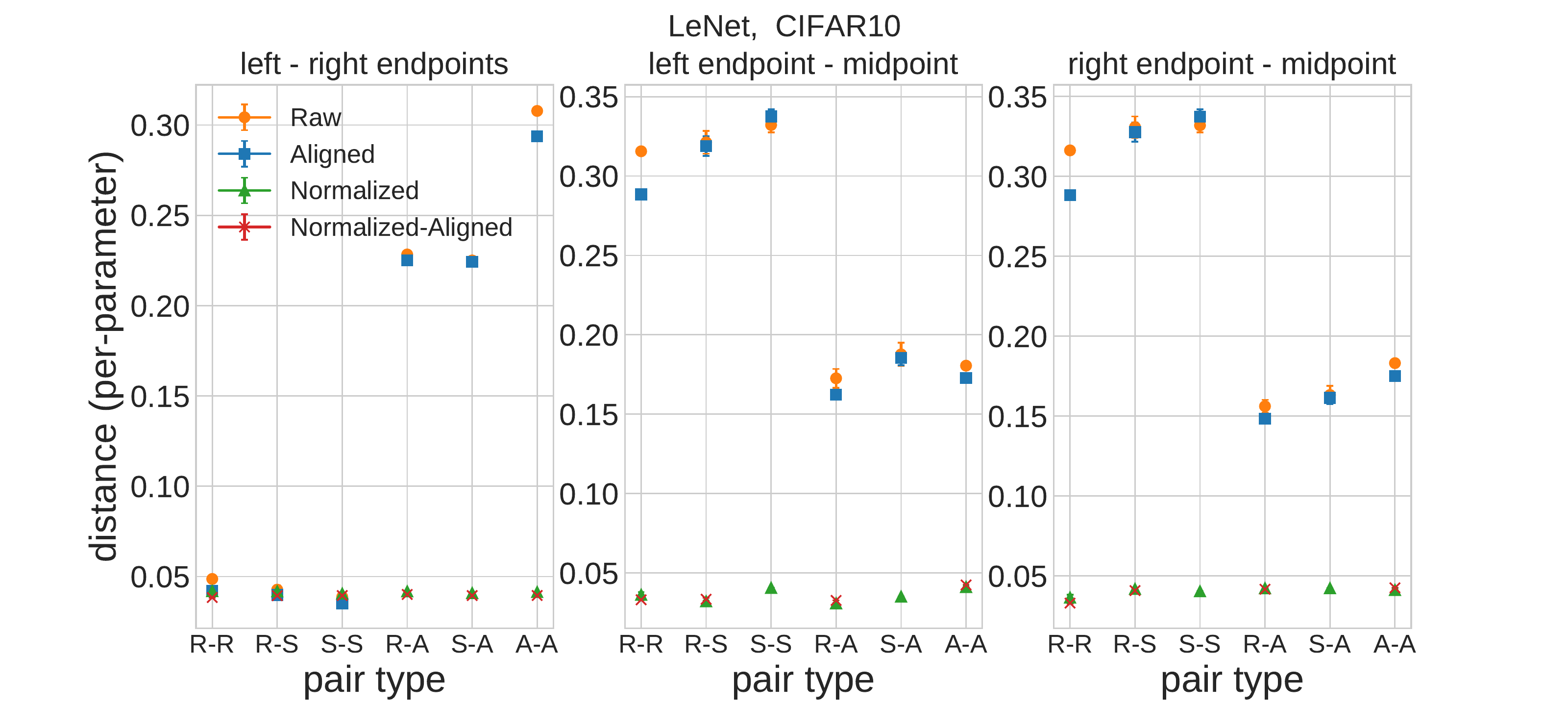}
    \caption{Distances between optimized configurations. For all networks/datasets: (Left panel) distance between configurations independently sampled by different algorithms (on the $x$-axis: R$\equiv$RSGD; S$\equiv$SGD; A$\equiv$ADV). (Middle panel) distance between the optimized midpoint initialized as the mean of the two configurations in the $x$-axis with the one indicated on the left; (left panel) same as the middle panel but the distance is between the optimized midpoint and the configuration indicated on the right.}
    \label{figSI:dist}
    \end{center}
    \vskip -0.2in
\end{figure}

We report in Fig.~\ref{figSI:dist} the distances between configurations and optimized midpoints for the networks/datasets studied in the main paper.






\section{Neural Networks with Binary weights\label{SI:binaryNNs}}
Here we provide implementation details for the experiments presented in Section~\ref{sec:binaryNNs}, as well as additional numerical tests.
In all our experiments we used the BinaryNet training scheme (see e.g. \citet{simons2019review}), which is a variant of SGD tailored to binary weights. Notice that our implementation differs from the original one~\cite{hubara2016binarized} because we use binary weights in the whole network, including the output layer.
In all the cases, for each solution type (ADV, SGD, RSGD) we were able to obtain solutions with zero or close to zero ($<1\%$) train error.

\paragraph{Local Energies}
The local energy provides a robust measure of the flatness of a given solution. It has been shown to be highly correlated with generalization errors \cite{Jiang2020Fantastic, pittorino2021}.
It is defined as the average error as a function of the distance from a reference solution.
In the case of binary weights neural networks,
we perturb a solution by changing sign to a random fraction $\varepsilon$ of the weights, and measure the train error for random choices of the perturbed weights.
$\delta E(w, \varepsilon) = \mathbb{E}_{\varepsilon}\left[ E(w, \varepsilon)\right] - E(w, \varepsilon=0)$, where $E(w, \varepsilon=0)$ is the train error of the unperturbed solution.
By varying $\varepsilon$ we are able to obtain the profile of the errors as a function of the hamming distance from the reference solution. 
In all the experiments, for each value of $\varepsilon$ we average the errors over $10$ independent realizations of the choice of the perturbed weights.

\paragraph{Random Linear Paths and Optimized Paths}

In order to explore the barriers between different solutions we measured the train error along the shortest paths connecting them. Given a source solution and a destination solution, we simply count the number of weights with different sign (that corresponds to the extensive hamming distance), progressively change the sign of those weights in the source solution in order to
approach the destination solution, and measure the errors along the path. 
In all the experiments we consider $3$ solutions for each ADV, SGD, RSGD (see Sec.~\ref{sec:hunt}).
The reported paths are averaged over all the $6$ possible paths among solutions of a given type (back and forth), and over $5$ realizations of the random path (the weight flipping order).

Optimized paths are random linear paths with one bend. We take a weight vector located at equal distance between two solutions, and use it to initialize SGD. In all the experiments the middle point has been optimized using the same parameters of SGD-type solutions. We then report the random linear paths between the source and the optimized middle point, and the optimized middle point and the destination. For each of the $6$ possible couples of solutions we averaged over $3$ independent choices of the middle point and $5$ realization of the random paths.

\paragraph{Shallow binary networks and the Hidden Manifold Model}

We trained both a binary perceptron and a binary fully-connected committee machine (CM) on synthetic data generated by the so-called Hidden Manifold Model (HHM) \cite{goldt2020modeling, gerace2020generalisation, baldassi2021learning}, also known as the Random Features Model \cite{mei2019generalization}.

In the HMM the data are first generated in a $D$-dimensional manifold, where a perceptron teacher $\{w^{T}_i\}_{i=1}^D$ assigns to a set of $P$ random i.i.d. patterns $\{\boldsymbol{\xi}^{\mu}\}_{\mu=1}^P$ a label, according to $y^{\mu} = \text{sign}\left( \boldsymbol{w}^T \cdot \boldsymbol{\xi}^{\mu} \right)$. Then the student sees a non-linear random projection of the data into an N-dimensional manifold: $x_{i}^{\mu} = \text{sign}\left(\sum_{k=1}^{D} F_{ik} \xi_{k}^{\mu} \right)$, where $F \in \mathbb{R}^{N \times D}$ is a fixed random matrix with i.i.d. elements, and classify them according to the teacher label.
In our experiments we fixed the size of the perceptron teacher (i.e. the dimension of the data points in their hidden manifold) to $D=501$ and the pattern number to $P=1503$, so that we are working at $\alpha_{T} \equiv P/D=3$. By increasing the size $N$ of the projection, we are able to study the system in the over-parameterized regime ($\alpha_{D} \equiv D/N \to 0$) (see also \cite{baldassi2021learning}).
In the case of the CM we fixed the hidden layer size to $H=101$ and vary the input size $N$.

\begin{figure}[ht]
    \begin{center}
    \includegraphics[width=0.23\linewidth]{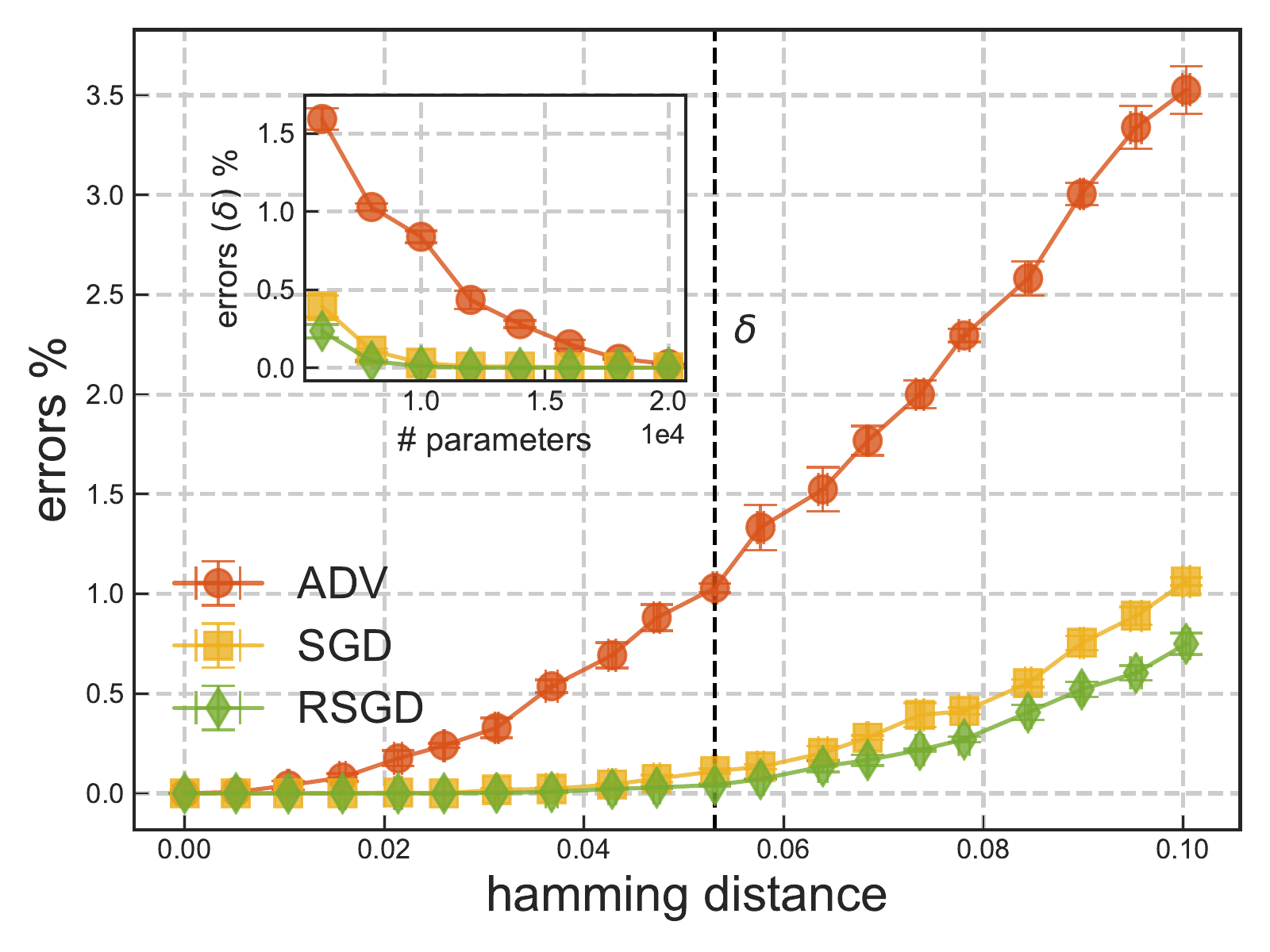}
    \includegraphics[width=0.23\linewidth]{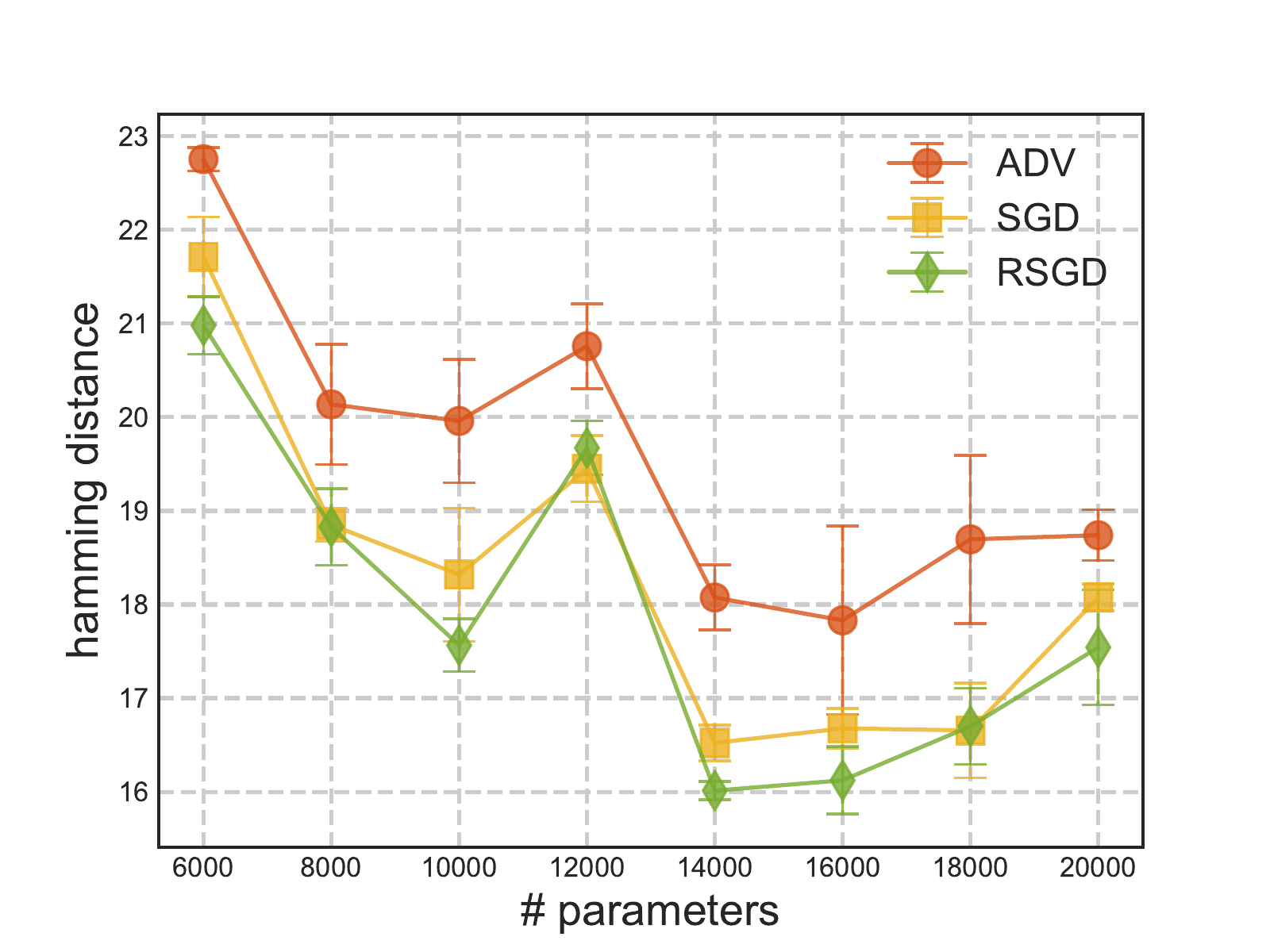}
    \includegraphics[width=0.23\linewidth]{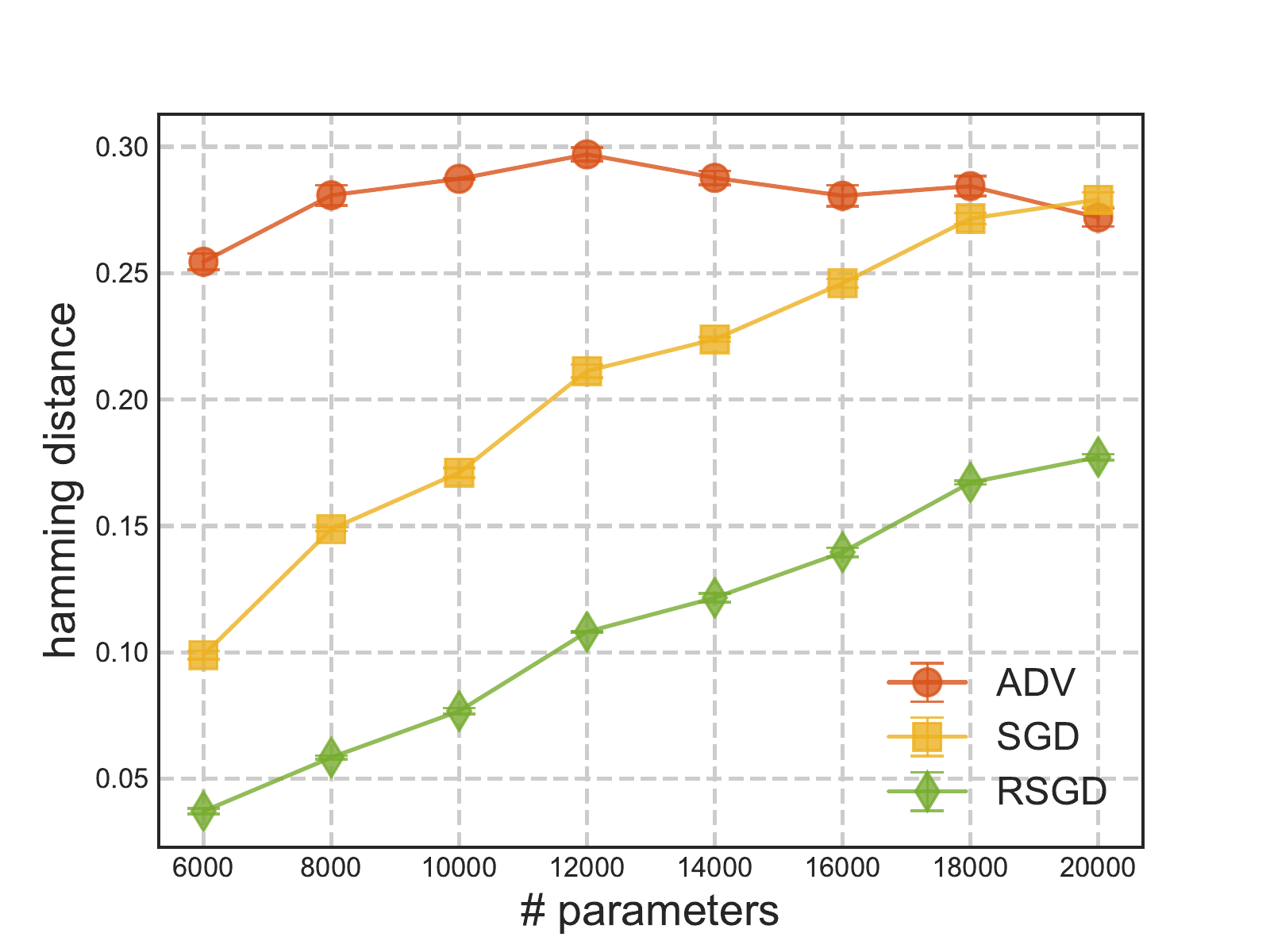}
    \includegraphics[width=0.23\linewidth]{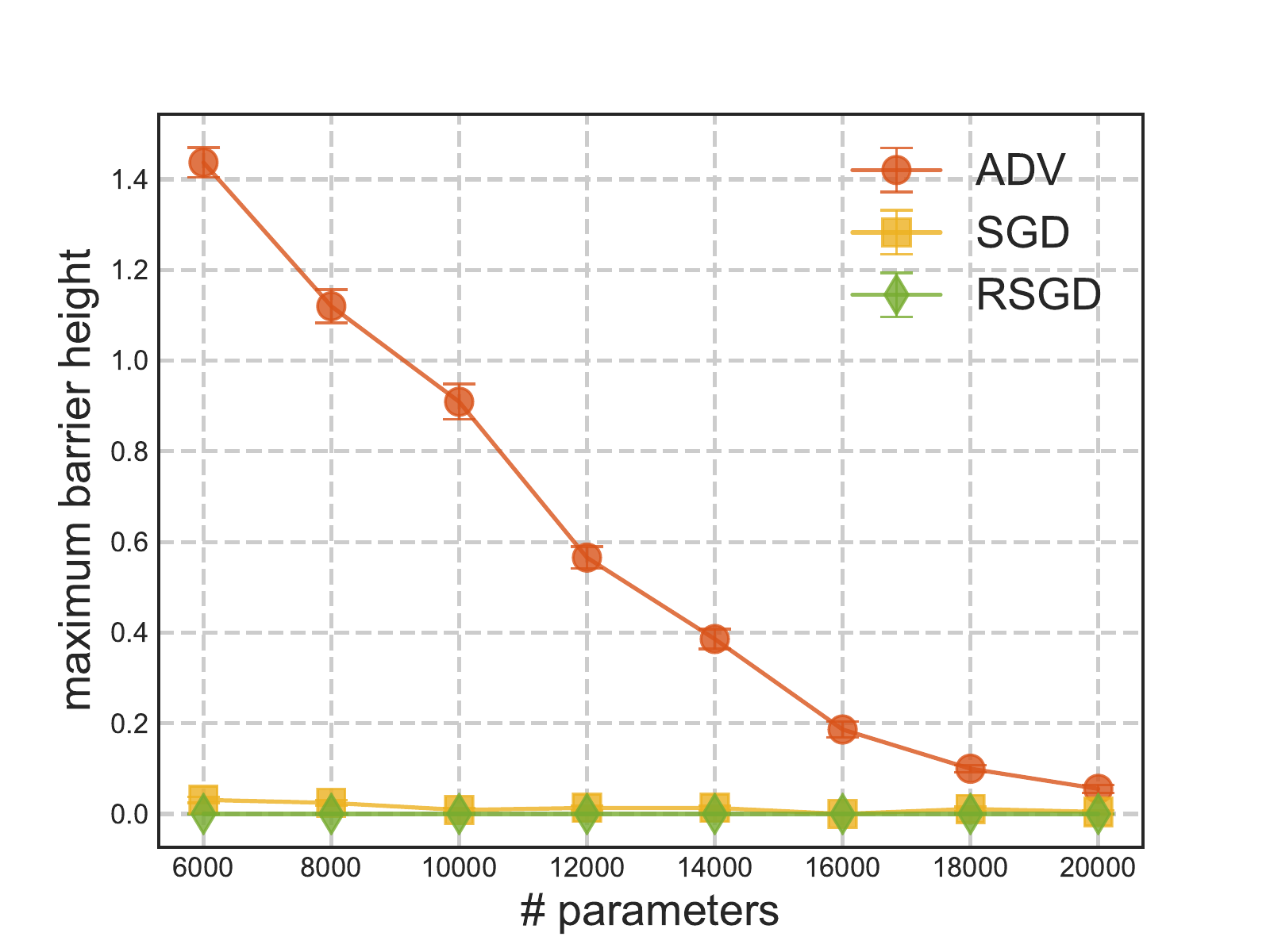}
    \caption{Binary Perceptron trained on data generated by the Hidden Manifold Model (see also Fig.~\ref{fig:binary_cm_hmm}) (Left)
    Local energy of different solutions. Inset: local energy at a fixed distance ($\delta$) as a function of the number of parameters. (Center Left) Generalization errors of different solutions as a function of the number of parameters. (Center Right) Average hamming distance of different solutions. (Right) Average maximum train error (percentage) along random linear path connecting different solutions.}
    \label{fig:binary_perc_hmm}
    \end{center}
\end{figure}

In Fig.~\ref{fig:binary_perc_hmm} we report the results for the binary perceptron, analogous to the ones presented in the main text in Fig.~\ref{fig:binary_cm_hmm}. In the perceptron case there is no redundancy in the function expressed in the student model, so that solutions do not need to be aligned.
As for the CM, in the limit $D/N \to 0$ we can see that solutions get flatter and the maximum error along random paths connecting them approaches zero. Even in this case, we observe a strong correlation among flatness (as measured with the local energy), generalization errors and maximum barrier heights.

We report training parameters for both models. Each model is trained with batchsize $100$ for $200$ epochs with binary cross entropy loss and SGD without momentum. 
For both the binary perceptron
and the CM we used the following parameters: 
(SGD) lr$=1.0$, 
(RSGD) lr=$1.0$, with $5$ replicas coupled with an elastic constant $\gamma(t) = 0.002 * (1.002)^t$, where $t$ is the epoch.
(ADV) networks have been first trained on a modified train set with randomized labels with lr$=10.0$ for $500$ epochs. The resulting configuration is used as an initial condition and has been optimized with SGD for $200$ epochs and lr$=5.0$.

\paragraph{Binary Multi-layer Perceptrons}
We consider two binary Multi-layer perceptron (MLP) architectures: a) a two hidden layer MLP trained on $10$k MNIST images (binary classification of the parity of digits), and b) a three hidden layer MLP trained on FashionMNIST. 
For architecture a) we analyzed the effect of removing symmetries as the network size is increased, by increasing the width of the hidden layers, while MLP b) has three hidden layer of fixed size $H=1001$. 
In case~a), for all hidden layer widths, we optimized the binary cross-entropy loss for $1000$ epochs using SGD with no momentum, with fixed batchsize $100$ and lr=$10.0$. For the ADV solutions we used as a starting point for SGD the results of $1000$ epochs of SGD optimization on a train set where the label have been randomized. For RSGD solutions we used $5$ replicas coupled with an epochs-dependent elastic constant $\gamma(t) = 0.002*(1.002)^t$.
In case b) we used Adam optimization with lr=$0.001$ for both SGD and RSGD solutions. For RSGD we used the same number of replica and elastic constant as in case a). The ADV solutions have been obtained by first training with SGD  with no momentum and lr=$5.0$ for $500$ on the train set with random labels, and then for other $1000$ epochs with the same lr and optimization algorithm.

Results for case b) are reported in the main text (see Fig.~\ref{fig:binary_deep_paths},~\ref{fig:binarymlp_fashion_plane}).

\begin{figure}[ht]
    \begin{center}
    \includegraphics[width=0.23\linewidth]{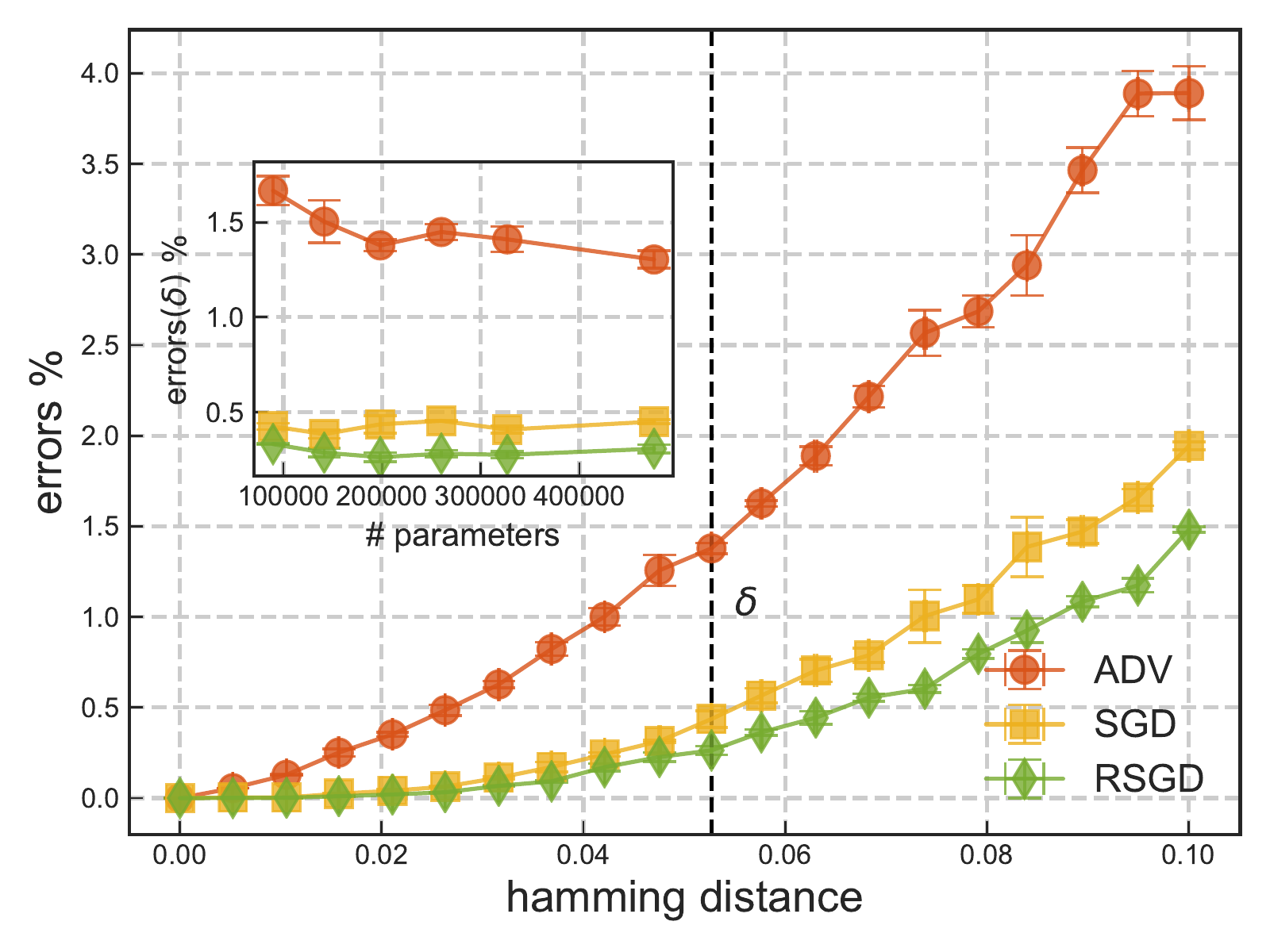}
    \includegraphics[width=0.23\linewidth]{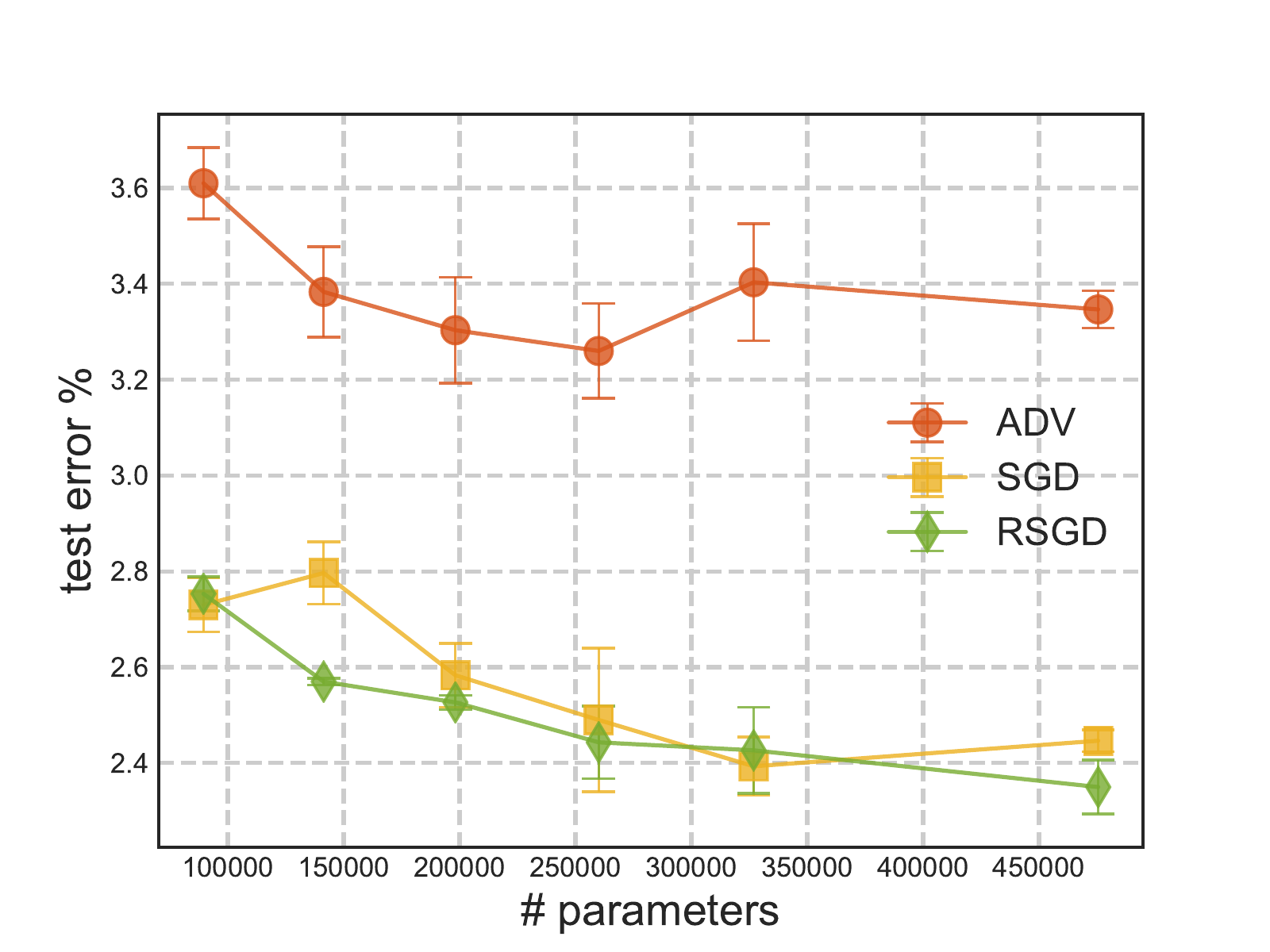} 
    \includegraphics[width=0.23\linewidth]{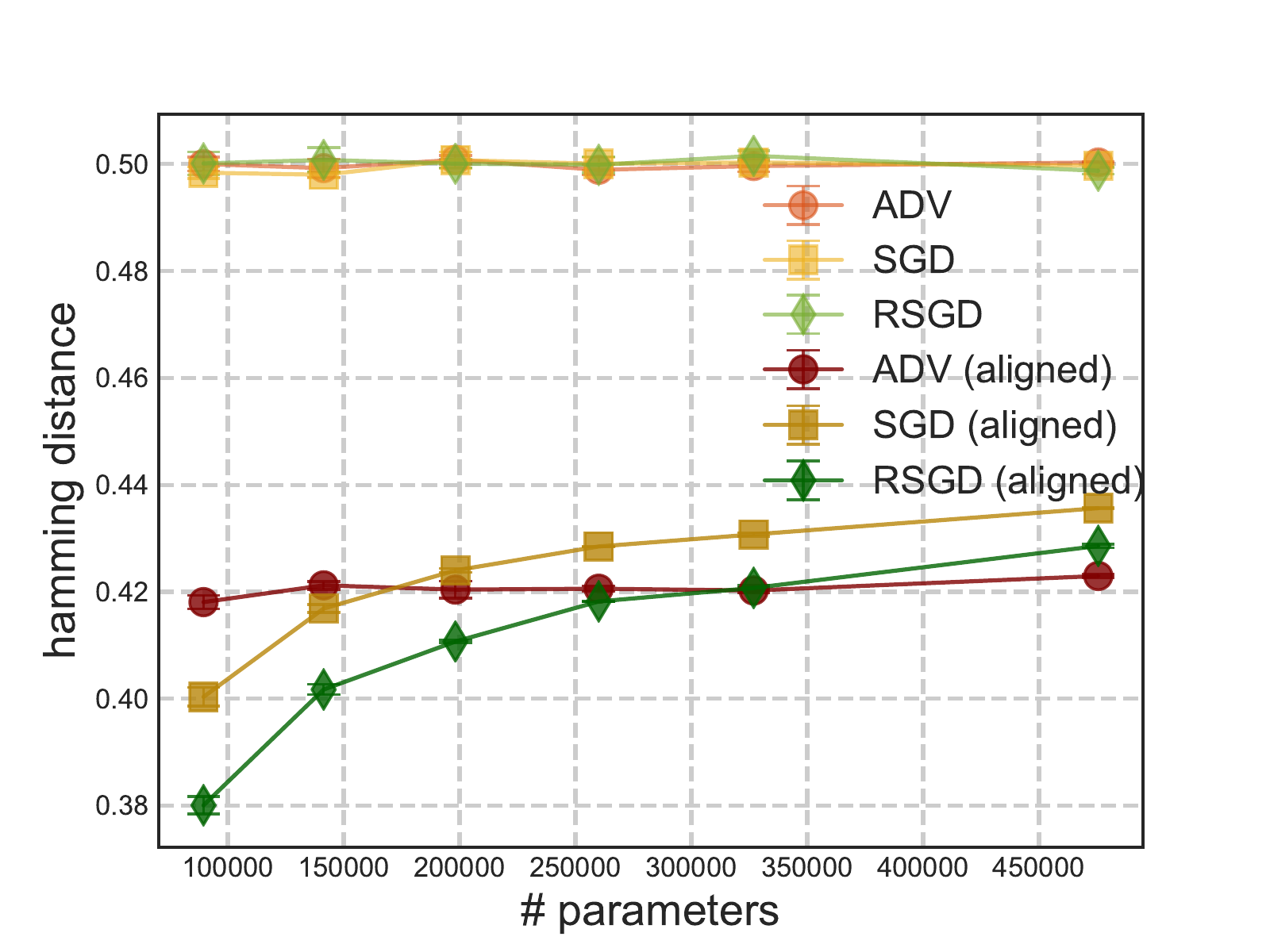}
    \includegraphics[width=0.23\linewidth]{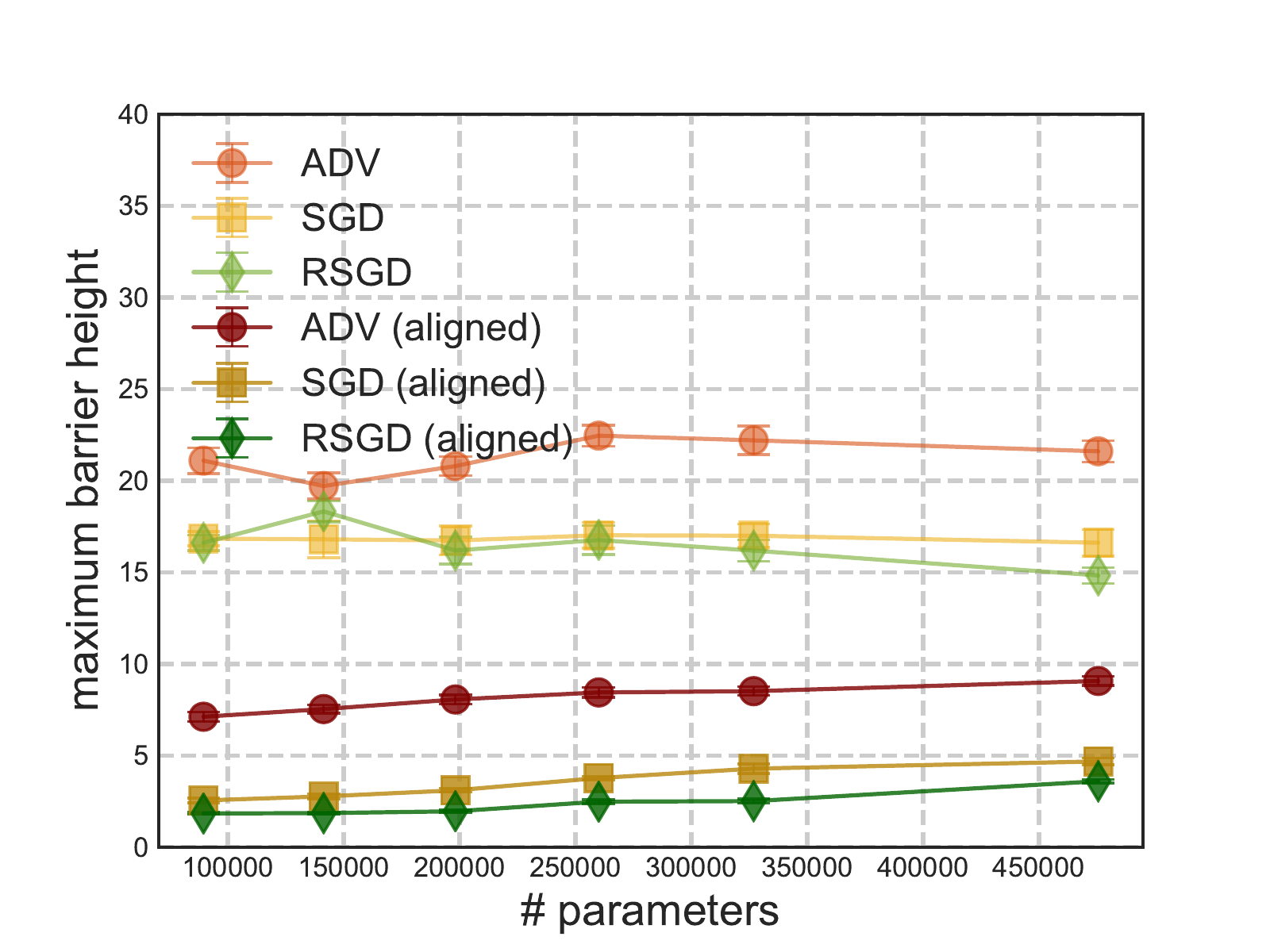}
    \caption{Binary MLP with $2$ hidden layers with $H$ units each, trained on $10$k examples of the MNIST dataset (binary classification). (Left) Local energy for the tree type of solutions considered at $H=201$. Inset: local energy at a fixed distance $\delta$ as a function of the number of parameters. (Center Left) Test errors for the three type of solutions as a function of the number of parameters. (Center Right) Average hamming distance among solutions, before (light colors) and after (dark colors) solutions have been aligned. (Right) Maximum barrier heights (in percentage of train errors) along random linear paths connecting solutions, before and after solutions have been aligned.}
    \label{fig:binary_mlp_oemnist}
    \end{center}
    \vskip -0.2in
\end{figure}

In case a) we performed analogous experiments as the one presented for shallow binary NNs on the HMM (see Fig.~\ref{fig:binary_perc_hmm},~\ref{fig:binary_cm_hmm}). The results are reported in Fig.~\ref{fig:binary_mlp_oemnist}. As in the simpler synthetic dataset scenario, there is a strong correlation between flatness, generalization errors and barrier heights. However, while in this case removing the permutation and sign-reversal symmetries between solutions considerably lowers the respective barriers and average distances, the barriers do not seem to approach zero error in the limit of a large number of parameters.

\paragraph{Binary Convolutional Network}

We trained a $5$ layers convolutional networks with binary weights. The first two layers are convolutional layers with $20$ and $50$ $5\times5$ filters with no padding, each followed by a $2\times2$ maxpool layer and sign activation function. The two convolutional layers are followed by $3$ fully connected layers with $2001$ units each. We used the following optimization parameters: (SGD) we trained the model for $400$ epochs with Adam optimization with batchsize=$128$ and lr=$0.005$ that is multiplied by $0.5$ every $50$ epochs. (RSGD) We used $3$ replicas coupled with an elastic constant $\gamma(t) = 0.001*(1.001)^t$, where $t$ is the epoch. All the other parameters are the same as SGD, except that we optimized for $300$ epochs.
(ADV) We first initialize the solutions by optimizing a modified train set where the labels have been randomized for $200$ epochs with SGD with lr=$10.0$, and then train them for $200$  epochs using Adam with lr=$0.01$ that is halved at epoch $10, 20, 50$ and then every $50$ epochs.

\begin{figure}[ht]
    \begin{center}
    \includegraphics[width=1.0\linewidth]{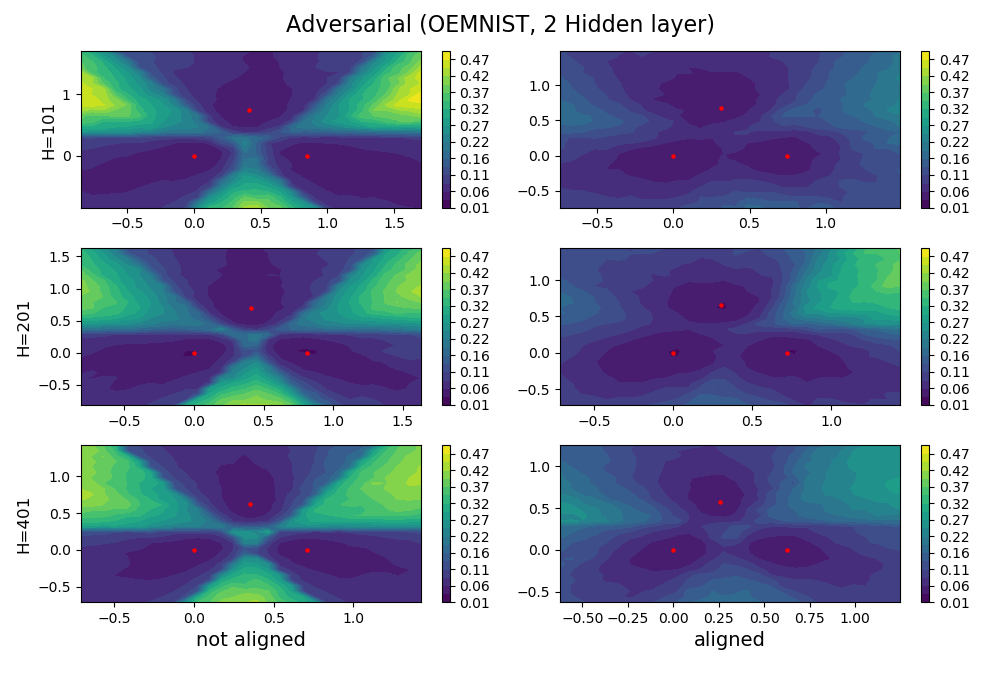}
    \caption{Bi-dimensional error landscape for ADV solutions of a $2$-hidden layer MLP trained on MNIST. From top to bottom the width of hidden layers is increased. Left column: raw solutions. Right column: top and right solutions have been aligned to the left solutions.}
    \label{fig:binary_mlp_oemnist_plane_adv}
    \end{center}
\end{figure}

\begin{figure}[ht]
    \begin{center}
    \includegraphics[width=1.0\linewidth]{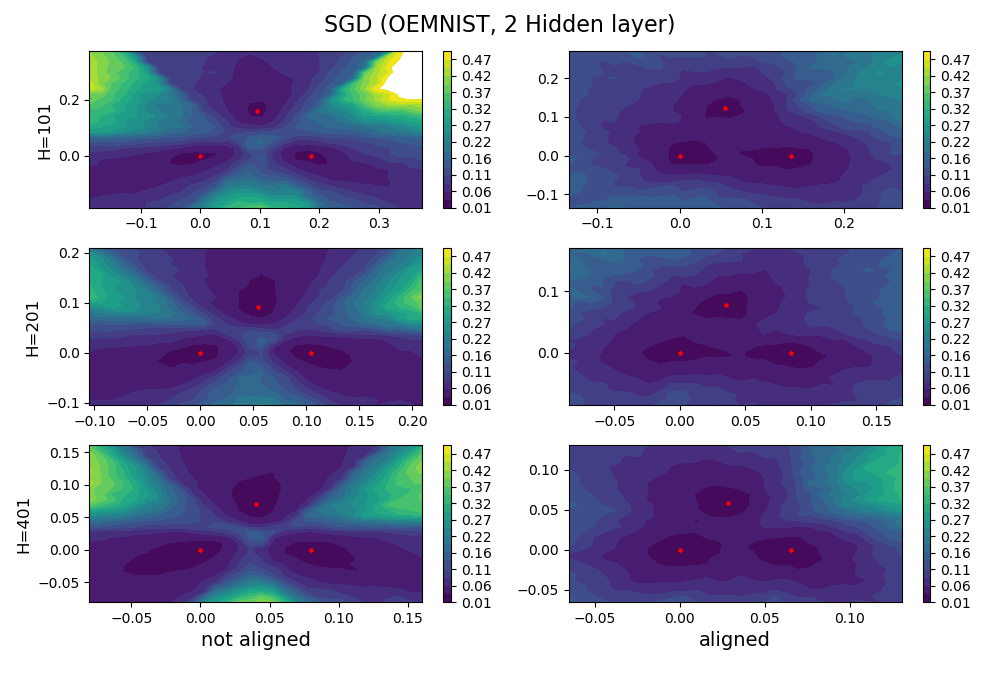}
    \caption{Bi-dimensional error landscape for SGD solutions of a $2$-hidden layer MLP trained on MNIST. From top to bottom the width of hidden layers is increased. Left column: raw solutions. Right column: top and right solutions have been aligned to the left solutions.}
    \label{fig:binary_mlp_oemnist_plane_sgd}
    \end{center}
\end{figure}

\begin{figure}[ht]
    \begin{center}
    \includegraphics[width=1.0\linewidth]{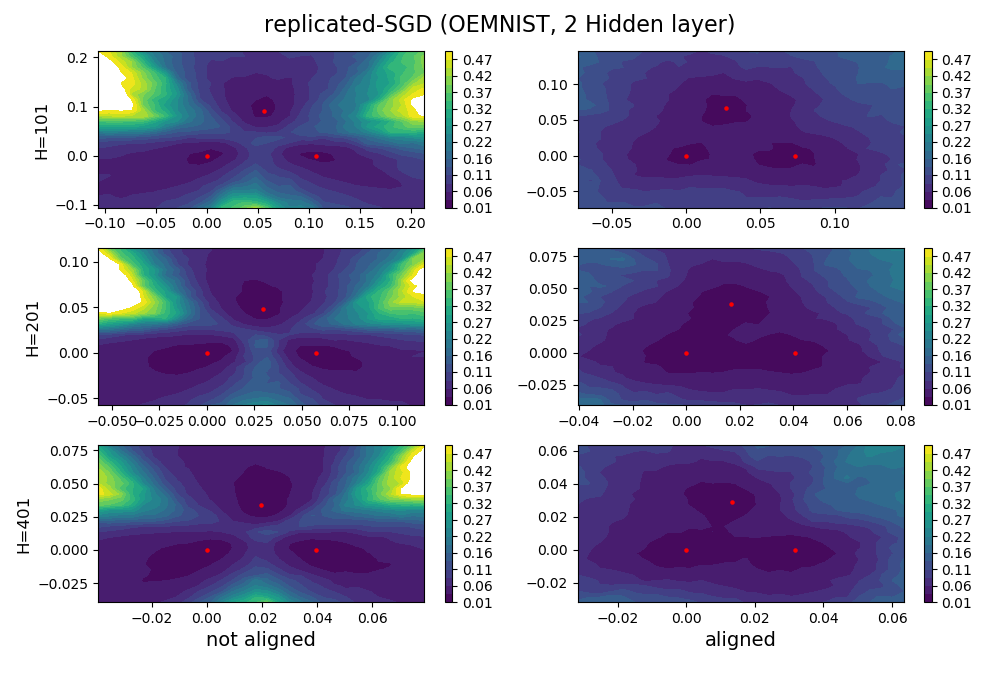}
    \caption{Bi-dimensional error landscape for RSGD solutions of a $2$-hidden layer MLP trained on MNIST. From top to bottom the width of hidden layers is increased. Left column: raw solutions. Right column: top and right solutions have been aligned to the left solutions.}
    \label{fig:binary_mlp_oemnist_plane_rsgd}
    \end{center}
\end{figure}

\begin{figure}[ht]
  \centering
    \includegraphics[width=1.0\linewidth]{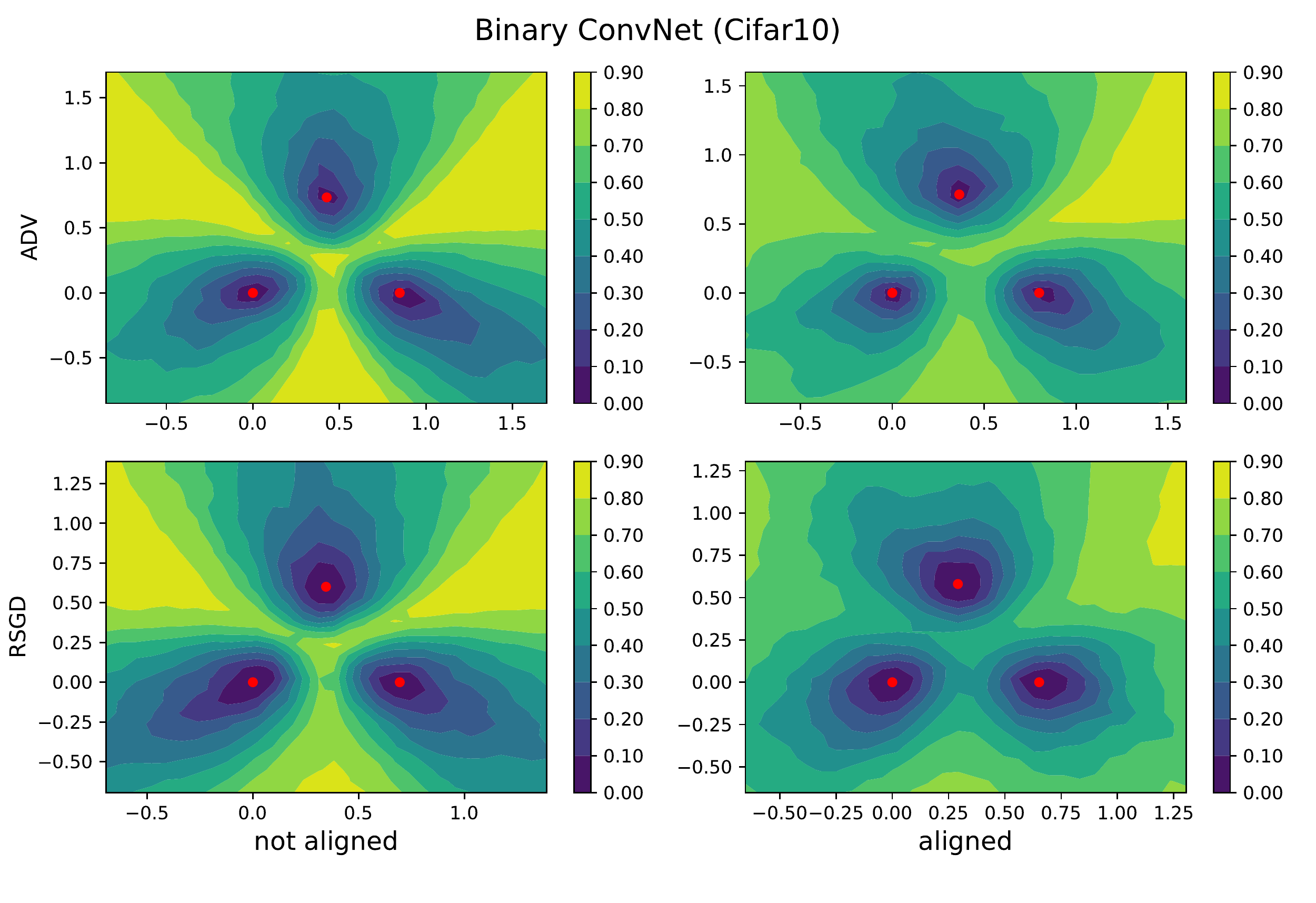}
  \caption{Train errors in the plane spanned by three solutions (for both ADV and RSGD solutions) for a binary convolutional NN trained on CIFAR-10 (see also Fig.~\ref{fig:binarymlp_fashion_plane}). Going from left to right panels one can appreciate the effect of removing symmetries in the error landscape.
  }
  \label{fig:binaryconv_cifar_plane}
\end{figure}

\paragraph{Bi-Dimensional Error Landscapes}
We describe the procedure to produce the bi-dimensional error landscape plot 
reported in the main text
(Fig.~\ref{fig:binarymlp_fashion_plane}).
Given three solutions, we pick the continuous weights associated with the binary ones\footnote{see \cite{hubara2016binarized} for more insights on the relation between continuous and binary weights in the BinaryNet optimization scheme} and use them to construct an orthonormal basis using the Gram-Schmidt procedure (as for the case of continuous NNs, Sec.~\ref{subsec:continuous_landscape}). At each point in the plane, we binarize the weights by taking their sign, and report the train error.
With this bi-dimensional projection of the error landscape one can graphically appreciate the effect of removing symmetries. 
In Fig.~\ref{fig:binary_mlp_oemnist_plane_adv},~\ref{fig:binary_mlp_oemnist_plane_sgd},~\ref{fig:binary_mlp_oemnist_plane_rsgd} we report the error landscape for the three types of solutions considered (ADV, SGD, RSGD) in the case of the $2$-hidden layer MLP trained on MNIST, as a function of the number of parameters. Without taking into account symmetries, the solutions appear to be isolated, while their flatness increase. However, once solutions are aligned by removing symmetries, a different landscape appears, where they are connected with low errors paths.
In Fig.~\ref{fig:binaryconv_cifar_plane} we show the effect of removing symmetries for ADV and RSGD solutions of the binary convolutional network (analogous to Fig.~\ref{fig:binarymlp_fashion_plane}).

\end{document}